\documentclass[10pt,twocolumn,letterpaper]{article}
\usepackage[T1]{fontenc}

  \usepackage[pagenumbers]{cvpr}   %

\usepackage{array} %
\usepackage{placeins} %

\usepackage{newtxtext}
\usepackage{indentfirst}
\usepackage{etoc}

\newcommand{\method}{\textsc{DriveWild3R}\xspace}
\newcommand{\bench}{\textsc{DriveWild3R-Bench}\xspace}
\newcommand{\G}{\mathcal{G}}
\newcommand{\views}{\mathcal{V}}
\newcommand{\render}{\mathcal{R}}

\definecolor{linkblue}{rgb}{0.21,0.49,0.74}
\usepackage[pagebackref,breaklinks,colorlinks,allcolors=linkblue]{hyperref}

\usepackage{xcolor}

\title{\method: Feedforward Transient Removal and Compact Reconstruction of Multi-Traversal Driving Data}

\author{%
  Finn K\"ohler$^{1*}$\quad
  Chensheng Peng$^{1*}$\quad
  Jerry Shang$^{1}$\quad
  Aaditya Agnihotri$^{1}$\\
  Thomas Monninger$^{2}$\quad
  Quentin HERAU$^{3}$\quad
  Masayoshi Tomizuka$^{1}$\quad
  Wei Zhan$^{3,1}$\\[0.6em]
  {\normalsize $^{1}$University of California, Berkeley\qquad
  $^{2}$Mercedes-Benz Research \& Development North America\qquad
  $^{3}$Applied Intuition}\\[0.2em]
  {\normalsize $^{*}$Equal contribution}%
}

\begin{document}
\etocdepthtag.toc{mainpaper}

\twocolumn[{%
\renewcommand\twocolumn[1][]{#1}%
\maketitle
\begin{center}
  \vspace{-1.2em}
  \centering
  \captionsetup{type=figure}
  \includegraphics[width=\textwidth]{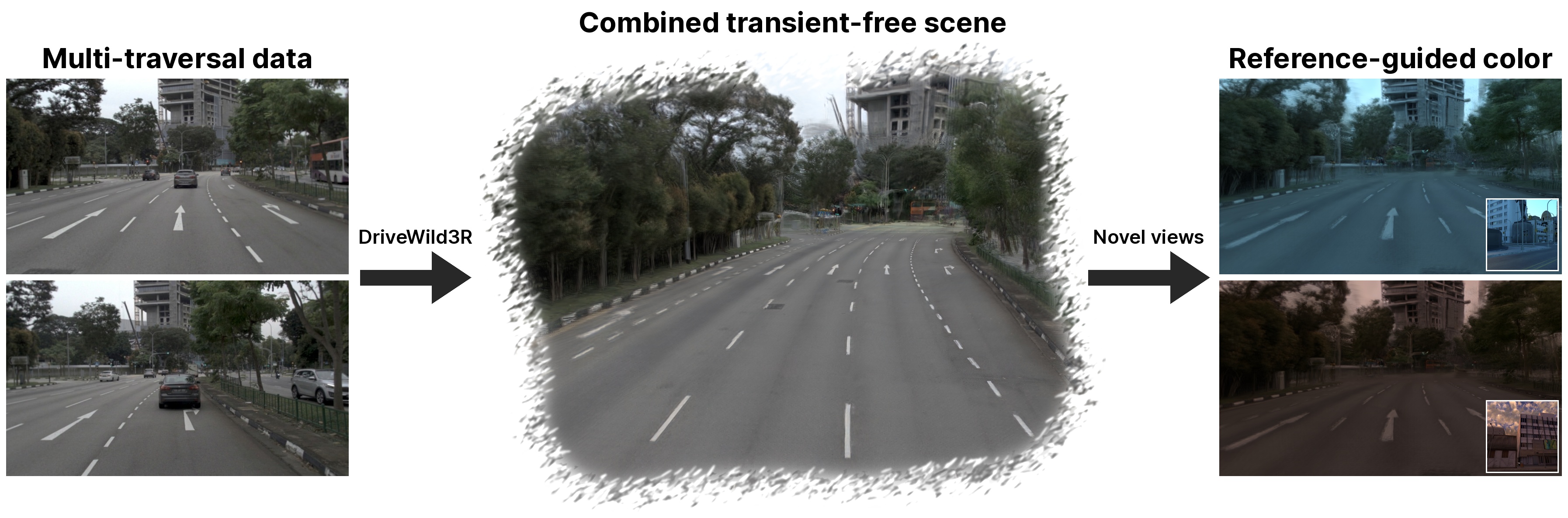}
  \captionof{figure}{\textbf{One transient-free scene from multiple drives.} Multi-traversal data of the same road see different traffic (left). \method combines them in a single feedforward pass into one shared Gaussian scene without the vehicles (center). Novel views of that scene can be rendered in the lighting and color of any reference image (right, references inset), while the geometry stays the same.}
  \label{fig:teaser}
\end{center}
\vspace{1em}
}]

\begin{abstract}
Multi-traversal drives observe the same road block under changing traffic and lighting, offering more observations from various viewpoints compared to single-traversal data. 
Existing feedforward 3D Gaussian Splatting approaches reconstruct a scene within seconds, yet on such data they either reproduce the transient vehicles in their input views or bake them into the scene as ghosting artifacts.
We present \method, a feedforward framework that reconstructs a shared, transient-free scene with controllable lighting from multi-traversal driving data. We first augment the feedforward predictor with a mask head that explicitly identifies transient objects, and with an entropy-aware sampling strategy that substantially reduces redundant Gaussians. 
We then introduce a lighting branch that combines per-image lighting embeddings with per-Gaussian features to transfer the color pattern of an arbitrary target image. To further improve reconstruction quality, a two-stage recurrent refiner corrects first the geometry and then the appearance of the initial prediction. 
To evaluate reconstruction quality and transient removal directly, we curate \bench, a simulated benchmark with traffic-free ground truth under diverse lighting and weather conditions, on which \method outperforms the strongest feedforward baseline by 3.9\,dB PSNR on traffic regions.
On the real-world MTGS and Ithaca365 datasets, \method achieves the lowest LPIPS and highest PSNR.

  Project page, code, models, and \bench: \href{https://drivewild3r.github.io}{drivewild3r.github.io}.

\end{abstract}
\section{Introduction}
\label{sec:intro}

Autonomous driving fleets traverse the same roads, so surfaces occluded in one traversal may be visible in another, and the combined observations constrain the scene better than a single drive.
Yet each traversal also records different traffic and lighting: vehicles park, pass, and leave, and illumination, exposure, and white balance change across drives.
Applications such as novel-view synthesis and sensor simulation therefore require a reconstruction that keeps the persistent scene, discards transient content, and can be rendered under the lighting of any traversal.

Per-scene methods such as Multi-Traversal Gaussian Splatting (MTGS)~\cite{li2025mtgs} model repeated drives with a shared static scene and traversal-specific components, but optimizing a single scene takes about 40 minutes even for eight images.
Feedforward Gaussian models predict a scene in a single forward pass~\cite{charatan2024pixelsplat,tian2025drivingforward,yang2025storm,jiang2025anysplat}, yet most of them keep the vehicles they observe without removing them.
Wild3R~\cite{furutani2026wild3r} learns to suppress transients and to handle appearance variation, but evaluates transient removal only qualitatively, since real data provides no ground truth for what lies behind a vehicle.
On \bench, a simulated benchmark with traffic-free ground truth, we find that existing feedforward models either reproduce the vehicles in their input views or bake them into the scene as ghosts (Fig.~\ref{fig:carla}), and that a single prediction also leaves blurred roads and low-opacity holes.

In this paper, we present \method, a feedforward framework that reconstructs a shared, transient-free scene with controllable lighting from multi-traversal driving data with the following key components:

\noindent \textbf{Transient-aware compact initialization.}
Building on Wild3R~\cite{furutani2026wild3r}, we augment it with a mask head that explicitly predicts per-pixel transient probabilities from the backbone features.
The predicted masks prevent the subsequent stages from reintroducing the transient traffic.
Since lifting every pixel of eight views yields about 1.6M Gaussians, we adopt an entropy-aware sampling strategy that focuses the budget on textured structure and down-weights sky and transient pixels.
A Gaussian head trained with the sampler in the loop covers the discarded pixels, producing a compact cloud of 100k--400k Gaussians small enough for learned refinement and efficient reconstruction.

\noindent \textbf{Decoupled lighting transfer.}
We transfer lighting in a decoupled branch that combines per-image lighting embeddings with per-Gaussian features to predict an affine color transform toward the lighting of an arbitrary target image.
Since this branch modifies colors only, the geometry is unaffected by the choice of target, and the same branch relights other feedforward methods without retraining.

\noindent \textbf{Two-stage recurrent refinement.}
To correct the blur and holes of a single feedforward prediction, we adopt the render--compare--update principle of ReSplat~\cite{xu2026resplat} and refine the compact cloud in two stages.
The geometry steps compare each context view with the scene relit to that view's lighting, so that lighting differences between traversals are not mistaken for geometric errors.
The color steps then compare the scene with reference frames of the target traversal and update the color patterns to match the target.

On \bench, \method outperforms the strongest feedforward baseline by 3.9\,dB PSNR on the traffic regions of its input views.
On held-out traversals of real-world datasets, MTGS~\cite{li2025mtgs} and Ithaca365~\cite{diazruiz2022ithaca365}, it achieves the lowest LPIPS among all compared methods and, given target-drive reference frames, also the highest PSNR, while reconstructing a scene in under two seconds. 

In summary, our contributions are as follows:
\begin{itemize}
  \item An efficient feedforward framework for multi-traversal data that explicitly predicts transient masks and produces compact Gaussians through entropy-aware sampling.
  \item A decoupled lighting branch that transfers the appearance of an arbitrary target image by modifying colors and generalizes to other feedforward methods without retraining.
  \item A two-stage recurrent refiner that corrects the geometry first and then the appearance of the initial prediction.
  \item \bench, a simulated benchmark built in CARLA with pixel-aligned, traffic-free ground truth under diverse lighting and weather conditions, which enables a direct quantitative evaluation of transient removal and reconstruction.
\end{itemize}
\section{Related Work}
\label{sec:related}
\paragraph{Street scenes and repeated traversals.}
Urban reconstruction methods such as EmerNeRF~\cite{yang2024emernerf}, Street Gaussians~\cite{yan2024streetgaussians}, Periodic Vibration Gaussians~\cite{chen2023pvg}, DrivingGaussian~\cite{zhou2024drivinggaussian}, OmniRe~\cite{chen2025omnire}, and DeSiRe-GS~\cite{peng2025desire} optimize one drive at a time, in minutes to hours, and separate its static background from dynamic content with tracked objects, scene graphs, time-dependent parameters, or self-supervised decomposition.
Repeated traversals are combined with per-capture appearance embeddings in Block-NeRF~\cite{tancik2022blocknerf}, as outliers of a robust optimization in 3DGM~\cite{li2024memorize}, and as a shared static scene with traversal-specific dynamic and appearance components in MTGS~\cite{li2025mtgs}, which uses LiDAR initialization and depth supervision; all three are optimized per scene.
Outside driving, per-scene methods suppress distractors with robust losses~\cite{sabour2023robustnerf}, predicted uncertainty~\cite{ren2024nerfonthego}, or pretrained image features~\cite{sabour2025spotlesssplats}, relying on many observations of one scene.
We instead predict only the shared static scene feedforward, refine it in a few learned steps, use LiDAR for training only, and remove vehicles rather than model them.
We train on nuPlan~\cite{caesar2021nuplan} and on the repeated routes of Ithaca365~\cite{diazruiz2022ithaca365}, whose held-out road blocks also serve for evaluation.

\paragraph{Feedforward reconstruction and learned refinement.}
Feedforward methods amortize 3D Gaussian Splatting~\cite{kerbl2023gaussian} over a training collection: pixelSplat~\cite{charatan2024pixelsplat} samples depth from per-pixel distributions, MVSplat~\cite{chen2024mvsplat} builds a cost volume, and DepthSplat~\cite{xu2025depthsplat} adds monocular depth features, all for a few small-baseline views of static scenes with constant lighting.
Driving models predict Gaussians from surround-view rigs~\cite{tian2025drivingforward}, from a shared volume~\cite{miao2025evolsplat}, or with per-Gaussian velocities~\cite{yang2025storm}, but reconstruct a single drive and keep its vehicles.
AnySplat~\cite{jiang2025anysplat} handles uncalibrated collections with the attention of VGGT~\cite{wang2025vggt}, and Wild3R~\cite{furutani2026wild3r}, trained on WildCity, handles transients and appearance variation; we adapt Wild3R to calibrated, repeated drives and correct its prediction.
Learned refinement goes back to the learned gradient descent of DeepView~\cite{flynn2019deepview}; G3R~\cite{chen2024g3r} lifts rendering-loss gradients into a 3D representation of driving scenes, and ReSplat~\cite{xu2026resplat} predicts recurrent Gaussian updates from rendering-error feedback, whose update mechanism we adapt to our initializer and to repeated drives.
For compaction, optimized scenes merge Gaussians into hierarchies~\cite{kerbl2024hierarchical}, GGN~\cite{zhang2024ggn} pools redundant pixel-aligned Gaussians across views, and SparseSplat~\cite{zhang2026sparsesplat} keeps pixels with a probability that grows with local image entropy; we adopt its sampler, down-weight sky and transients, and train our Gaussian head with it in the loop.

\paragraph{Appearance variation and relighting.}
NeRF in the Wild~\cite{martinbrualla2021nerfw} conditions color on a learned per-image embedding, as do Gaussian-splatting methods for photo collections~\cite{zhang2024gsw,xu2024splatfactow}; closest to our color transform, WildGaussians~\cite{kulhanek2024wildgaussians} maps per-Gaussian and per-image embeddings through an MLP to a per-channel affine color transform.
Wild3R~\cite{furutani2026wild3r} instead inserts a reference image into its transformer, which we find also changes the predicted geometry, so we keep target-drive images away from the geometry (Sec.~\ref{sec:appearance}).
Concurrent feedforward methods condition colors on an embedding or on tokens of a target-appearance image~\cite{spectralsplat2026,wildsplatter2026,wildsplat2026}; SpectralSplat~\cite{spectralsplat2026}, on driving data, decodes colors from a pooled image embedding with a shared MLP, is supervised by synthetically relit pairs, and leaves multi-traversal data to future work.
Repeated drives provide real pairs of the same place under different lighting, which we combine with synthetic relit pairs and unpaired losses, and we bring the lighting into the refiner.
\section{Method}
\label{sec:method}

\subsection{Problem Formulation and Overview}
\label{sec:problem}
Let $\views=\bigcup_{r=1}^{R}\views_r$ contain context views from $R$ traversals of a road segment.
View $i$ consists of an image $I_i$, camera calibration $C_i=(K_i,T_i)$, and a traversal identifier $r_i$.
Here $K_i$ is the intrinsic matrix and $T_i$ maps world coordinates to camera coordinates.
Sparse LiDAR depth $Z_i$ and transient labels $M_i^*$ may be available during training.
We seek a shared static Gaussian scene $\G$ that can be rendered from a query camera $C_q$, including cameras on a held-out traversal.
The query images and their labels are excluded from reconstruction.

\begin{figure*}[t]
  \centering
  \includegraphics[width=\linewidth]{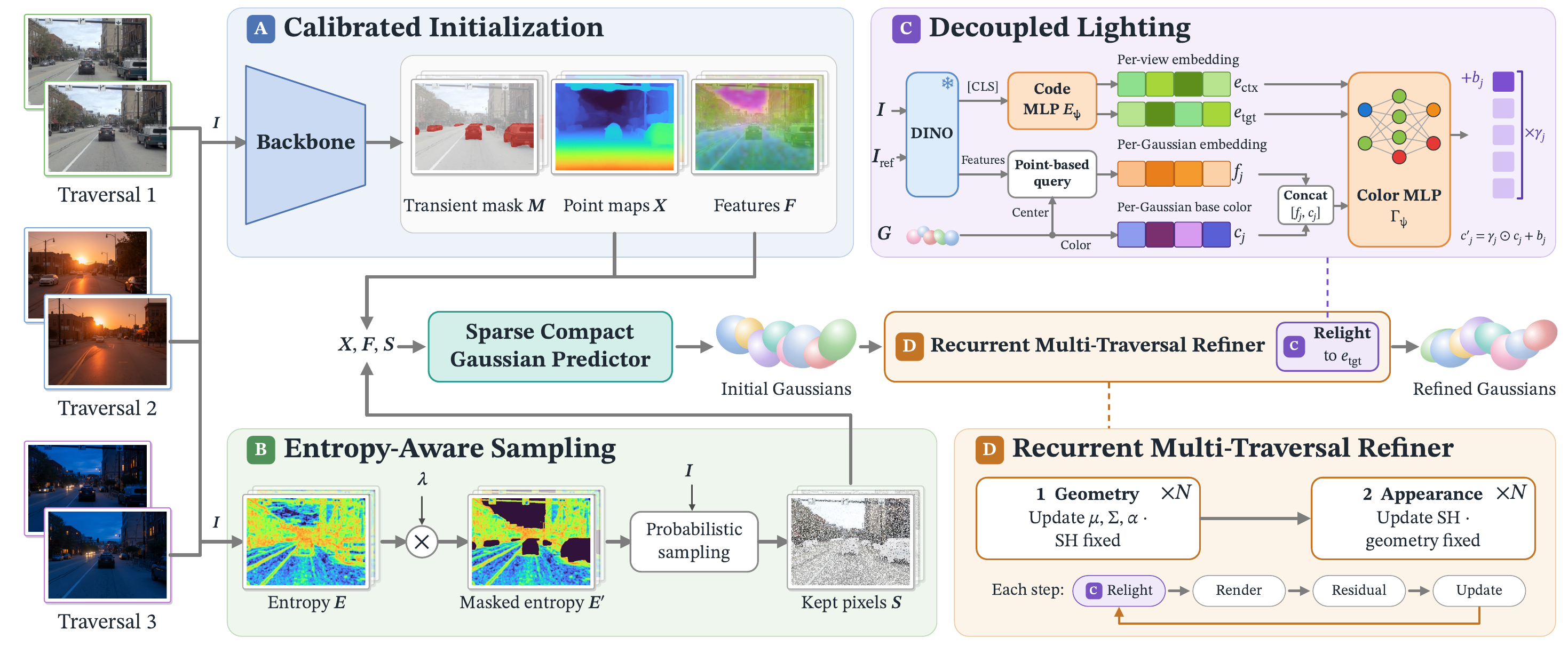}
  \caption{Overview of \method. A calibrated, transient-aware predictor initializes scene geometry; compact Gaussians are then updated using image evidence through a learned refiner. LiDAR and annotated transient masks provide training supervision.}
  \label{fig:pipeline}
\end{figure*}

The scene $\G$ is a set of $N_{\mathrm{g}}$ Gaussians, each with a center $\boldsymbol\mu_j$, covariance $\Sigma_j$, opacity $\alpha_j$, and appearance $\mathbf{c}_j$.
The renderer produces $\widehat I_i=\render(\G,C_i)$.
Our pipeline first predicts per-pixel Gaussian centers and transient probabilities, samples a subset of $B$ pixels, lets a Gaussian head predict the Gaussians at those pixels, and applies learned updates:
\begin{equation}
  \begin{aligned}
    (\mathcal{X},\widehat M)&=F_\theta(\views),\\
    \mathcal{S}&=\mathrm{Sample}_B(\views,\widehat M),\\
    \G^{(0)}&=H_\omega(\mathcal{X},\mathcal{S}),\\
    \G^{(n+1)}&=U_\phi(\G^{(n)},\views,\widehat M),\\
    \G^{\star}&=\mathcal{A}_\psi\bigl(\G^{(N)},\mathbf{e}_{\mathrm{tgt}}\bigr).
  \end{aligned}
  \label{eq:pipeline}
\end{equation}
The notation $F_\theta$ includes the image backbone, the depth head, which together with the cameras gives one 3D center per pixel ($\mathcal{X}$), and the separate mask head.
$\mathrm{Sample}_B$ selects the subset $\mathcal{S}$ of $B$ pixels that become Gaussians (Sec.~\ref{sec:compact}), and the Gaussian head $H_\omega$ predicts their attributes.
After $N$ refinement steps, the appearance branch $\mathcal{A}_\psi$ (Sec.~\ref{sec:appearance}) transfers every Gaussian's color to a target lighting code $\mathbf{e}_{\mathrm{tgt}}$; without a code, $\mathcal{A}_\psi$ is the identity.
Figure~\ref{fig:pipeline} summarizes the modules and supervision.

\subsection{Calibrated Feedforward Initialization}
\label{sec:initialization}
We initialize the image encoder, multi-view transformer, and reconstruction heads from Wild3R~\cite{furutani2026wild3r}, which predicts per-view depth and Gaussians from sparse context views and already tolerates transients and lighting changes.
It was trained on synthetic photo collections, not on driving data, so we adapt it: we train rank-16 low-rank adapters~\cite{hu2021lora} on the transformer and fine-tune the depth and Gaussian heads, keeping the pretrained prior.
Training renders views of a held-out traversal with an image loss on LaMa-filled targets, in which car pixels are replaced by an inpainting of the empty road~\cite{suvorov2022lama}, plus a coverage term $1-\alpha$ on non-sky pixels.

\paragraph{Camera priors and LiDAR.}
Driving datasets provide calibrated cameras and LiDAR, and we exploit both.
We lift every pixel $\mathbf{u}$ with its predicted depth $D_i(\mathbf{u})$ and the ground-truth camera,
\begin{equation}
  \mathbf{x}_i(\mathbf{u})=T_i^{-1}\bigl(D_i(\mathbf{u})\,K_i^{-1}\widetilde{\mathbf{u}}\bigr),
  \label{eq:unproject}
\end{equation}
where $\widetilde{\mathbf{u}}$ is the homogeneous pixel coordinate, and render with the same cameras; the transformer still receives images only.
Compared with Wild3R's own cameras, whose focal lengths are off by 4--6\,\%, this is worth about 1.5\,dB (Sec.~\ref{sec:backboneeval}).
During training, LiDAR supervises the predicted depth with an L1 loss on log depth; it is not used at inference.

\subsection{Explicit Transient Prediction}
\label{sec:transients}
Wild3R's features already separate content that is consistent across views from content that is not, but nothing downstream uses this.
Without explicit removal, the sampler spends Gaussians on vehicles, and the refiner, which is driven by image residuals, restores them, leaving ghosts and artifacts in the scene (Sec.~\ref{sec:transienteval}).
We make it explicit with a transient head that reads the distinction out of the features, and remove the predicted transients wherever an image is compared against.
We attach a small DPT-style head~\cite{ranftl2021dpt} to four intermediate layers of the transformer and predict a per-pixel transient probability $\widehat M_i(\mathbf{u})\in[0,1]$ for every view.
The head reads the features of the frozen backbone, so it adds masks without changing the transformer or its predicted Gaussians.
The masks of the context views come from the same features as the reconstruction and cost no extra pass; only the lighting references, which the reconstruction does not process, need a pass of their own.
The head is supervised with traffic that an image-editing model~\cite{wu2025qwenimage} inserts into clean WildCity~\cite{furutani2026wild3r} frames, pasted back onto the untouched frame for an exact mask, and with labeled traffic in nuPlan~\cite{caesar2021nuplan}.
At inference we threshold $\widehat M_i$ at 0.5.

Given the thresholded transient mask $\widehat M_i$, define the static observation weight
\begin{equation}
  W_i(\mathbf{u})=A_i(\mathbf{u})\bigl(1-\widehat M_i(\mathbf{u})\bigr),
  \label{eq:weights}
\end{equation}
where $A_i$ denotes valid image support after rectification.
The masks enter the pipeline wherever an image is compared against: they down-weight transient pixels when choosing which pixels become Gaussians (Sec.~\ref{sec:compact}), they remove transient pixels from the rendering errors that drive the refiner (Sec.~\ref{sec:refinement}), and they mask the lighting-reference frames (Sec.~\ref{sec:appearance}).
Masked regions are filled from other views and from the learned prior.

\subsection{Compact Gaussian Representation}
\label{sec:compact}
Lifting every pixel of eight context views gives about 1.6M Gaussians, far more than a learned refiner can handle.
A compact cloud makes refinement practical: at 400k Gaussians, four refinement steps peak at 11.5\,GiB instead of 40\,GiB at 1.6M, two unrolled steps become trainable, and the cloud resembles the compact inputs the original refiner was built for.
We therefore turn only a subset of $B$ pixels into Gaussians and predict their attributes with a head that is trained for this setting.
The budget $B$ is a free parameter: one head and one refiner serve any value, and we study $B$ from 100k to 400k.
Every Gaussian sits at the unprojected center of its source pixel and carries its view's index, which the lighting branch uses.

\paragraph{Entropy-aware pixel sampling.}
We follow SparseSplat~\cite{zhang2026sparsesplat} and keep each pixel with a probability $P(\mathbf{u})$ proportional to the entropy of its local intensity histogram, weighted by 0.1 on sky and transient pixels and scaled so that the expected number of kept pixels equals $B$.
The budget thus goes to textured static structure rather than to sky or transient objects.

\paragraph{Gaussian head for the sampled pixels.}
Dropping pixels leaves holes that the remaining Gaussians must cover.
We therefore train the Gaussian head of the backbone with the sampler in the loop, so that it learns to compensate for discarded pixels.
Following SparseSplat, which predicts a point's attributes from its 3D neighborhood, we condition the per-pixel head on the local sampling density.
Two additional input channels carry the keep probability $P$ and the realized density of kept pixels in a $17{\times}17$ window; their weights are zero-initialized, so training starts from the pretrained head.
The head runs on all pixels, and only the outputs at the kept pixels become Gaussians.
Their centers come from the depth head and the ground-truth cameras (Sec.~\ref{sec:initialization}), and the head predicts scale, rotation, opacity, and color.
Where few pixels survive, it learns to predict larger Gaussians, and where many survive, smaller ones.
The color is predicted as lit in the source view (Sec.~\ref{sec:appearance}).
The backbone is frozen during this stage.
To make one head serve any budget, each training step draws $B$ from \{100k, 200k, 400k\}; at inference any number of Gaussians can be sampled with the same weights (Sec.~\ref{sec:samplingeval}).
Each step renders views of a held-out traversal with the loss of Sec.~\ref{sec:initialization}, including the LaMa-filled car targets and the coverage term.

\subsection{Learned Multi-Traversal Refinement}
\label{sec:refinement}
\begin{figure}[t]
  \centering
  \includegraphics[width=\linewidth]{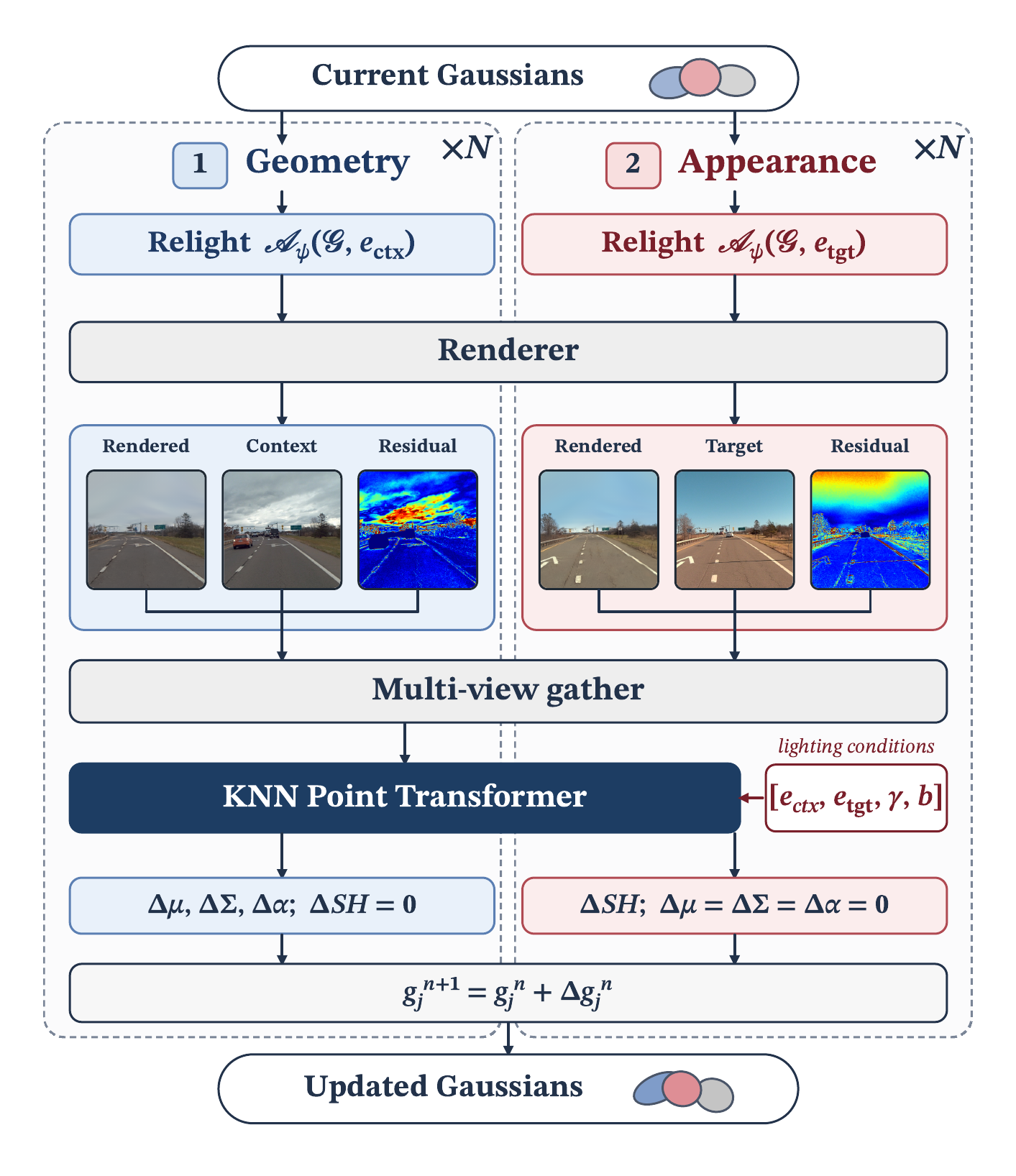}
  \caption{\textbf{Recurrent refinement.} Geometry steps (1) compare renders relit to each context view's code with the context images and update only means, covariances, and opacities. Color steps (2) run afterwards, compare with target-drive references, and update only the spherical harmonics.}
  \label{fig:recurrent}
\end{figure}

A single forward pass cannot check its prediction against the images, and it leaves blurred roads, low-opacity holes, and incomplete distant structure.
Per-scene optimization would fix this, but at minutes to hours per scene.
We therefore correct the compact cloud with a learned refiner that follows the render--compare--update principle of ReSplat~\cite{xu2026resplat}: it renders the Gaussians into the context views, compares the renders with the images, and predicts updates, in two steps and without per-scene gradient descent.
At step $n$, the feedback from context view $i$ is the mask-aware residual
\begin{equation}
  \rho_i^{(n)}=W_i\odot\bigl(I_i-\render(\G^{(n)},C_i)\bigr).
  \label{eq:feedback}
\end{equation}
A recurrent update reads this feedback for every Gaussian and predicts changes to its parameters and to a per-Gaussian hidden state:
\begin{equation}
  (\Delta g_j^{(n)},h_j^{(n+1)})=
  U_\phi(\G^{(n)},h^{(n)},\{\rho_i^{(n)}\},\views).
  \label{eq:update}
\end{equation}
Here $g_j=(\boldsymbol\mu_j,\Sigma_j,\alpha_j,\mathbf{c}_j)$ collects the parameters of Gaussian $j$, and $g_j^{(n+1)}=g_j^{(n)}+\Delta g_j^{(n)}$; updates use the appropriate parameterization for positive scales, rotations, and bounded opacity.
ReSplat initializes the hidden state $h^{(0)}$ from the multi-view features of its own image encoder (its condition features).
We predict these features with a lightweight adapter (10.8M parameters) instead, so the pipeline needs no second image encoder; it maps the backbone tokens and the compact head's features, normalized per channel with fixed statistics, to feature maps of the same shape and is trained to regress ReSplat's condition features; $h^{(0)}$ is read from these maps by the lookup described below.
Figure~\ref{fig:recurrent} shows the loop.

\paragraph{Position-based feedback lookup.}
As in ReSplat, each view's residual is encoded into a feature map $\mathbf{F}_i^{(n)}$ at a quarter of the image resolution (the difference between the features of the rendering and of the image plus the projected color residual), and one attention block over the tokens of all views exchanges information across views in image space; this attention is not geometry-aware.
It remains to map this pixel-level feedback to individual Gaussians.
ReSplat reads each Gaussian's feedback by index at a fixed cell of a quarter-resolution grid, but our Gaussians sit at irregularly sampled pixels, so some cells hold several Gaussians and many hold none.
We therefore replace the lookup by index with a lookup by position.
At every step, Gaussian $j$ is projected into the context views at its current position, giving pixel $\mathbf{u}_{ij}$ and camera depth $z_{ij}$, and it reads the feature maps of the views that actually observe it:
\begin{equation}
  \begin{aligned}
    \bar{\mathbf{F}}_j^{(n)}&=\sum_i w_{ij}\,\mathbf{F}_i^{(n)}(\mathbf{u}_{ij}),\\
    w_{ij}&\propto\mathbb{1}\bigl[\mathbf{u}_{ij}\in\Omega_i,\; z_{ij}\le(1{+}\delta)\,\widehat D_i^{(n)}(\mathbf{u}_{ij})\bigr],
  \end{aligned}
  \label{eq:gather}
\end{equation}
where $\Omega_i$ is the image domain, $\widehat D_i^{(n)}$ the depth rendered from the current Gaussians, $\delta{=}0.05$, and the weights are normalized over views.
The depth test excludes views in which the Gaussian is hidden behind another surface, so it does not receive the residual of whatever is drawn in front of it.
A Gaussian that passes the test in no view falls back to the in-frustum view whose rendered depth disagrees least.
The initial recurrent state $h_j^{(0)}$ is read from the conditioning maps in the same way.
Because the lookup depends only on the current Gaussian positions, it follows each Gaussian as it moves, and one refiner serves our irregular sampling at every budget.
The color steps of Sec.~\ref{sec:appearance} use the same depth-tested projection to read the residuals of the target-drive references.

\subsection{Decoupled Appearance Transfer}
\label{sec:appearance}
Besides sampling and refinement, reference frames of the target drive are the third place where the pipeline compares the scene against images, and like transients they must not reach the geometry.
A held-out traversal is recorded under different illumination, exposure, and white balance than the context drives, so a model that sees only the context can at best reproduce one of their lighting conditions.
Wild3R conditions appearance on a reference image inside the transformer, but that reference also changes the predicted geometry.
We therefore keep target-drive images away from the geometry and transfer lighting in a separate branch that only changes colors.

\paragraph{Lighting codes and per-Gaussian transfer.}
Every Gaussian $j$ is lifted from a pixel $\mathbf{u}_j$ of a context view $v_j$, so the head predicts its base color $\mathbf{c}_j$ as lit in that view.
A small MLP on frozen DINOv2 features~\cite{oquab2024dinov2} gives every image a lighting code and every Gaussian a local feature $\mathbf{f}_j$, read at its source pixel, so that, \eg, road and sky can respond differently to the same change of lighting.
Following WildGaussians~\cite{kulhanek2024wildgaussians}, a second MLP $\Gamma_\psi$ combines them with the base color and predicts a per-Gaussian affine color transform from the code $\mathbf{e}_{\mathrm{ctx}}$ of its source context view $v_j$ to a target code $\mathbf{e}_{\mathrm{tgt}}$:
\begin{equation}
  \begin{aligned}
    (\log\boldsymbol\gamma_j,\mathbf{b}_j)&=\Gamma_\psi\bigl(\mathbf{e}_{\mathrm{ctx}},\mathbf{e}_{\mathrm{tgt}},\mathbf{f}_j,\mathbf{c}_j\bigr),\\
    \mathbf{c}'_j&=\boldsymbol\gamma_j\odot\mathbf{c}_j+\mathbf{b}_j .
  \end{aligned}
  \label{eq:relight}
\end{equation}
WildGaussians optimizes its appearance model per scene; ours is pretrained once across scenes and runs feedforward.
With $\mathbf{e}_{\mathrm{tgt}}=\mathbf{e}_{\mathrm{ctx}}$ the transform is the identity, so without a target code every Gaussian keeps its source lighting.

\paragraph{Lighting inside the refiner.}
The geometry steps compare each context view with the cloud relit to that view's own code $\mathbf{e}_{\mathrm{ctx}}$, $\render(\mathcal{A}_\psi(\G^{(n)},\mathbf{e}_{\mathrm{ctx}}),C_i)$ in Eq.~\ref{eq:feedback}.
Lighting differences between drives therefore do not show up as residuals, which the refiner could only explain by editing geometry; these steps update only means, covariances, and opacities.
Afterwards, the color steps relight the cloud to $\mathbf{e}_{\mathrm{tgt}}$, compare it with reference frames of the target drive, and update only the spherical harmonics (Fig.~\ref{fig:recurrent}).
Because every geometry step finishes before a reference is used, the refined geometry is bit-identical with and without references.
The refined cloud is relit to $\mathbf{e}_{\mathrm{tgt}}$ once at the end (Eq.~\ref{eq:pipeline}).

\paragraph{Lighting references.}
Every other frame of the target drive serves as a reference, $\mathbf{e}_{\mathrm{tgt}}$ is their mean code, and only the remaining frames are scored, for every method.
Because the references lie next to the scored views, the color steps also correct view-dependent color; Suppl.\ Sec.~\ref{sec:supp_refdist} and Table~\ref{tab:refdist} separate this from lighting transfer.
Any image can supply $\mathbf{e}_{\mathrm{tgt}}$, so a street can also be rendered under, \eg, WildCity sunset or night lighting (Suppl.\ Fig.~\ref{fig:relight}).

\subsection{Training and Inference}
\label{sec:training}
Training proceeds in five stages, each with all earlier modules frozen: (1)~the backbone, (2)~the transient head, (3)~the compact Gaussian head, (4)~the appearance branch, and (5)~the refiner, fine-tuned from ReSplat's released weights with the appearance branch in the loop.
Losses and schedules are in the supplement.
At inference we use $N{=}2$ recurrent steps, two geometry and, with lighting, two color steps; more steps worsen LPIPS, even when the refiner is trained for them (Sec.~\ref{sec:steps}).
\section{Experiments}
\label{sec:experiments}
\paragraph{Real data and baselines.}
\label{sec:protocol}
We evaluate on 18 MTGS~\cite{li2025mtgs} and 30 Ithaca365~\cite{diazruiz2022ithaca365} splits from road blocks unseen in training, each with eight context views and query views on a drive that contributes no context, and compute PSNR, SSIM~\cite{wang2004ssim}, and LPIPS~\cite{zhang2018lpips} on static pixels.
Baselines are Wild3R, ReSplat, MVSplat, and DepthSplat with their released checkpoints, none trained on driving data, and MTGS on the same eight images; the supplement gives the full protocol.

\paragraph{\bench benchmark.}
\label{sec:carlabench}
Real data cannot show what is behind a removed vehicle, so we render 29 road locations in CARLA~\cite{dosovitskiy2017carla}, each target drive with and without traffic along the same poses, for a pixel-aligned traffic-free ground truth; context views and lighting references keep the traffic.
The 173 splits range from daylight with context in the target's lane to night and fog, and for 50 we also re-render the context views.
Scores exclude the sky: \emph{Full} uses all remaining pixels, \emph{Traffic} only those of transient objects in the real frame, and \emph{Static} the rest.

\subsection{Reconstruction on Real Drives}
\label{sec:mainresults}
\begin{table}[t]
  \centering
  \caption{\textbf{Held-out traversals}, static pixels, odd frames. Lower block: baselines relit by our branch; only \method runs color steps. Gaussians: ReSplat 100k, \method 200k, rest 1.6M; \emph{dense}: no refiner.}
  \label{tab:main}
  \footnotesize
  \setlength{\tabcolsep}{2pt}
  \resizebox{\linewidth}{!}{%
  \begin{tabular}{@{}lccccccr@{}}
    \toprule
    & \multicolumn{3}{c}{MTGS (18)} & \multicolumn{3}{c}{Ithaca365 (30)} & \\
    \cmidrule(lr){2-4}\cmidrule(lr){5-7}
    & PSNR & SSIM & LPIPS & PSNR & SSIM & LPIPS & Time \\
    \midrule
    \multicolumn{8}{@{}l}{\emph{No image of the target drive}} \\
    Wild3R & 13.56 & .458 & .423 & 16.00 & .363 & .413 & 0.27\,s \\
    ReSplat & 19.38 & .623 & .329 & \textbf{19.04} & \textbf{.477} & .333 & 0.38\,s \\
    MVSplat & 17.20 & .557 & .333 & 18.12 & .434 & .334 & 0.98\,s \\
    DepthSplat & 18.40 & .582 & .285 & 18.53 & .447 & .268 & 0.27\,s \\
    MTGS & 19.30 & .586 & .264 & -- & -- & -- & 38\,min \\
    \method, dense & 19.18 & .623 & \textbf{.227} & 18.35 & .427 & .255 & 0.28\,s \\
    \method & \textbf{19.88} & \textbf{.626} & .235 & 18.65 & .419 & \textbf{.249} & 0.80\,s \\
    \midrule
    \multicolumn{8}{@{}l}{\emph{Target-drive references; baselines relit by our branch}} \\
    ReSplat & 19.91 & .625 & .328 & 19.78 & \textbf{.484} & .331 & 0.41\,s \\
    MVSplat & 17.62 & .561 & .329 & 18.69 & .447 & .326 & 1.02\,s \\
    DepthSplat & 18.92 & .585 & .283 & 19.19 & .456 & .261 & 0.31\,s \\
    \method & \textbf{21.26} & \textbf{.649} & \textbf{.220} & \textbf{19.90} & .456 & \textbf{.218} & 1.15\,s \\
    \bottomrule
  \end{tabular}}
\end{table}

\begin{figure}[t]
  \centering
  \includegraphics[width=\linewidth]{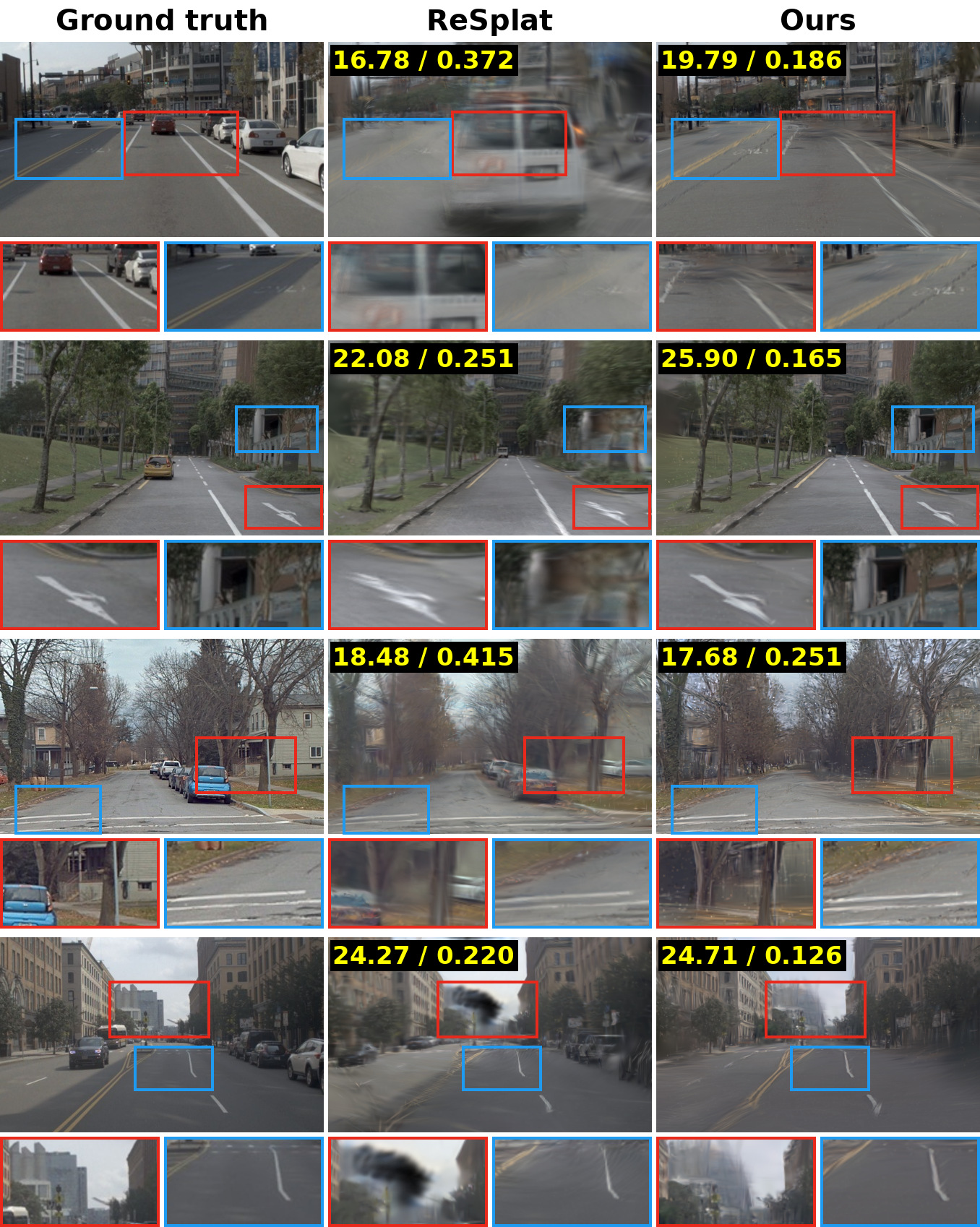}
  \caption{\textbf{Held-out traversals.} Ground truth, ReSplat, and \method (200k Gaussians, lighting references), each with two zoomed crops. Context: same-direction drives (rows 1, 3), one other drive (row 2), opposing-direction drives (row 4). Static PSNR/LPIPS of the full frame; frames hand-picked.}
  \label{fig:qualitative}
\end{figure}

Without any image of the target drive (Table~\ref{tab:main}, upper block), \method has the lowest LPIPS on both datasets and the highest PSNR on MTGS; on Ithaca365 it is 0.4\,dB behind ReSplat, whose smoother renders also score a higher SSIM.
Per-scene optimization with MTGS needs about 38\,min on the same eight images and is still outperformed by \method on all three metrics.
Given target-drive references, every feedforward method receives the same lighting code from our branch, which adds only 0.03--0.04\,s to a baseline.
\method then leads in PSNR, clearly on MTGS (1.35\,dB over relit ReSplat) and narrowly on Ithaca365 (0.12\,dB), and its LPIPS is 0.063 and 0.043 below relit DepthSplat.
In Figure~\ref{fig:qualitative}, ReSplat blurs detail and renders vehicles of the context drives, which real data cannot score (Sec.~\ref{sec:transienteval}); \method leaves streaks on the road.
Over all scored frames, \method beats relit ReSplat on both metrics on 79\,\% of the MTGS and 54\,\% of the Ithaca365 frames; on nearly all others it has a lower PSNR but a much better LPIPS.
Since the lighting branch changes only colors, a reference mainly moves PSNR: it raises ours by 1.4\,dB on MTGS and 1.3\,dB on Ithaca365 but lowers LPIPS by only 0.015 and 0.031 (Table~\ref{tab:abl_light}).
Our LPIPS lead over ReSplat, 0.08--0.11 with and without references on both datasets, therefore reflects a sharper reconstruction rather than better colors (Fig.~\ref{fig:qualitative}).

\subsection{Results on \bench}
\label{sec:transienteval}
\begin{table}[t]
  \centering
  \caption{\textbf{Transient removal against traffic-free ground truth} (PSNR; LPIPS on Full). \emph{Input}: re-rendered context views; \emph{held-out}: views without context. \emph{Ref.}: lighting references; \emph{Mask}: our predicted transient masks. \emph{Relit}: baselines given the lighting code of our branch from the same references; only \method runs color steps. \emph{Real frame}: unmodified. \emph{Dense}: 1.6M Gaussians, no refiner; other rows of \method: 200k. Full PSNR is reported on held-out views.}
  \label{tab:carla}
  \footnotesize
  \setlength{\tabcolsep}{2pt}
  \begin{tabular}{@{}lcc@{\hspace{5pt}}cc@{\hspace{5pt}}ccc@{}}
    \toprule
    & & & \multicolumn{2}{c}{Input (50)} & \multicolumn{3}{c}{Held-out (173)} \\
    \cmidrule(r){4-5}\cmidrule(l){6-8}
    & Ref. & Mask & Traffic & Static & Traffic & Full & LPIPS \\
    \midrule
    Real frame & & & 10.47 & 30.96 & 10.44 & -- & -- \\
    Wild3R & & & 16.48 & 15.64 & 13.88 & 12.67 & .489 \\
    ReSplat & & & 14.39 & 23.96 & 14.08 & 14.00 & .455 \\
    MVSplat & & & 13.28 & 18.85 & 13.15 & 13.76 & .455 \\
    DepthSplat & & & 12.89 & \textbf{24.78} & 13.22 & 13.61 & .392 \\
    ReSplat, relit & \checkmark & & 14.43 & 23.34 & 16.03 & 15.39 & .447 \\
    MVSplat, relit & \checkmark & & 13.27 & 18.60 & 14.88 & 15.01 & .444 \\
    DepthSplat, relit & \checkmark & & 12.87 & 24.25 & 14.93 & 14.90 & .383 \\
    \midrule
    \method, dense & & & 18.24 & 21.68 & 15.29 & 14.51 & .351 \\
    \method & & & 18.67 & 22.01 & 15.25 & 14.58 & .370 \\
    \method & & \checkmark & 20.34 & 22.10 & 15.58 & 14.64 & .365 \\
    \method & \checkmark & & 18.82 & 21.86 & 18.63 & 17.13 & .330 \\
    \method & \checkmark & \checkmark & \textbf{20.38} & 21.94 & \textbf{19.05} & \textbf{17.17} & \textbf{.325} \\
    \bottomrule
  \end{tabular}
\end{table}

\begin{table}[t]
  \centering
  \caption{\textbf{Where the traffic removal comes from} (PSNR on \bench, 200k Gaussians, no lighting reference). \emph{Mask}: transient labels given to the sampler and, with the refiner, to its residuals; same weights in every row. Without a refiner the rows show the compact head alone. Traffic covers only about 5\,\% of a frame, so Full PSNR changes little.}
  \label{tab:abl_masks}
  \footnotesize
  \setlength{\tabcolsep}{3pt}
  \begin{tabular}{@{}llcccc@{}}
    \toprule
    & & \multicolumn{2}{c}{Input (50)} & \multicolumn{2}{c}{Held-out (173)} \\
    \cmidrule(r){3-4}\cmidrule(l){5-6}
    Mask & Refiner & Traffic & Static & Traffic & Full \\
    \midrule
    \multicolumn{2}{@{}l}{Dense, 1.6M} & 18.24 & 21.68 & 15.29 & 14.51 \\
    \midrule
    None & & 18.93 & 20.99 & 15.40 & 14.47 \\
    None & \checkmark & 18.67 & 22.01 & 15.25 & 14.58 \\
    Predicted & & 19.51 & 21.04 & 15.53 & 14.51 \\
    Predicted & \checkmark & \textbf{20.34} & 22.10 & \textbf{15.58} & 14.64 \\
    Ground truth & & 19.59 & 21.05 & 15.54 & 14.51 \\
    Ground truth & \checkmark & 20.36 & \textbf{22.14} & 15.56 & \textbf{14.65} \\
    \bottomrule
  \end{tabular}
\end{table}

\begin{table*}[t]
  \footnotesize
  \setlength{\tabcolsep}{3pt}
  \def\abltabh{66pt}%
  \begin{minipage}[b]{0.31\textwidth}
    \centering
    \caption{\textbf{Refiner} (200k, no reference). \emph{Frozen}: ReSplat's released weights.}
    \label{tab:abl_refiner}
    \resizebox*{!}{\abltabh}{%
    \begin{tabular}{@{}lcccc@{}}
      \toprule
      & \multicolumn{2}{c}{MTGS} & \multicolumn{2}{c}{Ithaca365} \\
      \cmidrule(lr){2-3}\cmidrule(l){4-5}
      Refiner & PSNR & LPIPS & PSNR & LPIPS \\
      \midrule
      None & 19.13 & .262 & 18.32 & .277 \\
      Frozen, 2 steps & 19.40 & .294 & 18.48 & .320 \\
      Frozen, 4 steps & 19.17 & .313 & 18.54 & .337 \\
      Fine-tuned, 2 steps & \textbf{19.88} & \textbf{.235} & \textbf{18.65} & \textbf{.249} \\
      \bottomrule
    \end{tabular}}
  \end{minipage}\hfill
  \begin{minipage}[b]{0.30\textwidth}
    \centering
    \caption{\textbf{Filling the holes of sampling} (MTGS). All rows keep the same pixels.}
    \label{tab:abl_compact}
    \resizebox*{!}{\abltabh}{%
    \begin{tabular}{@{}lccc@{}}
      \toprule
      100k, no refiner & PSNR & SSIM & LPIPS \\
      \midrule
      Discard the rest & 10.51 & .304 & .530 \\
      Merge the rest & 18.75 & \textbf{.636} & .359 \\
      Our head & \textbf{18.88} & .616 & \textbf{.274} \\
      \textcolor{gray}{Dense, 1.6M} & \textcolor{gray}{19.18} & \textcolor{gray}{.623} & \textcolor{gray}{.227} \\
      \bottomrule
    \end{tabular}}
  \end{minipage}\hfill
  \begin{minipage}[b]{0.33\textwidth}
    \centering
    \caption{\textbf{Lighting branch} (200k). The geometry is identical in all three rows.}
    \label{tab:abl_light}
    \resizebox*{!}{\abltabh}{%
    \begin{tabular}{@{}lcccc@{}}
      \toprule
      & \multicolumn{2}{c}{MTGS} & \multicolumn{2}{c}{Ithaca365} \\
      \cmidrule(lr){2-3}\cmidrule(l){4-5}
      Lighting & PSNR & LPIPS & PSNR & LPIPS \\
      \midrule
      No reference & 19.88 & .235 & 18.65 & .249 \\
      + lighting code & 20.71 & .231 & 19.39 & .238 \\
      + color steps & \textbf{21.26} & \textbf{.220} & \textbf{19.90} & \textbf{.218} \\
      \bottomrule
    \end{tabular}}
  \end{minipage}
\end{table*}

\begin{figure}[t]
  \centering
  \includegraphics[width=\linewidth]{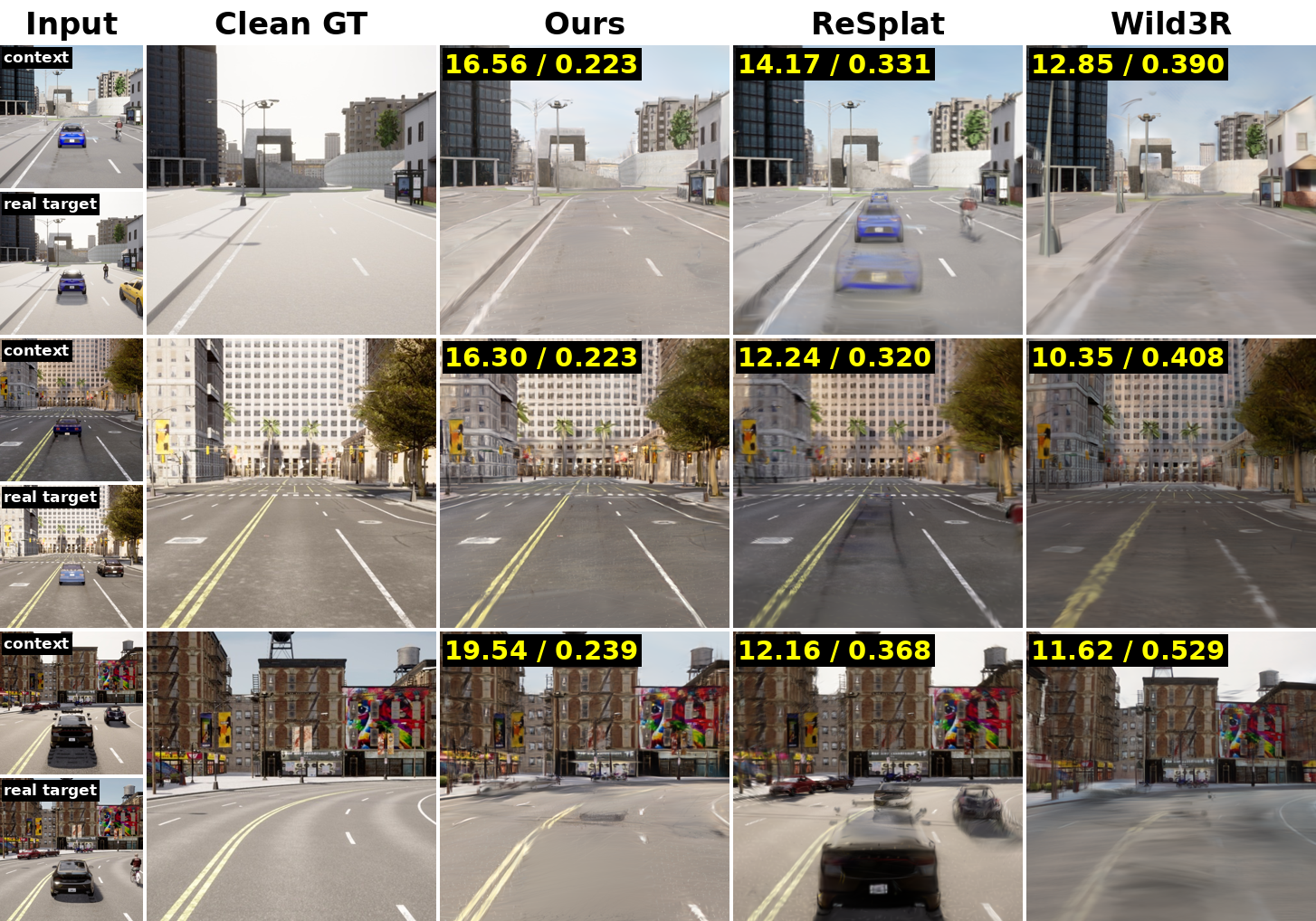}
  \caption{\textbf{Baked-in transients on \bench.} Left: nearest context view above the real target frame. Context from one drive. ReSplat renders context vehicles as ghosts; \method keeps the road clear, sometimes leaving a blurred patch. Wild3R leaves smeared vehicle artifacts. Full PSNR/LPIPS against the clean frame.}
  \label{fig:carla}
\end{figure}

\paragraph{Removing traffic from the input.}
We first ask whether a method can remove a vehicle it sees, by re-rendering its input views (Table~\ref{tab:carla}).
The clean copy contains no vehicles, pedestrians, or cyclists at all, so Traffic PSNR shows directly how well transients are removed: a method only scores well on these pixels if it renders the empty road.
As a reference point, returning the unmodified input yields 10.5\,dB.
ReSplat and DepthSplat reach 24--25\,dB on static pixels but only 14.4 and 12.9\,dB on traffic, because they largely reproduce what they see.
Wild3R, which was designed to suppress transients, is the best baseline at 16.5\,dB, but it has the lowest static PSNR of all methods.
\method reaches 20.4\,dB, 3.9\,dB above the best baseline with and without a lighting reference, and it wins on 48 of the 50 splits, while also improving on Wild3R by 6.3\,dB on static pixels.
The gain is geometric: a reference changes the Traffic score by only 0.04\,dB, and relighting the baselines with our branch changes theirs by at most as much.
ReSplat and DepthSplat copy their input, which scores well on static pixels but renders the vehicle; \method trades 2.0--2.8\,dB of static PSNR for the empty road.

\paragraph{A clean scene from a new drive.}
On held-out frames the target's own traffic was never in the input, so Traffic PSNR measures how well the background is recovered, and Full PSNR also counts ghosts anywhere in the view, such as the context vehicles that ReSplat bakes in (Fig.~\ref{fig:carla}).
\method is ahead at every difficulty level, and in Full PSNR it beats the best relit baseline on 159 of the 173 splits.
Even without a lighting reference, it is 1.5\,dB above the best baseline in Traffic PSNR.
A reference from the target drive adds another 3.5\,dB by correcting the colors; the geometry stays the same.
Relit with the same references, the baselines gain 1.7--2.0\,dB of Traffic PSNR, but \method still leads relit ReSplat by 3.0\,dB in Traffic and 1.8\,dB in Full PSNR, at every difficulty level (Suppl.\ Table~\ref{tab:carla_difficulty}), and its LPIPS is 0.058 below relit DepthSplat.

\paragraph{Where the removal comes from.}
Table~\ref{tab:abl_masks} separates the stages without a lighting reference, so only geometry changes.
The fine-tuned backbone does most of the removal: its dense cloud reaches 18.2\,dB in Traffic PSNR on the input views, 1.8\,dB above the best baseline, and the compact head, trained on targets with inpainted vehicles, removes more even without masks.
The predicted masks then matter at both later stages.
In the sampler, they move the budget from vehicles to the static scene (+0.6\,dB for the head alone, 49 of 50 splits).
In the refiner, they decide what the residuals ask for: with masks, refinement adds 0.8\,dB of Traffic PSNR; without them, it fits the vehicles the context views still show and loses 0.3\,dB, while improving static pixels by 1.0\,dB in both cases.
On held-out views the effects are smaller but have the same sign.
Removing the masks costs 1.6--1.7\,dB on the input views and 0.3--0.4\,dB on held-out views, and vehicles can return as ghosts (Suppl.\ Fig.~\ref{fig:nomask}).
Predicted masks perform like ground-truth masks (within 0.02\,dB here, indistinguishable on real data), so no external transient-mask source is needed.

\subsection{Ablations}
\label{sec:backboneeval}
\label{sec:steps}
\label{sec:samplingeval}
\label{sec:appearanceeval}

\paragraph{Refiner.}
Applied frozen to our Gaussians, ReSplat's refiner raises PSNR by at most 0.3\,dB and worsens LPIPS; fine-tuned, it improves both (Table~\ref{tab:abl_refiner}).
Two recurrent steps give the best LPIPS, even against a refiner trained for four.

\paragraph{Compact Gaussians.}
Keeping a sixteenth of the pixels and discarding the rest severely degrades the rendering, because the dense Gaussians are the size of a pixel (Table~\ref{tab:abl_compact}).
Our head, trained with the sampler in the loop, has a much better LPIPS than merging the dropped Gaussians into their neighbors, at equal PSNR.
One head and one refiner serve every budget with PSNR within 0.25\,dB from 100k to 400k, while LPIPS improves with larger budgets (Suppl.\ Fig.~\ref{fig:budget}).

\paragraph{Lighting.}
The lighting code alone contributes 0.7--0.8\,dB, and the color steps, which need our refiner, add another 0.5\,dB (Table~\ref{tab:abl_light}).
The code also carries over to other methods: applied without retraining to ReSplat, MVSplat, and DepthSplat, it adds 0.4--0.5\,dB on MTGS and 0.6--0.7\,dB on Ithaca365 (Table~\ref{tab:main}).

\paragraph{Cameras and context views.}
Giving Wild3R known intrinsics in its unprojection adds about 1.5\,dB; its row in Table~\ref{tab:main} uses its own predicted cameras, aligned to the ground truth by one similarity transform.
More context views help without retraining: going from 8 to 16 adds 0.3\,dB on MTGS and 0.1\,dB on Ithaca365.
\section{Conclusion}
\label{sec:conclusion}
\label{sec:limitations}
We presented \method, a feedforward framework for reconstructing shared static scenes from multi-traversal driving images.
Explicit transient masks guide compact Gaussian prediction with entropy-aware sampling priors, significantly reducing redundant Gaussians.
A two-stage recurrent refiner updates geometry and appearance, allowing target-drive references to improve colors without changing the reconstructed geometry.
Finally, \bench, with its traffic-free ground truth, provides a benchmark for evaluating both the performance of reconstruction and transient removal.
Experiments on MTGS and Ithaca365 demonstrate improved perceptual quality and, with target-drive references, the highest PSNR among the baselines, with fine-grained reconstruction within two seconds.

\paragraph{Limitations.}
Only \method is trained on driving data, and we use the ground-truth poses for calibrated cameras.
Transient removal can only be measured directly in simulation, and background hidden in every context view is smeared rather than recovered.
The lighting branch transfers color only: it cannot move shadows or add light sources, and it works best with reference frames of the target drive.
\clearpage
\section*{LLM Usage Statement}
\label{sec:llm}
We used a large language model coding assistant (Claude, Anthropic) throughout this project.
In the research, it wrote and modified code for data preparation, training, and evaluation, launched and monitored training and evaluation runs on our servers, and aggregated results into tables and comparison panels.
In the writing, it drafted, revised, and reviewed parts of the manuscript text, tables, captions, and the architecture figure.
The authors designed the method and the experiments, decided which results to report, and checked the reported numbers against the evaluation outputs.
All scientific claims, analyses, and conclusions are the authors' own, and the authors take full responsibility for the content of this paper.
Separately from writing assistance, the method itself uses generative models as components: LaMa~\cite{suvorov2022lama} inpaints the road behind vehicles in training targets, and generative image editing produces part of the transient-mask training data (Sec.~\ref{sec:transients}).

  \clearpage
  \setcounter{page}{1}
\maketitlesupplementary
\raggedbottom
\renewcommand{\topfraction}{0.9}\renewcommand{\dbltopfraction}{0.9}
\renewcommand{\bottomfraction}{0.6}\renewcommand{\textfraction}{0.07}
\renewcommand{\floatpagefraction}{0.75}\renewcommand{\dblfloatpagefraction}{0.75}
\setcounter{topnumber}{3}\setcounter{dbltopnumber}{2}\setcounter{bottomnumber}{2}\setcounter{totalnumber}{5}
\setcounter{section}{0}
\renewcommand{\thesection}{\Alph{section}}
\renewcommand{\theHsection}{supp.\Alph{section}}
\renewcommand{\theHsubsection}{\theHsection.\arabic{subsection}}

\etocdepthtag.toc{supplement}
\begingroup
  \renewcommand{\contentsname}{Supplementary Contents}
  \etocsettagdepth{mainpaper}{none}
  \etocsettagdepth{supplement}{subsection}
  \tableofcontents
\endgroup

\section{Implementation and Training Details}
\label{sec:supp_impl}
This section lists what is needed to reproduce \method: preprocessing, every module, the five training stages in one table (Table~\ref{tab:supp_recipe}), and the default inference configuration (Table~\ref{tab:supp_inference}); results that deviate from it are listed in Table~\ref{tab:supp_variants}.
\subsection{Calibration and LiDAR Processing}
\label{sec:supp_calib}
\paragraph{Images and intrinsics.}
The driving datasets provide pinhole intrinsics $K$ and OpenCV distortion coefficients for images that are still distorted.
We undistort every image to the same $K$, so the intrinsics stay valid.
Ithaca365 frames are distributed without rectification, and we use their intrinsics, rotation, and distortion coefficients; an epipolar check with SIFT matches and the RTK poses confirmed that this is the correct camera model.
Each image is then resized so that its short side is 448 pixels and center-cropped to $448\times448$.
The focal lengths and the principal point are scaled by the resize factors, and the principal point is shifted by the crop offset.
Labels and masks go through the same rectification and crop.

\paragraph{World frame and scale.}
Poses are given in a frame local to each road block.
For a split we express them relative to the first context view.
The backbone predicts depth up to a global scale, which we fit from the distances between camera pairs, using the cameras of the camera head and the ground-truth cameras.
The scale is detached, it divides the ground-truth translations, and depth is unprojected with these cameras and the ground-truth intrinsics (Sec.~\ref{sec:initialization}).

\paragraph{LiDAR.}
LiDAR is used only to train the backbone and to evaluate depth.
nuPlan sweeps are given in the ego frame of the sweep and are mapped to the world with that sweep's ego pose; each image uses the sweep nearest in time, at most 25\,ms away.
Ithaca365 sweeps are mapped with the sensor-to-world transform of the sweep with the frame's timestamp.
Points are then transformed into the camera, projected with the rectified $K$, and rounded to the nearest pixel.
Depth is the camera-axis distance $z$, restricted to at most 80\,m.
Where several returns fall on one pixel we keep the nearest; there is no further occlusion test, so a return that the camera cannot see stays valid if no nearer return lands on its pixel.
With the validity indicator $V_i$ of the projected returns, the backbone is supervised with
\begin{equation}
  \mathcal{L}_{\mathrm{depth}}=\frac{1}{|\views|}\sum_{i}
  \frac{\sum_{\mathbf{u}}V_i(\mathbf{u})\,
  \bigl|\log Z_i(\mathbf{u})-\log\bigl(s\,D_i(\mathbf{u})\bigr)\bigr|}
  {\sum_{\mathbf{u}}V_i(\mathbf{u})+\epsilon},
  \label{eq:depthloss}
\end{equation}
where $s$ is the detached scale at which the cloud is placed, so the loss ties depth to the scale used for rendering; the term has a constant weight of 0.5.
Returns on transient objects are kept in this loss.
Depth supervision complements the image losses, since a reconstruction can improve its renders without improving its metric geometry.

\subsection{Network Modules}
\label{sec:supp_modules}
\paragraph{Backbone.}
The transformer receives images only and has no pose conditioning.
With the same ground-truth cameras and 1.6M Gaussians, adapting Wild3R to driving data raises the dense renders from 17.57 to 19.18\,dB PSNR on MTGS and from 17.98 to 18.35\,dB on Ithaca365, and lowers LPIPS from 0.313 to 0.227 and from 0.341 to 0.255.
Images, intrinsics, masks, and projected depth share the same rectified camera model after resizing and cropping; replacing the cameras only at rendering time would not correct a cloud lifted with inconsistent geometry.
Low-rank adapters sit on the attention projections (query/key/value and output) and the second MLP layer of all 48 alternating-attention blocks~\cite{wang2025vggt}.
Each low-rank adapter adds $B(A(x))$ to the output of its linear layer, with $A$ and $B$ of rank 16, a scale of 1, and $B$ initialized to zero, so training starts from the pretrained model.

\paragraph{Transient head.}
The head reads the 2048-dimensional tokens of layers 4, 11, 17, and 23 of the transformer.
They are projected to 256, 512, 1024, and 1024 channels, brought to four spatial scales, and fused by a DPT decoder with 256 features; the output is upsampled to full resolution and gives one logit per pixel.
It predicts transients only; sky and the ego hood come from semantic labels (Sec.~\ref{sec:supp_transient_data}).

\paragraph{Compact Gaussian head.}
The image path of the Wild3R head is widened from three to five input channels for the keep probability and the local density of Sec.~\ref{sec:compact}.
Pixels are sampled as in SparseSplat~\cite{zhang2026sparsesplat}.
For every context image we compute the Shannon entropy $E$ of the grayscale histogram (256 levels) in a $7{\times}7$ window, multiply it by a class weight $\lambda$ that is $0.1$ for sky and transient pixels and $1$ otherwise, and keep pixel $\mathbf{u}$ with probability
\begin{equation}
  P(\mathbf{u})=\operatorname{clip}\!\Bigl(\tau\,\frac{\lambda(\mathbf{u})\,E(\mathbf{u})}{\log 256},\,0,\,1\Bigr),
  \label{eq:sampling}
\end{equation}
drawing one independent Bernoulli variable per pixel.
The scalar $\tau$ is fitted jointly over all context views so that the expected number of kept pixels equals the budget $B$.
Sky and transient pixels, which would otherwise take about 10\,\% and 6\,\% of a uniform budget, receive little.
The sky class in $\lambda$ comes from the semantic labels of the context views (shipped with MTGS, SegFormer-B2 pseudo-labels on Ithaca365, the simulator's labels on \bench), not from a network of ours.
Each kept pixel receives three scales limited to at most 15, a unit quaternion, an opacity, and spherical harmonics of degree 3 per color channel, 59 values including its center.

\paragraph{Condition adapter.}
The adapter upsamples the backbone tokens ($32\times32$, 2048 channels) bilinearly to the $112\times112$ grid of the head features (128 channels).
Both are projected to 512 channels by $1{\times}1$ convolutions and added.
Two residual blocks (group normalization, GELU, $3{\times}3$ convolutions, last convolution initialized to zero) and a final $1{\times}1$ convolution produce the 512-channel maps.
It was trained once, on the features of the Wild3R backbone, and the refiner was trained with it.
In the final pipeline its weights are kept, and its inputs, the compact head's features, are mapped per channel to the statistics of the features it was trained on, with fitted constants.

\paragraph{Feedback.}
A frozen ResNet-18 encodes the rendering and the image at a quarter of the image resolution, and the feedback is the difference of the two feature maps plus a projection of the pixel-unshuffled color residual.
One attention block then runs across the tokens of all views.
Average-pooled transient masks multiply the feedback, so transient pixels send no signal to the update.
The feedback is read for every Gaussian with the lookup of Eq.~\ref{eq:gather}, giving a 256-dimensional vector (Fig.~\ref{fig:recurrent}).

\paragraph{Recurrent update.}
The hidden state has 512 dimensions and is initialized by a $1{\times}1$ convolution of the adapter's maps, read with the same lookup.
Each Gaussian concatenates its 59 parameters, the state, and the feedback, and a linear layer maps them to 512 dimensions.
Four blocks of a point transformer then exchange information between the eight nearest neighbors of each Gaussian in 3D, and their output is added to the state.
A four-layer MLP, with its last layer initialized to zero, predicts the 59 parameter changes.
The neighbors are recomputed in the geometry steps; color steps do not move the Gaussians, so they reuse them.

\paragraph{Constrained updates.}
Centers, scales, and the unnormalized quaternion are updated additively, opacity is updated in logit space, and scales are clamped to stay positive; the quaternion is normalized afterwards.
Geometry steps apply only the changes to centers, scales, rotations, and opacities, and color steps apply only those to the spherical harmonics.
In a color step, a Gaussian seen by at least one lighting reference takes its feedback from the references, rendered in the target lighting; otherwise it keeps the feedback of the context views.
A linear layer initialized to zero adds a projection of the two 32-dimensional lighting codes and the Gaussian's color transform $(\boldsymbol\gamma_j,\mathbf{b}_j)$ to that feedback.

\paragraph{Appearance branch.}
A frozen DINOv2-S/14 encoder~\cite{oquab2024dinov2}, separate from the geometry backbone, embeds an image $I$.
An MLP $E_\psi$ maps its class token, mean patch token, and global color statistics (per-channel mean and standard deviation, five luminance quantiles, and a $4{\times}4$ grid of mean colors) to a 32-d lighting code $\mathbf{e}=E_\psi(I)$.
The 16-d local feature $\mathbf{f}_j$ is a linear projection of the DINOv2 patch feature at the Gaussian's source pixel $\mathbf{u}_j$.
The last layer of $\Gamma_\psi$ is zero-initialized, and an identity loss makes $\mathbf{e}_{\mathrm{tgt}}=\mathbf{e}_{\mathrm{ctx}}$ return $\mathbf{c}_j$ unchanged.
For spherical-harmonic colors $\mathbf{c}=Y_0\,\mathbf{k}_0+\tfrac12+\sum_{l>0}(\cdot)$, every band is scaled by $\boldsymbol\gamma_j$ and the DC coefficient absorbs the offset, $\mathbf{k}'_0=\boldsymbol\gamma_j\odot\mathbf{k}_0+\bigl(\mathbf{b}_j+\tfrac12(\boldsymbol\gamma_j-1)\bigr)/Y_0$, which applies Eq.~\ref{eq:relight} exactly.
In the refiner, the color steps read the reference residuals with the depth-tested projection of Eq.~\ref{eq:gather}, and a 70-d lighting vector $[\mathbf{e}_{\mathrm{ctx}},\mathbf{e}_{\mathrm{tgt}},\boldsymbol\gamma_j,\mathbf{b}_j]$ enters the update through a zero-initialized linear projection.

\paragraph{Model size.}
Counted from the checkpoints: the Wild3R model has 1{,}191M parameters (transformer 909M, camera head 216M, depth and Gaussian heads 33M each), of which we train 8.65M low-rank adapter parameters and the 65.45M of the depth and Gaussian heads.
The camera head only sets the metric scale (Sec.~\ref{sec:supp_calib}).
The transient head has 32.7M parameters, the compact Gaussian head 32.8M, the condition adapter 10.8M, and the refiner's update modules 14.6M (including the 0.7M of the truncated ResNet-18 feature encoder).
$E_\psi$ and $\Gamma_\psi$ have 0.32M; the DINOv2-S encoder they read (22M) stays frozen.

\subsection{Training Stages}
\label{sec:supp_losses}
\label{sec:supp_appearance}
Training proceeds in five stages, each with all earlier modules frozen (Sec.~\ref{sec:training}); Table~\ref{tab:supp_recipe} gives data, losses, initialization, optimization, steps, and time for each.
All stages use AdamW with gradient clipping at 1 on one NVIDIA RTX PRO 6000 Blackwell.
We first define the two losses shared by the rendering stages.

\begin{figure}[ht]
  \centering
  \includegraphics[width=\linewidth]{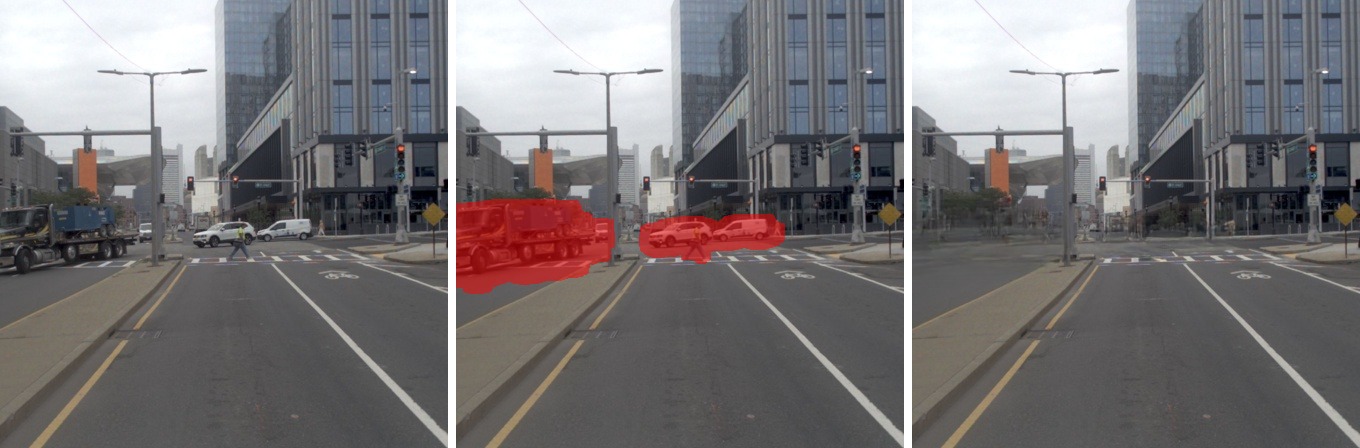}
  \caption{\textbf{Inpainted training target} (nuPlan training block). Left: target frame. Middle: region filled by LaMa (red). Right: the LaMa fill that supervises these pixels.}
  \label{fig:supp_lama}
\end{figure}

\paragraph{Photometric loss.}
Stages (1), (3), (4)(iii), and (5) render held-out target views $q$ of another traversal from the predicted Gaussians with ground-truth cameras and compare them to a target image $I^\ast_q$.
Masked pixels are handled by compositing, not by reweighting, so that LPIPS still sees a complete image:
\begin{equation}
  \begin{aligned}
    \widetilde I_q&=\Omega_q\odot\widehat I_q+(1-\Omega_q)\odot I^\ast_q,\\
    \mathcal{L}_{\mathrm{img}}&=\lambda_{2}\,\bigl\|\widetilde I_q-I^\ast_q\bigr\|_2^2+\lambda_{\mathrm{p}}\,\mathrm{LPIPS}\bigl(\widetilde I_q,I^\ast_q\bigr),
  \end{aligned}
  \label{eq:imageloss}
\end{equation}
where the squared error is averaged over all pixels and LPIPS uses the AlexNet backbone.
Pixels outside the supervised set $\Omega_q$ are copied from the target, so they contribute no error and no gradient.
In stages (1), (3), and (4)(iii), the transient pixels of the target are replaced by a LaMa~\cite{suvorov2022lama} inpainting of the empty road and are part of $\Omega_q$; only the ego vehicle is excluded.
The inpainted region is the transient mask of the target frame, extended downward by a quarter of each object's height to cover its shadow and then dilated (Fig.~\ref{fig:supp_lama}).
In stage (5), $I^\ast_q$ is the real frame and $\Omega_q$ is the static weight $W_q$ of Eq.~\ref{eq:weights}, so transient pixels are excluded.
Colors are compared as rendered; the target's global color is not factored out, except on the no-reference steps of stage (4)(iii).

\paragraph{Coverage loss.}
All four rendering stages add
\begin{equation}
  \mathcal{L}_{\mathrm{cov}}=\frac{1}{|\mathcal{C}_q|}\sum_{\mathbf{u}\in\mathcal{C}_q}\bigl(1-\widehat\alpha_q(\mathbf{u})\bigr),
  \label{eq:covloss}
\end{equation}
where $\widehat\alpha_q$ is the accumulated opacity and $\mathcal{C}_q$ holds all non-sky, non-ego pixels of the target, transient pixels included.

\begin{table*}[t]
  \centering\footnotesize
  \caption{\textbf{Training stages of \method.} All stages: AdamW, gradient clipping at 1, one RTX PRO 6000 Blackwell. Time is wall-clock for the final run of each stage; for (5) it excludes the 40k warm-start steps. The adapter's run did not log its duration.}
  \label{tab:supp_recipe}
  \label{tab:supp_appearance}
  \setlength{\tabcolsep}{3pt}
  \begin{tabular}{@{}>{\raggedright\arraybackslash}p{0.10\textwidth}>{\raggedright\arraybackslash}p{0.16\textwidth}>{\raggedright\arraybackslash}p{0.22\textwidth}>{\raggedright\arraybackslash}p{0.21\textwidth}>{\raggedright\arraybackslash}p{0.16\textwidth}>{\raggedright\arraybackslash}p{0.08\textwidth}@{}}
    \toprule
    Stage & Trained / initialization & Data & Losses (weights) & Optimization & Steps, time \\
    \midrule
    (1) Backbone & Rank-16 LoRA on the transformer, depth and Gaussian heads; from released Wild3R & 1{,}653 splits (1{,}383 distinct), 274 nuPlan and Ithaca365 blocks (IT00/01/02/04), $\sim$60\,\% multi-traversal; 3 targets/step & 0.8\,MSE + 0.2\,LPIPS on LaMa targets, 1.0 coverage, 0.5 log-depth LiDAR (Eq.~\ref{eq:depthloss}) & lr $10^{-4}$ LoRA, $2\cdot10^{-5}$ heads; 200-step warm-up, cosine to 1/10; wd 0.01 & 4k; 1.7\,h \\
    (2) Transient head & New DPT head on frozen backbone tokens & $\sim$700 WildCity frames with inserted traffic, $\sim$1{,}200 nuPlan frames with SegFormer-B2 labels & Binary cross-entropy + Dice (1:1), context views & lr $2\cdot10^{-4}$, cosine to 0; wd $10^{-4}$ & 40 epochs (best 23); $<$2\,h \\
    (3) Compact head & Gaussian head of (1) + 2 zero-initialized input channels & Splits of (1); 2 targets/step; budget drawn from \{100k, 200k, 400k\} & 0.8\,MSE + 0.2\,LPIPS on LaMa targets, 1.0 coverage & lr $2\cdot10^{-5}$, cosine to 1/10; wd 0.01 & 4k (best 1{,}750); 2.0\,h \\
    (4)(i) Image-space transfer & $E_\psi$, $\Gamma_\psi$ from scratch & WildCity relit pairs, 93 scenes (6 held out); batch 8 & MSE + 0.1\,LPIPS (VGG) + 1.0 identity; exposure, white-balance, gamma jitter & lr $3\cdot10^{-4}$ & 30k; 1.6\,h \\
    (4)(ii) Real-domain alignment & Continues (i) & Driving frames (10k steps), then + Photo Tourism pairs with $\geq$100 shared COLMAP points (10k); batch 8 & Identity 1.0, same-lighting 1.0, cycle 0.5, color statistics 0.3; WildCity loss continues & lr $10^{-4}$ & 20k; 1.6\,h \\
    (4)(iii) Gaussian space & Continues (ii); head and backbone frozen & Refiner splits below; 2 target + 2 context views/step; far reference with $p{=}0.75$ & 0.8\,MSE + 0.2\,LPIPS on LaMa targets, 1.0 coverage; context 0.5, identity 1, WildCity 0.5 & lr $3\cdot10^{-5}$, 200-step warm-up, cosine to 1/10 & 12k (best 11.5k); 5.5\,h \\
    Adapter & New, on frozen backbone features & 19 nuPlan blocks, 4 views/step & MSE to per-channel standardized ReSplat condition features & lr $10^{-3}$, one-cycle; wd $10^{-4}$ & 4k; not logged \\
    (5) Refiner & Update modules + lighting projection; warm start (text) & 1{,}001 splits (434 nuPlan, 567 Ithaca365) from 240 blocks; 2 targets/step; budget drawn per step & 1.0\,MSE + 0.5\,LPIPS + 1.0 coverage on the real frame, transients excluded & lr $10^{-5}$, cosine to $10^{-6}$; wd 0 & 8k; 7.0\,h \\
    \bottomrule
  \end{tabular}
\end{table*}

\paragraph{(1) Backbone.}
The renders of three target views per step are supervised with Eqs.~\ref{eq:imageloss}--\ref{eq:covloss} and the LiDAR term of Eq.~\ref{eq:depthloss}, at a peak of 17\,GB.

\paragraph{(2) Transient head.}
\label{sec:supp_transient_train}
We insert vehicles and pedestrians into clean, transient-free WildCity~\cite{furutani2026wild3r} frames with an image-editing model~\cite{wu2025qwenimage}.
The model changes the entire image slightly, so we segment the inserted objects with SAM~3~\cite{carion2025sam3} and paste them onto the original clean frame.
A training frame therefore equals the clean frame outside the mask, and the mask is exact.
The head trains on about 700 of these generated frames and 1{,}200 nuPlan frames with SegFormer-B2~\cite{xie2021segformer} labels of real traffic.
Little supervision is needed: on the original Wild3R features, a head trained on about 700 frames reaches 0.97 AP and 0.86 IoU on six held-out MTGS blocks, and more data did not help (0.85 IoU with about 1{,}170 frames, 0.84 with about 1{,}640), which is consistent with the transformer already separating transient content and the head only making it explicit.
Ithaca365 is absent from the head's training data, but the backbone it reads was fine-tuned on part of it.

\paragraph{(3) Compact Gaussian head.}
Each step samples the budget, runs the sampler of Eq.~\ref{eq:sampling} with the transient labels of the context views, and renders two target views.

\paragraph{(4) Appearance branch.}
The geometry model stays frozen throughout, and training proceeds in three sub-stages.
\emph{(i)~Supervised image-space transfer.}
WildCity~\cite{furutani2026wild3r} renders each synthetic scene under many HDR environment lightings, giving pixel-aligned pairs of the same camera under two lightings.
Every pixel stands in for a Gaussian.
The source and target codes are computed from \emph{other} cameras of the scene, so they can carry lighting but not the target's content.
\emph{(ii)~Real-domain alignment without pairs.}
Real data has no relit pairs, so codes of real driving frames (nuPlan and Ithaca365 training blocks) and of Photo Tourism photographs~\cite{snavely2006phototourism,martinbrualla2021nerfw} are aligned with pair-free losses.
These are identity; same-lighting consistency (a code from another frame of the same drive must leave colors unchanged); cycle consistency (transfer to another drive and back); and color-statistics matching to a second drive in the same direction, or to a co-visible Photo Tourism photo.
\emph{(iii)~Gaussian space.}
Compact Gaussians from the frozen geometry model are relit with Eq.~\ref{eq:relight} and rendered at the target views of the held-out traversal.
The target code comes from a frame of the held-out drive outside the reconstructed window.
On some steps no reference is given: the target code is then that of a context view, and on cross-traversal targets a detached per-image affine color fit removes the unknowable global color from the loss.
Two context views, each rendered with every Gaussian relit to that view's own code, supervise the transfer between traversals inside the context.
Training the Gaussian head jointly in this stage degraded its geometry, so the head stays frozen and only $E_\psi$ and $\Gamma_\psi$ are updated.

\paragraph{Condition adapter.}
The adapter renders nothing, so no mask is involved; it regresses ReSplat's condition features, standardized per channel, from the backbone features (Sec.~\ref{sec:supp_modules}).

\paragraph{(5) Refiner.}
The refiner is unrolled for the full inference schedule ($N{=}2$ geometry steps, then $N{=}2$ color steps with a far frame of the target drive as the lighting reference), and the loss is applied only to the final, relit output; gradients flow back through all unrolled steps.
It is trained with the lighting branch frozen and in the loop and with the geometry-first schedule it uses at inference; an interleaved schedule (geometry, color, geometry, color) lets the references leak into the geometry (Sec.~\ref{sec:supp_light_refiner}).
The run of Table~\ref{tab:supp_recipe} is warm-started from a chain of earlier refiner runs with the same loss on the same compact head, which start from ReSplat's released DL3DV~\cite{ling2024dl3dv} weights: one budget, then all budgets, then with the appearance branch in the loop and an interleaved schedule (the interleaved-trained refiner of Table~\ref{tab:appearance_refiner}); in total the refiner has seen 48k steps.
The four-step refiner of Table~\ref{tab:steps} is initialized from the two-step refiner and trained for 4k steps with the same settings and $N{=}4$ (eight passes), at a peak of about 85\,GB.

\subsection{Inference Configuration}
\label{sec:supp_inference}
Table~\ref{tab:supp_inference} gives the default configuration.
It applies to every result unless the result is listed in Table~\ref{tab:supp_variants}.

\begin{table}[ht]
  \centering\small
  \caption{\textbf{Default inference configuration.} Results that deviate from it are listed in Table~\ref{tab:supp_variants}.}
  \label{tab:supp_inference}
  \setlength{\tabcolsep}{3pt}
  \begin{tabular}{p{0.36\linewidth}p{0.58\linewidth}}
    \toprule
    Setting & Value \\
    \midrule
    Context views, resolution & 8 views, $448\times448$ \\
    Gaussian budgets & 100k, 200k, 300k, 400k \\
    Sampling & Entropy map; sky and transient pixels weighted $\times0.1$ \\
    Maximum Gaussian scale & 15 \\
    Refinement & $N{=}2$: 2 geometry steps, then 2 color steps \\
    Feature lookup & All views that see the Gaussian, 5\,\% depth tolerance \\
    Adapter input & Backbone tokens + compact-head features, normalized per channel \\
    Transient masks & Mask head on backbone tokens (one extra backbone pass for the lighting references only), threshold 0.5 \\
    Sky and ego masks & Semantic labels of the dataset \\
    Lighting references & Even-numbered target frames (9--12 per split); odd frames scored \\
    Rendering & Color normalized by accumulated alpha, with a soft ramp to white around alpha 0.5 \\
    Precision & bf16 for backbone and update transformer \\
    \bottomrule
  \end{tabular}
\end{table}
\subsection{Configurations Behind Each Result}
\label{sec:supp_variants}
Table~\ref{tab:supp_variants} lists every result that deviates from the default of Table~\ref{tab:supp_inference}.
Three earlier versions of the pipeline appear.
\emph{Stock-pass adapter}: the same trained weights as the final pipeline, but the adapter reads a second pass of the unadapted Wild3R model instead of the normalized compact-head features; at 200k with references it scores 21.18\,dB / 0.219 on MTGS and 19.93\,dB / 0.215 on Ithaca365, against 21.26 / 0.220 and 19.90 / 0.218 for the final pipeline.
\emph{Reference-slot pipeline}: a separately trained backbone, head, and refiner that receive a target-drive frame in Wild3R's reference slot instead of using our appearance branch.
\emph{Pre-lighting pipeline}: the final backbone and shared compact head with a refiner trained before the appearance branch existed, run without any target-drive image and scored on all target frames.
\begin{table}[t]
  \centering\footnotesize
  \caption{\textbf{Results that deviate from the default configuration.} The three earlier pipeline versions are defined in Sec.~\ref{sec:supp_variants}.}
  \label{tab:supp_variants}
  \setlength{\tabcolsep}{3pt}
  \begin{tabular}{@{}p{0.30\linewidth}p{0.66\linewidth}@{}}
    \toprule
    Result & Deviation from Table~\ref{tab:supp_inference} \\
    \midrule
    Tables~\ref{tab:steps}, \ref{tab:appearance_code}; upper block of~\ref{tab:appearance_refiner} & One far target-drive frame as reference; all target frames scored \\
    Table~\ref{tab:backbone}, full pipeline & Stock-pass adapter \\
    Tables~\ref{tab:appearance_code}, \ref{tab:appearance_refiner} ($\dagger$); Sec.~\ref{sec:supp_refslot} & Reference-slot pipeline \\
    Sec.~\ref{sec:supp_shared}, single-budget heads & Pre-lighting pipeline \\
    Sec.~\ref{sec:supp_shared}, single-budget refiners & Reference-slot pipeline \\
    Table~\ref{tab:views} & 50\,m windows; 10 Ithaca365 splits (IT03, IT05) \\
    Table~\ref{tab:sampling}, Fig.~\ref{fig:ellipsoids} & 100k, no refiner, no relighting \\
    Table~\ref{tab:settings} & 100k (ReSplat's budget) \\
    \bottomrule
  \end{tabular}
\end{table}

\section{Data and Evaluation Protocol}
\label{sec:supp_protocol}
No test block is used in any training stage, and within a comparison all methods receive the same context images and ground-truth cameras and are scored on the same static pixels.

\subsection{Real Data and Context Construction}
We use the driving scenes distributed with MTGS~\cite{li2025mtgs} and repeated-route data from Ithaca365~\cite{diazruiz2022ithaca365}.
MTGS contributes 18 test splits from six road blocks, which we name A--F (Table~\ref{tab:supp_manifest}); Ithaca365 contributes 30 splits from snow-free drives of five held-out road blocks (IT03, IT05, IT08, IT13, IT19).
Test blocks are geographically separate from the data of every training stage: the nearest training cameras are roughly 200\,m from the held-out Ithaca365 and MTGS test blocks.

Every query lies on a drive without context; the context comes from one other drive (\emph{single held-out}), several drives in the query's direction (\emph{same direction}), or both directions (\emph{opposing direction}).
Every split has eight context views in total; they span about 20\,m of road on MTGS (median 20.7\,m) and about 50\,m on Ithaca365 and \bench.
Within a comparison, methods receive identical context images and preprocessing.
The eight views are a fixed total shared among the context traversals, so multi-traversal splits receive no more images than single-traversal ones; Table~\ref{tab:views} varies the view count.

We call all three settings held-out synthesis rather than geometric extrapolation: the query traversal is held out, but its cameras lie somewhere in the road window that the context covers.
Results are reported per setting as well as in aggregate over all three settings.

\begin{table}[ht]
  \centering\footnotesize
  \caption{\textbf{Real-data test blocks.} Context drives, held-out drive(s), and target frames per split (half are scored). Each MTGS block has one split per setting; each Ithaca365 block has three windows with two settings each, and a window never uses its held-out drive as context.}
  \label{tab:supp_manifest}
  \setlength{\tabcolsep}{3.5pt}
  \begin{tabular}{lllll}
    \toprule
    Name & Block & Context drives & Held-out & Targets \\
    \midrule
    A & MTGS 331220 & 0, 1, 2, 7 & 8 & 21 \\
    B & MTGS 587400 & 0, 23, 31 & 2 & 23 \\
    C & MTGS 587640 & 0, 2, 3, 23 & 41 & 20 \\
    D & MTGS 365000 & 3, 4 & 5 & 17 \\
    E & MTGS 365530 & 1, 2, 5 & 6 & 34 \\
    F & MTGS 587860 & 1, 4, 16, 35 & 33 & 19 \\
    \midrule
    & IT03 & 2, 4, 5, 6 & 7 & 24 \\
    & IT05 & 2, 4, 5, 6, 7 & 2, 5 & 24 \\
    & IT08 & 0--6, 8 & 0, 2 & 24 \\
    & IT13 & 0, 1, 2, 3, 5, 6, 7, 8 & 4 & 24 \\
    & IT19 & 0--6, 8 & 0, 7 & 24 \\
    \bottomrule
  \end{tabular}
\end{table}
Training and test data are separated by road block, so repeated drives share locations only within a block.
No MTGS block is used for training.
The backbone and the compact head see nuPlan blocks and the Ithaca365 blocks IT00, IT01, IT02, and IT04 (Sec.~\ref{sec:initialization}).
The refiner and the appearance branch also see the 13 remaining Ithaca365 blocks; the refiner trains on 1{,}001 splits (434 nuPlan, 567 Ithaca365) from 240 blocks.
The five Ithaca365 test blocks are used in no training stage.

\subsection{Metrics and Support}
\label{sec:supp_metrics}

We report PSNR, SSIM~\cite{wang2004ssim}, and LPIPS~\cite{zhang2018lpips} with the AlexNet backbone.
The metrics reward different things: PSNR is a per-pixel squared error, which favors matching colors and a smooth render over a sharp one that is a few pixels off, while LPIPS compares unit-normalized deep features and responds mainly to structure.
A global offset in exposure or color cast therefore lowers PSNR even when every structure is in place, and moves LPIPS much less.
Three of our results show the difference.
On the simulated input views, ReSplat and DepthSplat reach 24--25\,dB of static PSNR while keeping the vehicles (Table~\ref{tab:carla}).
The lighting code alone moves PSNR by 0.6--0.8\,dB and LPIPS by about 0.01 (Table~\ref{tab:refdist}), so a global color offset costs that much PSNR with the structure unchanged.
And on most frames where \method has a lower PSNR than ReSplat, its LPIPS is much better (Sec.~\ref{sec:mainresults}).
Later sections refer back to this paragraph rather than repeating it.

Every method renders the rectified $448\times448$ crop of each query camera and is scored against the identically preprocessed ground-truth image; both are 8-bit images mapped to $[0,1]$, and to $[-1,1]$ for LPIPS.
Rectification keeps the pinhole intrinsics, which for the barrel distortion of these cameras leaves no empty border, so every pixel of the crop holds image content.
Scores are restricted to static pixels: a pixel is excluded if the semantic label of the query frame (shipped with MTGS; on Ithaca365, a SegFormer-b2 Cityscapes pseudo-label) marks it as sky or as a transient class (person, rider, car, truck, bus, train, motorcycle, bicycle), or if it lies on the ego vehicle.
Labels are rectified and cropped like the images, with nearest-neighbor sampling.
All methods are scored on the same pixels, and predicted masks never determine which pixels receive a score.
PSNR uses the mean squared error over the valid pixels.
SSIM is computed as a dense map over the whole crop ($7\times7$ uniform window) and averaged over the valid pixels, so windows at a mask boundary also see excluded pixels.
For LPIPS, both images are set to zero outside the valid pixels before the network is applied; the zeroed region is identical in both images and contributes no difference by itself, but it enters the receptive fields of features near its boundary.
Masked LPIPS therefore compares the images on a common support rather than restricting the metric exactly to it.

Each split is scored by the mean over its scored frames, and a setting by the mean over its splits.
Every road block contributes the same number of splits to each setting (one on MTGS, three on Ithaca365), so blocks carry equal weight; the aggregate of a dataset is the mean over all its splits, and datasets are reported separately.
Depth is evaluated in two ways, both as the median of $|d-z|/z$ over LiDAR returns $z$ projected into the rectified crop (nearest return per pixel), with $d$ the predicted depth along the optical axis.
\emph{Backbone depth} (Table~\ref{tab:backbone}) uses context views, with $d$ the depth of the backbone's point at that pixel; views with fewer than 200 returns are skipped, the error is averaged over views, and range bins restrict the returns by LiDAR depth.
\emph{Rendered depth} (Table~\ref{tab:refined_depth}) uses the scored frames of the held-out traversals, with $d$ the expected depth rendered from the Gaussians (opacity-weighted mean depth, normalized by the accumulated opacity); both methods use the same rasterizer.
It counts returns within 1--80\,m on the static pixels used for the image metrics, so returns on transient objects in the query frame are excluded, while a vehicle that a method renders over static road counts as an error; pixels must have an accumulated opacity above 0.5 for every compared method, and the error is pooled over the scored frames of a split.
No scale is fitted: every method receives the ground-truth cameras in meters and renders metric depth (median $d/z$ between 0.90 and 1.11 for ours, 0.86 on MTGS and 1.05 on Ithaca365 for ReSplat).
We additionally report mask AP/IoU, Gaussian count, reconstruction latency, peak memory, and rendering throughput.

\subsection{Baselines}
\label{sec:supp_baselines}

We compare the pretrained Wild3R model, ReSplat, MVSplat~\cite{chen2024mvsplat}, DepthSplat~\cite{xu2025depthsplat}, the dense initialization of our adapted Wild3R without refinement, and our compact pipeline at four budgets.
All feedforward baselines run with their official code and released checkpoints, zero-shot, on the same $448\times448$ context crops and ground-truth cameras; none was trained on driving data, and none has a notion of transient objects.
\begin{itemize}
\item \emph{Wild3R}: released checkpoint, run as published: it receives images only and predicts its own cameras, and its appearance reference is its first input, our first context view.
Its cloud is carried to the metric ground-truth frame by one oriented similarity transform fitted between its predicted and the ground-truth context cameras, and rendered with its own rasterizer.
Injecting the ground-truth cameras instead is studied in Sec.~\ref{sec:supp_backbone}.
\item \emph{ReSplat}: released DL3DV base checkpoint for eight views (trained at $512\times960$), run at $448\times448$, which gives one Gaussian per $4{\times}4$ cell, about 100k, with its own initialization and four refinement steps. Table~\ref{tab:views} also uses its released 16-view checkpoint.
\item \emph{MVSplat}: RE10K checkpoint, near/far planes 1 and 100.
\item \emph{DepthSplat}: base RE10K+DL3DV checkpoint ($448\times768$, 2--6 views), near/far 0.5 and 200, Gaussian scales clamped at 0.1.
\end{itemize}
For MVSplat and DepthSplat, the choice of checkpoint (MVSplat RE10K vs.\ ACID, DepthSplat base vs.\ small), input resolution, the scale applied to the metric poses (0.1 for both), and DepthSplat's scale clamp was made once on 24 splits from training blocks (12 nuPlan, 12 Ithaca365), by static PSNR $-$ 20 LPIPS; no test split was used.
The only code change is that their rasterizer receives the principal point, which it otherwise assumes to be centered.
Three comparisons do not depend on the zero-shot setting: \bench is unseen by every method, the lower rows of Table~\ref{tab:carla} isolate our own components, and Table~\ref{tab:transfer} holds the Gaussians fixed and varies only the refiner.
We also test a frozen ReSplat refiner on Wild3R Gaussians to isolate transfer from fine-tuning.

\paragraph{Per-scene optimization.}
As a per-scene optimization reference we run MTGS~\cite{li2025mtgs} with its official code on every MTGS test split, optimized from the same eight context images (front camera) and rendered at the same query cameras.
MTGS keeps its full recipe, including an initialization from the stacked LiDAR scans of the context traversals and LiDAR and monocular depth losses, so it receives more input than the feedforward methods.
We report a 10k-iteration schedule, with all of its schedules scaled by a third, because it scores better with eight views than the default 30k on all three metrics (19.30 vs.\ 19.07\,dB, SSIM 0.586 vs.\ 0.577, LPIPS 0.264 vs.\ 0.273).
MTGS trains on brightness-normalized images (each frame's value channel is scaled by a factor that ships with the data); we map its renders back by the mean factor of the context frames of the query's drive, or of its nearest context drive if the query drive is held out, which uses no information from the query.

\paragraph{Relighting the baselines.}
\label{sec:supp_baseline_relight}
The lower block of Table~\ref{tab:main} and the \emph{relit} rows of Tables~\ref{tab:carla}, \ref{tab:settings}, \ref{tab:carla_difficulty}, \ref{tab:carla_static}, and~\ref{tab:carla_locations} relight the final Gaussians of ReSplat (after its four refinement steps), MVSplat, and DepthSplat with our lighting branch, unchanged and without retraining.
The branch needs each Gaussian's source view and pixel: MVSplat and DepthSplat predict one Gaussian per context pixel, and ReSplat one per $4{\times}4$ cell, in view-major order, so we take the pixel or cell center; projecting the Gaussians into their source views confirms this order (median distance below 2.6 pixels).
Every method gets the same source codes, local features, and target code (the mean code of the even target frames); only \method also runs the color steps.
Relighting a baseline takes 0.03\,s for the codes of the context views and references plus the color transform of its Gaussians (under 1\,ms for ReSplat's 100k, 0.011\,s for 1.6M), timed as in Table~\ref{tab:runtime}; the times in Table~\ref{tab:main} add this to each baseline's own reconstruction time.
The lighting codes are computed from the unmasked images, so relighting needs no transient masks; the 0.28\,s that \method spends on masking the references (Table~\ref{tab:runtime}) serves only the residuals of its color steps, which the baselines do not run.
Without the target code, the same runs reproduce each baseline's scores in Table~\ref{tab:main} exactly.
Our dense 1.6M cloud, too large to refine, reaches 19.75 and 19.03\,dB with the code alone, 1.5 and 0.9\,dB below \method.
We did not relight Wild3R or MTGS.

\begin{figure*}[t]
  \centering
  \includegraphics[width=\textwidth]{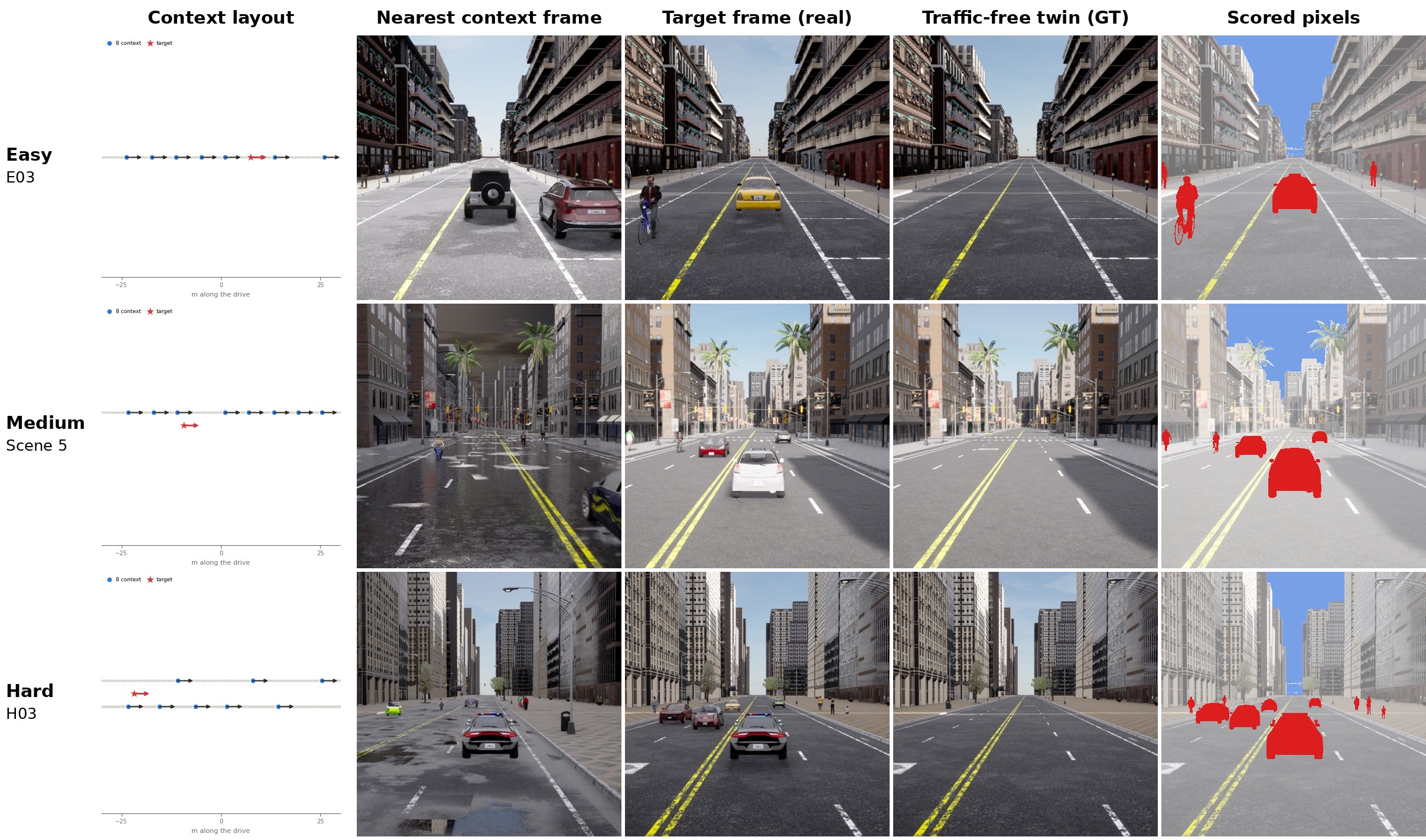}
  \caption{\textbf{The \bench benchmark}, one split per difficulty level (drawn at random with a fixed seed, not chosen by hand). Columns: context cameras (blue) and target (red) from above, the nearest context frame, the real target frame, its traffic-free twin used as ground truth, and the scored pixels (red: \emph{Traffic}; blue: sky, excluded everywhere; everything else is \emph{Static}, and \emph{Full} is Static plus Traffic).}
  \label{fig:carla_examples}
\end{figure*}
\subsection{The \bench Benchmark}
\label{sec:supp_carla}
We render \bench in CARLA~0.9.16~\cite{dosovitskiy2017carla}: 29 road locations of 50\,m in six towns (Towns 02, 03, 05, 10, 12, and 13), 36 captures in total.
\emph{Locations.}
Seven locations in Towns 02, 03, 05, and 10 (Blend to Tree Shade in Table~\ref{tab:carla_locations}) are captured twice with different drive layouts: once with all drives in the same direction under daylight weather, and once with the context drives facing away from two test drives under a weather pool that includes night, fog, dust, and heavy rain.
Ten locations lie in the downtown district of Town 12 and are captured once (Scenes 1--10).
The last twelve, in Towns 10, 12, and 13, are captured once each and are designed for one difficulty level each (four per level; E01--H04).
\emph{Drives.}
Every capture has four or five context drives and one or two target drives, recorded with a front camera on a Nissan Patrol at 10\,Hz while the ego vehicle follows a fixed route at 5\,m/s.
Each drive uses one of CARLA's named weather presets, which combine a sky condition (clear, cloudy, wet, wet-cloudy, soft, mid, or hard rain, fog, dust) with a time of day (morning, noon, afternoon, sunset, night).
\emph{Traffic.}
Each normal drive spawns its traffic from a seeded, per-drive layout along the route: 1--5 moving vehicles (mean 3.9), 1--3 cyclists (mean 2.5), and 2--8 pedestrians on the sidewalks (mean 7.7), all of them moving; the Town 12 and 13 captures also add 0--3 parked vehicles (mean 0.2).
Every target drive is rendered a second time along identical camera poses with no actors at all, parked vehicles included, since a parked car can be gone in another drive; each target frame thus has a pixel-aligned traffic-free ground truth (43 clean target drives, 5{,}287 clean frames).
Context views and lighting references always come from the frames with traffic, as they would on a real vehicle.
Images have the distortion of the nuPlan front camera and are rectified and cropped as in the real datasets.
No stage of any method is trained or tuned on this benchmark.
Each split uses eight context views from a 50\,m window and at most 24 target frames, of which 12 are scored.
Figure~\ref{fig:carla_examples} shows one split per difficulty level.

\paragraph{Difficulty levels.}
We group the 173 splits into three difficulty levels by capture conditions.
\emph{Easy} (44 splits): daylight, dry target drive, context in the target's lane (median lateral distance below 1\,m).
\emph{Medium} (81 splits): the same conditions with context 3--8\,m to the side; 21 of these splits take their context from several drives in the target's direction.
\emph{Hard} (48 splits): lateral distances of 3--15\,m with large changes of weather and time of day between drives, including night, fog, dust, and heavy rain; in 14 of these splits the context mixes drives from both directions.
For the last twelve locations the level is a property of the location.
For the first 17 it is set per split: in the same-direction captures and in Town 12, each context drive gives one single-drive split (Easy if it drives in the target's lane, Medium otherwise) and all drives together give one multi-drive split (Medium), while all splits of the second capture of the seven locations are Hard.
Table~\ref{tab:carla_locations} lists the splits per level for every location.

\paragraph{Scores.}
Every score uses the traffic-free frame as ground truth and excludes its sky.
\emph{Full} scores all remaining pixels, so traffic and baked-in ghosts both count.
\emph{Static} also excludes the transients of the real target frame, as the real-data metrics do.
\emph{Traffic} PSNR is restricted to pixels that hold a transient in the real target frame but not in the clean one; frames with fewer than 50 such pixels are left out of it.

\section{Additional Results on Real Data}
\label{sec:supp_real}
At equal budget and with the same target-drive references, \method leads relit ReSplat in every setting, though only by 0.04\,dB with same-direction context on Ithaca365; without any image of the target drive it leads on MTGS except with opposing-direction context, and trails ReSplat on Ithaca365 in PSNR while keeping a much lower LPIPS.
It outperforms per-scene MTGS on all three metrics at about 2{,}000 times lower cost.

\subsection{Comparison with the Baselines}

Table~\ref{tab:main} compares \method with ReSplat, MVSplat, DepthSplat, and Wild3R on the same context views, cameras, and scored pixels.
With target-drive references, all feedforward methods receive the same lighting code from our branch (Sec.~\ref{sec:supp_baseline_relight}).
Across budgets (Table~\ref{tab:budgets}), \method is then 1.21--1.35\,dB above relit ReSplat on MTGS and between 0.03\,dB below (400k) and 0.21\,dB above (100k) on Ithaca365, and its LPIPS is lower by about 0.1 (0.214--0.232 vs.\ 0.328 on MTGS, 0.218--0.225 vs.\ 0.331 on Ithaca365).
Relit ReSplat keeps the higher SSIM on Ithaca365 (0.484 vs.\ 0.450--0.461).
SSIM sides with PSNR here, not with LPIPS: on Ithaca365, where every method stays below 0.5, we lose to ReSplat with and without references, while on MTGS we match or exceed it.
We attribute this to ReSplat's smoother renders (Sec.~\ref{sec:supp_metrics}). Our ablations show the same preference, as merging Gaussians gives the highest SSIM but a far worse LPIPS than our head (Table~\ref{tab:sampling}); we did not test this explanation further.
Without references, \method beats ReSplat on both metrics on 59\,\% of the MTGS and 23\,\% of the Ithaca365 frames.
Relit DepthSplat is the strongest baseline in LPIPS (0.283 on MTGS, 0.261 on Ithaca365) but stays 2.20--2.34\,dB (MTGS) and 0.56--0.80\,dB (Ithaca365) below \method in PSNR; MVSplat is below every other pipeline except Wild3R.
ReSplat, MVSplat, and DepthSplat reconstruct the vehicles of the context drives together with the static scene; at a held-out query they reappear as smeared ghosts over the road and facades, which the static-pixel metrics still penalize because the scored pixels behind them are static.
Sec.~\ref{sec:transienteval} measures this against traffic-free ground truth.
Without any image of the target drive, \method leads on MTGS (19.74--19.97 vs.\ 19.38\,dB for ReSplat) and on Ithaca365 matches or exceeds DepthSplat but stays below ReSplat (18.53--18.73 vs.\ 18.53 and 19.04\,dB), while keeping its LPIPS advantage (Table~\ref{tab:budgets}).

Two training seeds of the same model differ by about 0.15\,dB in PSNR before refinement and by at most 0.06\,dB after it, so we do not treat differences below 0.1\,dB as significant.

\subsection{Per-Setting Results}

\begin{table}[t]
  \centering
  \caption{\textbf{Per-setting results at equal budget} (PSNR/LPIPS, 100k Gaussians, 6 MTGS and 15 Ithaca365 splits per setting, scored as in Table~\ref{tab:main}). Bold marks the better of each pair. ReSplat's single held-out rows coincide on both datasets in PSNR and LPIPS; their SSIM differs (0.614 vs.\ 0.459). Ithaca365 has no opposing-direction drives in the held-out blocks.}
  \label{tab:settings}
  \footnotesize
  \setlength{\tabcolsep}{2.5pt}
  \resizebox{\linewidth}{!}{%
  \begin{tabular}{lcccc}
    \toprule
    & \multicolumn{2}{c}{No target-drive image} & \multicolumn{2}{c}{Target-drive references} \\
    \cmidrule(lr){2-3}\cmidrule(lr){4-5}
    Setting & ReSplat & \method & ReSplat, relit & \method \\
    \midrule
    \multicolumn{5}{l}{\emph{MTGS}} \\
    Single held-out & 18.77/.340 & \textbf{19.61}/\textbf{.234} & 19.58/.339 & \textbf{21.43}/\textbf{.216} \\
    Same direction & 19.45/.324 & \textbf{20.30}/\textbf{.234} & 19.96/.323 & \textbf{21.58}/\textbf{.217} \\
    Opposing direction & \textbf{19.93}/.323 & 19.33/\textbf{.273} & 20.20/.322 & \textbf{20.37}/\textbf{.262} \\
    All & 19.38/.329 & \textbf{19.74}/\textbf{.247} & 19.91/.328 & \textbf{21.12}/\textbf{.232} \\
    \midrule
    \multicolumn{5}{l}{\emph{Ithaca365}} \\
    Single held-out & \textbf{18.77}/.340 & 18.59/\textbf{.261} & 19.45/.337 & \textbf{19.81}/\textbf{.229} \\
    Same direction & \textbf{19.31}/.325 & 18.88/\textbf{.248} & 20.12/.324 & \textbf{20.16}/\textbf{.220} \\
    All & \textbf{19.04}/.333 & 18.73/\textbf{.255} & 19.78/.331 & \textbf{19.99}/\textbf{.225} \\
    \bottomrule
  \end{tabular}}
\end{table}

Table~\ref{tab:settings} shows where the gains come from, at ReSplat's budget of 100k, comparing methods that receive the same information.
With target-drive references, \method exceeds relit ReSplat by 1.85\,dB on single held-out synthesis, 1.62\,dB on same-direction and 0.17\,dB on opposing-direction synthesis on MTGS, and by 0.36 and 0.04\,dB on single held-out and same-direction synthesis on Ithaca365.
Without them, \method exceeds ReSplat on MTGS by 0.84 and 0.85\,dB on single held-out and same-direction synthesis but trails it by 0.60\,dB with opposing-direction context, and it trails ReSplat on Ithaca365 by 0.18 and 0.43\,dB.
The lead is thus largest when the context comes from one other drive and smallest with opposing-direction context, where the drives share few surfaces.
ReSplat's LPIPS is 0.05--0.12 higher than ours in every setting, with and without references.
DepthSplat falls below ReSplat and below \method without references in PSNR in every setting (MTGS 17.87--18.70\,dB, Ithaca365 18.37--18.68\,dB).

\subsection{Qualitative Comparison}
\label{sec:supp_qual}
\begin{figure*}[t]
  \centering
  \includegraphics[width=\textwidth]{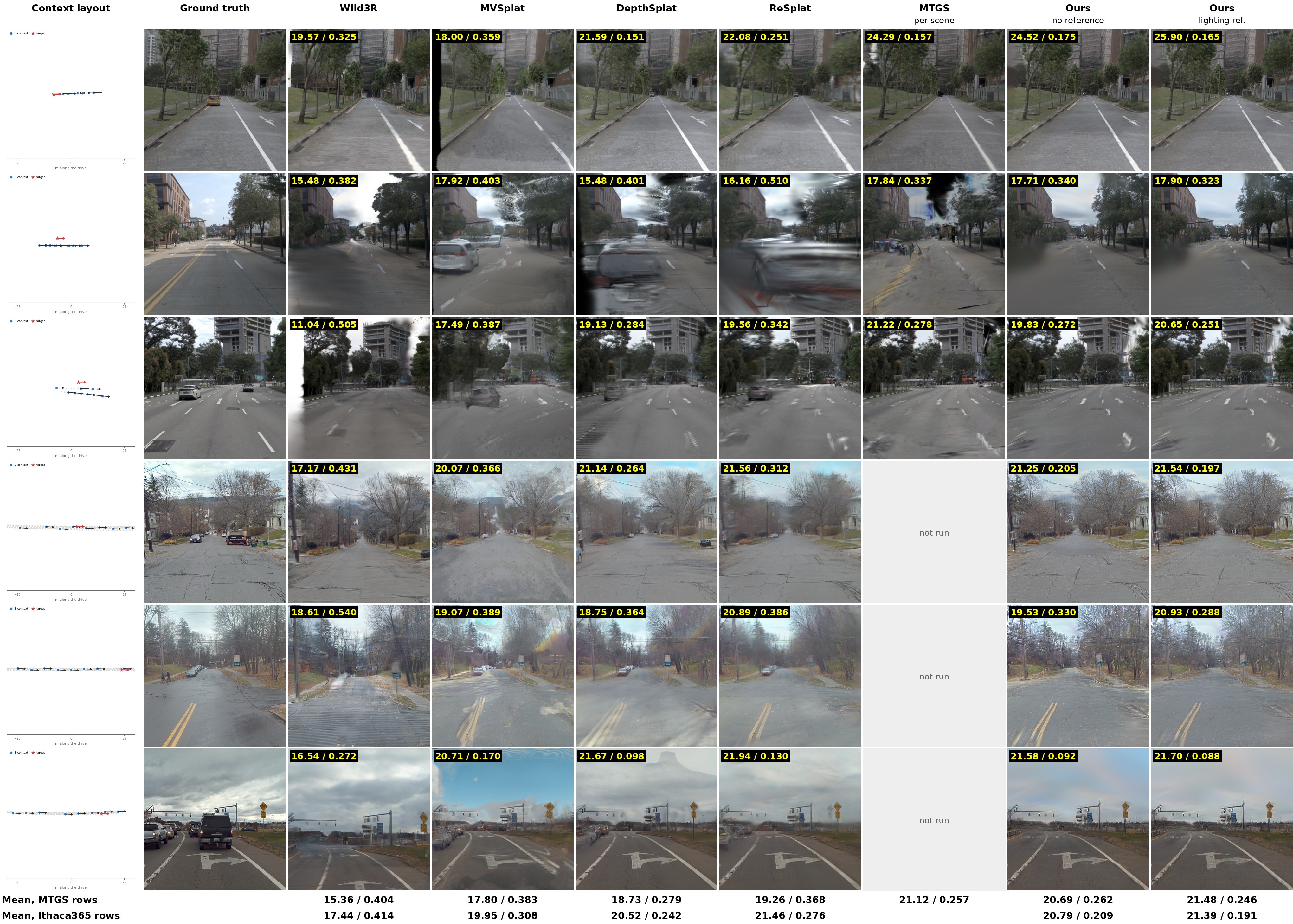}
  \caption{\textbf{Unselected comparison on held-out traversals.} Rows: three MTGS and three Ithaca365 splits, each from a different road block, and one scored frame per split, all drawn at random with a fixed seed. Columns 3--8 receive no image of the target drive (Wild3R, MVSplat, DepthSplat, ReSplat, per-scene MTGS, which was run on MTGS only, and \method at 200k); the last column is \method with lighting references. Static PSNR/LPIPS per frame; the footer gives the means over the rows of each dataset.}
  \label{fig:random_qual}
\end{figure*}

Figure~\ref{fig:random_qual} compares all methods on frames that were not chosen by hand: three MTGS and three Ithaca365 splits, each from a different road block, and one scored frame of each, all drawn with a fixed random seed.
In row 2, MVSplat, DepthSplat, and ReSplat render a vehicle of the context drive onto the empty road, and MTGS breaks up; MVSplat and DepthSplat also leave dark regions at the image border (rows 1 and 2).
\method keeps the road free of context vehicles but blurs some distant structure and lane markings.
On the three Ithaca365 rows, ReSplat has a higher PSNR than \method without a reference, as in Table~\ref{tab:main}, while its LPIPS is worse.
The qualitative figure of the main paper (Fig.~\ref{fig:qualitative}) is hand-picked; the failure cases are in Fig.~\ref{fig:failures}.

\begin{figure*}[p]
  \centering
  \includegraphics[width=\textwidth]{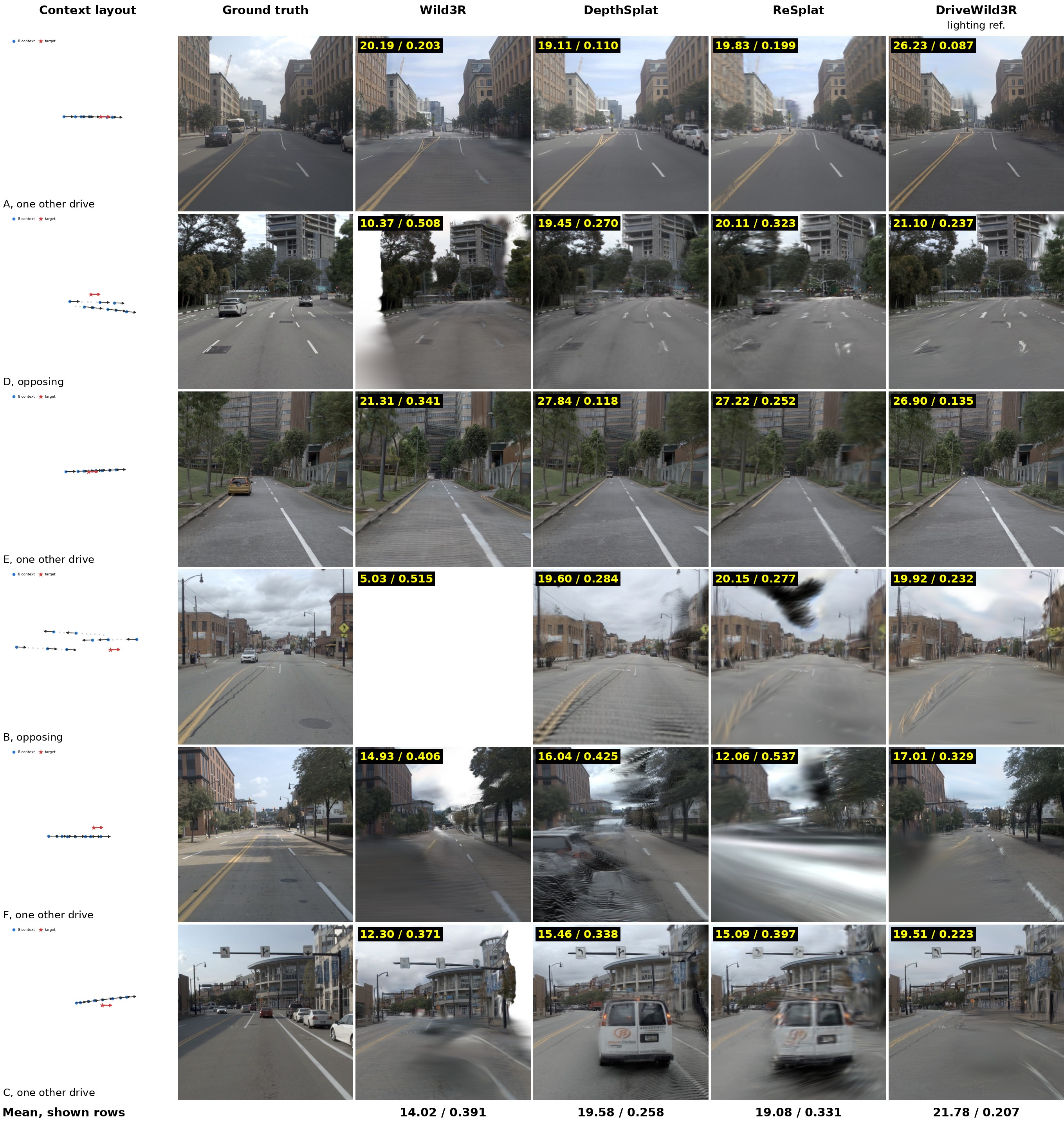}
  \caption{\textbf{Unselected comparison on MTGS.} One split per road block (A--F) and one scored frame per split, drawn at random with a fixed seed. Columns: context cameras (blue) and target (red) from above, labeled with block and setting; ground truth; Wild3R, DepthSplat, and ReSplat; \method at 200k with lighting references. Static PSNR/LPIPS per frame; the footer gives the means over the shown rows.}
  \label{fig:supp_qual_mtgs}
\end{figure*}

\begin{figure*}[p]
  \centering
  \includegraphics[width=\textwidth]{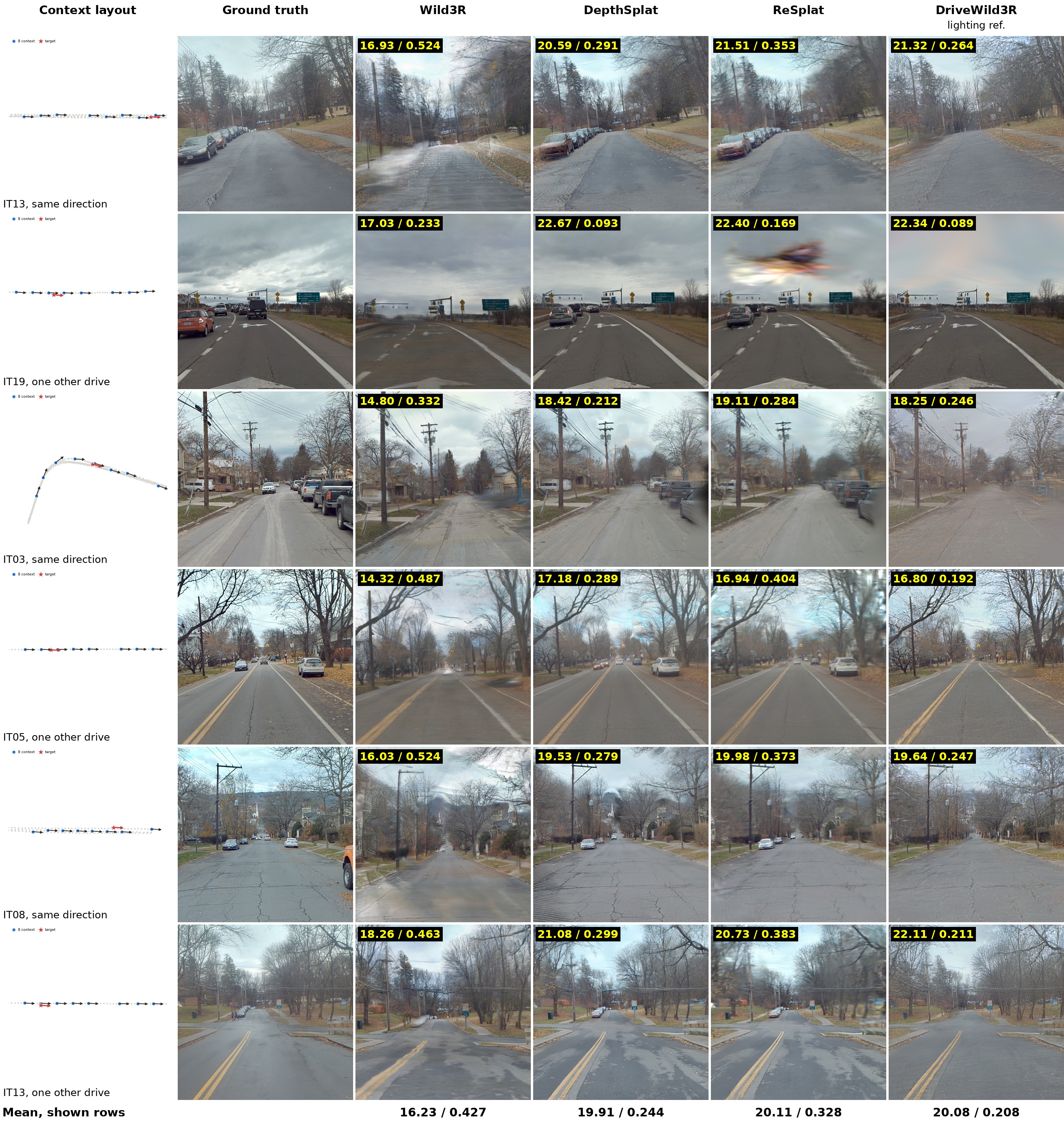}
  \caption{\textbf{Unselected comparison on Ithaca365.} Splits from all five test blocks plus a second split of one block, one scored frame each, drawn at random with a fixed seed. Columns and scores as in Fig.~\ref{fig:supp_qual_mtgs}, except that all methods use the LiDAR-registered poses (Sec.~\ref{sec:supp_ithaca_reg}).}
  \label{fig:supp_qual_ithaca}
\end{figure*}

\begin{figure*}[p]
  \centering
  \includegraphics[width=\textwidth]{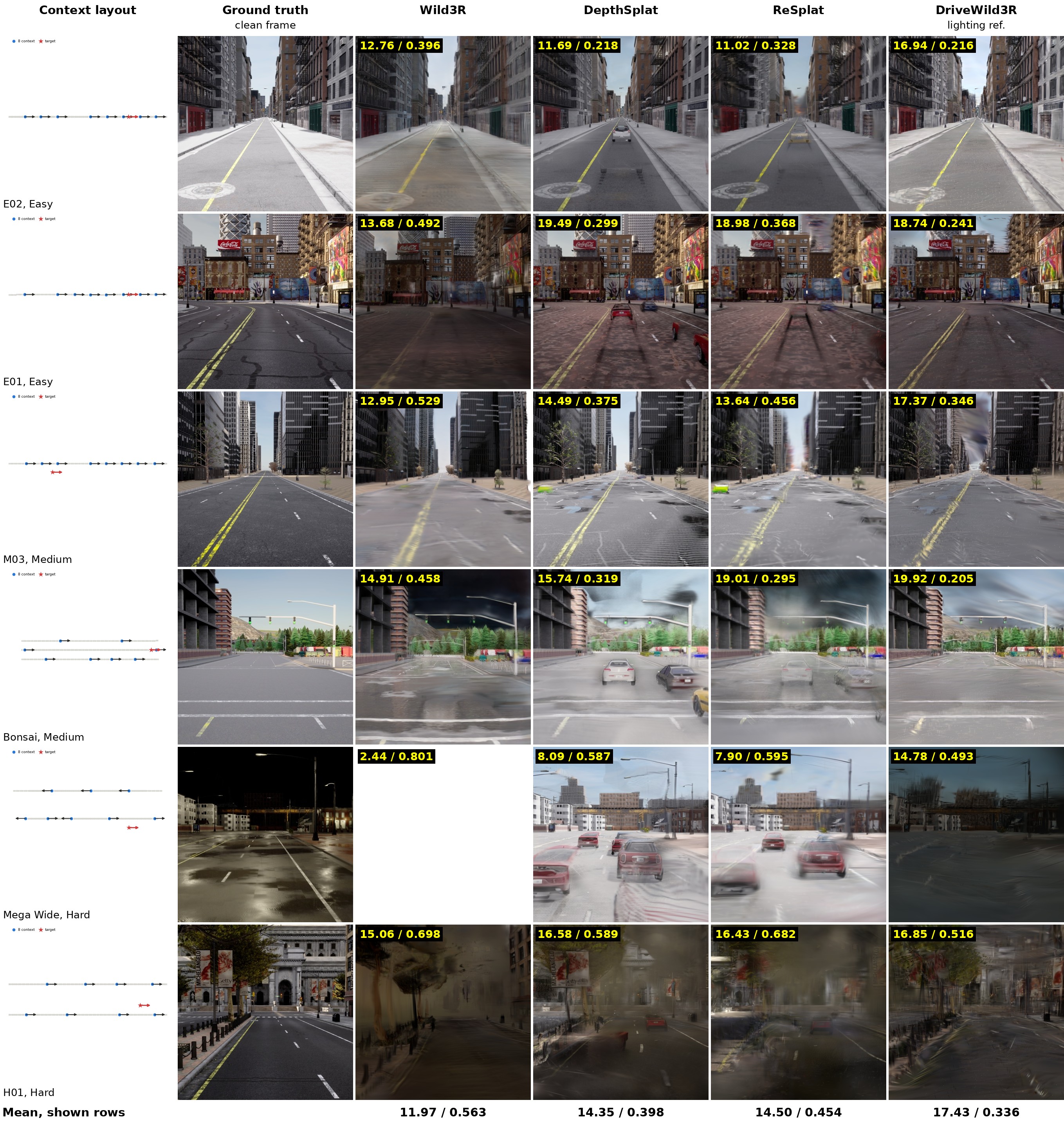}
  \caption{\textbf{Unselected comparison on \bench.} Two held-out splits per difficulty level from six different locations (names as in Table~\ref{tab:carla_locations}), one scored frame each, drawn at random with a fixed seed. The ground truth is the traffic-free twin of the target frame. \method uses predicted transient masks, as in Table~\ref{tab:carla}. Full PSNR/LPIPS (sky excluded) against the clean frame per frame; the footer gives the means over the shown rows.}
  \label{fig:supp_qual_carla}
\end{figure*}

Figures~\ref{fig:supp_qual_mtgs}--\ref{fig:supp_qual_carla} add six further unselected frames per dataset for MTGS, Ithaca365, and \bench, comparing \method with lighting references against Wild3R, DepthSplat, and ReSplat.
Rows come from different road blocks (MTGS: one per block; Ithaca365: all five test blocks and a second split of one of them; \bench: two splits per difficulty level from six locations), and no frame of Fig.~\ref{fig:random_qual} is repeated.
On MTGS, \method has the highest PSNR on four and the lowest LPIPS on five of the six rows. DepthSplat and ReSplat render vehicles of the context drives onto the road (rows 2, 5, and 6), and Wild3R fails on the opposing-direction split of row 4, where it renders an empty frame.
In Fig.~\ref{fig:supp_qual_ithaca}, all methods use the LiDAR-registered poses of Sec.~\ref{sec:supp_ithaca_reg}.
On the six rows, ReSplat has the highest PSNR on three, DepthSplat on two, and \method on one, while \method has the lowest LPIPS on five; \method also removes cars that are parked in the target frame.
On \bench, \method has the highest PSNR on five and the lowest LPIPS on all six rows. DepthSplat and ReSplat copy the traffic of the context drives into the clean frame, and on the night target of row 5 they render the daylight context, while \method follows the target's lighting.
All methods fail on the Hard split of row 6, whose context drives are recorded in fog.

\subsection{Fusing Traversal-Specific Shadows}
\label{sec:supp_shadows}
\begin{figure*}[t]
  \centering
  \includegraphics[width=\textwidth]{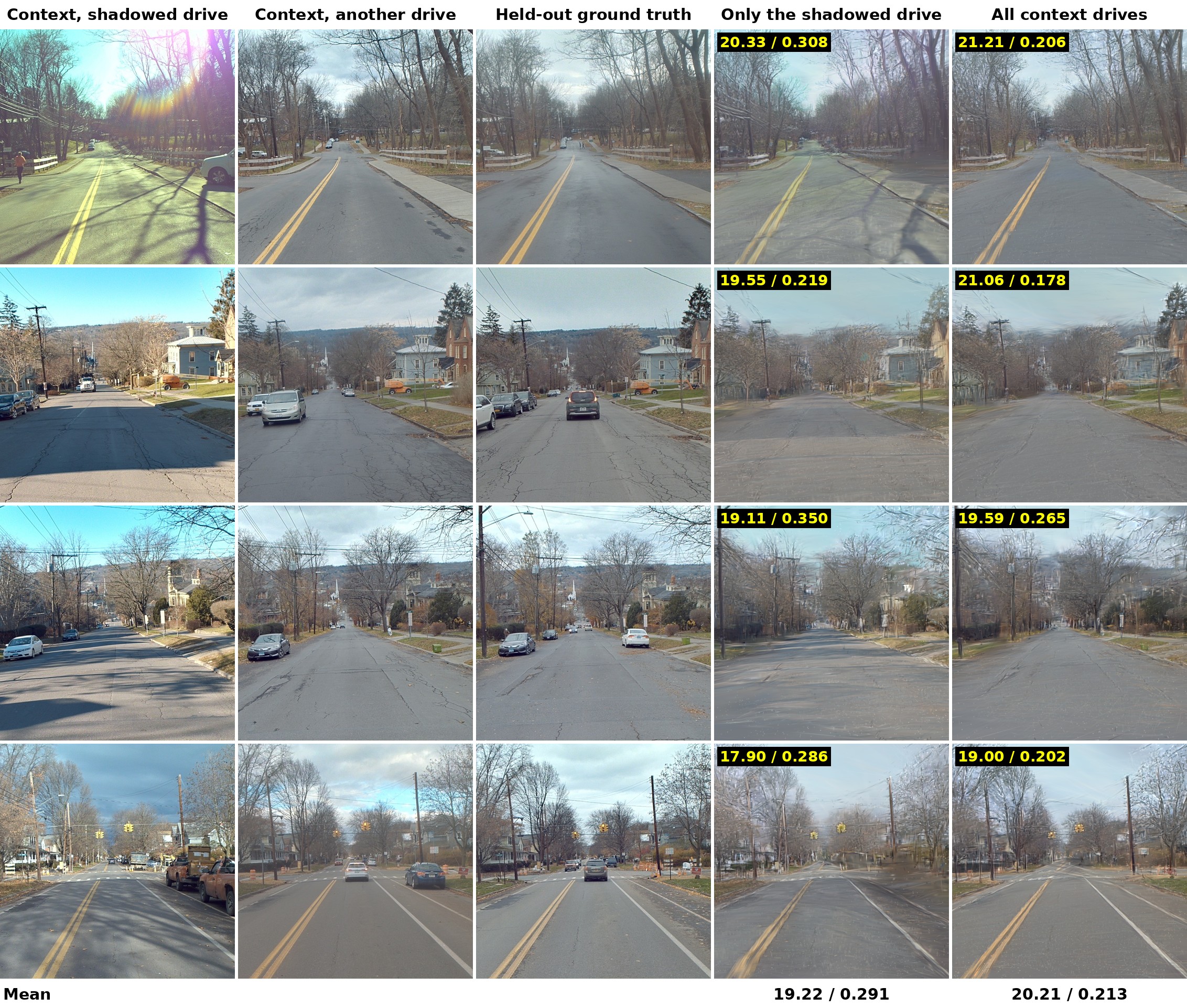}
  \caption{\textbf{Shadows observed by only one context drive.} Four held-out frames from Ithaca365, selected to show the effect. Full pipeline at 200k Gaussians, refined, with lighting references from the held-out drive and the LiDAR-registered poses (Sec.~\ref{sec:supp_ithaca_reg}). Columns 1--2: the nearest context frame of the drive with hard shadows and of another drive. Columns 4--5: reconstruction from eight views of the shadowed drive alone, and from all context drives. Numbers are static PSNR/LPIPS; the footer gives the means over the four frames.}
  \label{fig:shadows}
\end{figure*}
Cast shadows change with the time of day, so they differ between traversals, much like transient objects.
Figure~\ref{fig:shadows} shows four held-out frames from Ithaca365 where one context drive has hard shadows on the road.
A reconstruction from that drive alone reproduces them, while a reconstruction from all context drives renders the road evenly lit.

\subsection{Per-Scene Optimization}

MTGS, optimized on each scene from the same eight front-camera images and additionally supervised with LiDAR, reaches 19.30\,dB and LPIPS 0.264 on MTGS.
Like \method without a lighting reference, it receives no image of the target drive, and \method in that setting exceeds it at every budget (19.74--19.97\,dB, LPIPS 0.228--0.247; Table~\ref{tab:budgets}).
With references, \method is 1.82--1.96\,dB above it.
It is level with ReSplat in PSNR (19.08--19.43 vs.\ 18.77--19.93\,dB per setting) and well ahead in LPIPS (0.251--0.284 vs.\ 0.323--0.340).
With eight views its static model absorbs vehicles that stay in front of the ego car in every context frame, and these reappear at the held-out query, as for the feedforward baselines.
The two approaches differ most in cost.
MTGS needs about 38\,min per scene for its 10k iterations, computed from the per-iteration time of our 30k runs.
\method reconstructs a scene in a single feedforward pass followed by two geometry and two color refinement steps, in 1.15\,s in total at 200k (Table~\ref{tab:runtime}), about 2{,}000 times faster, and with higher quality.

\section{Transient Removal}
\label{sec:supp_transient}
On \bench \method leads every baseline at every difficulty level, also when the baselines are relit with our lighting branch: on held-out frames its Traffic PSNR is 3.0\,dB above the best relit baseline (relit ReSplat), which receives the same lighting references.
Its predicted masks perform like ground-truth masks, and removing the masks costs mainly on the pixels of the traffic.

\subsection{Held-Out Frames}

With masks from its own transient head, \method leads every column on all three metrics (Tables~\ref{tab:carla_difficulty} and~\ref{tab:carla_static}).
Against the best un-relit baseline it wins on 166 and 164 of the 173 splits in Full and Static PSNR; against the best relit baseline on 159 and 159, and in Traffic PSNR on 141.
On held-out frames Traffic PSNR measures background recovery: \method reaches 19.05\,dB, against 14.08\,dB for the best un-relit baseline (ReSplat) and 16.03\,dB for the best relit one (relit ReSplat), while the real target frame, which keeps its vehicles, scores 10.44\,dB.
Full and Static stay within 0.5\,dB of each other for every method, because they hide only the traffic of the target frame.
The vehicles that baselines bake in come from the context drives and appear elsewhere in the target view, on pixels that both metrics score (Fig.~\ref{fig:carla}).
Without a lighting reference, \method uses no image of the target drive; it then leads the un-relit ReSplat, MVSplat, and DepthSplat, which receive no such image either, in every column (15.58\,dB Traffic PSNR, 1.5\,dB above the best of them).
The relit baselines do receive the references, and without them \method trails relit ReSplat (15.58 vs.\ 16.03\,dB Traffic, 14.64 vs.\ 15.39\,dB Full) while keeping a lower LPIPS (0.365 vs.\ 0.383 for relit DepthSplat).

\begin{table}[t]
  \centering
  \caption{\textbf{Held-out views of \bench per difficulty level} (splits in brackets; 200k Gaussians, predicted masks). The clean frame is the ground truth. Upper rows of each block: no image of the target drive; \emph{relit} rows and the last row: lighting references from the target drive (Sec.~\ref{sec:supp_baseline_relight}).}
  \label{tab:carla_difficulty}
  \footnotesize
  \setlength{\tabcolsep}{3.5pt}
  \resizebox{\linewidth}{!}{%
  \begin{tabular}{@{}lcccc@{}}
    \toprule
    & Easy (44) & Medium (81) & Hard (48) & All (173) \\
    \midrule
    \multicolumn{5}{l}{\emph{Full, PSNR/LPIPS}} \\
    Wild3R & 13.98/.433 & 13.15/.465 & 10.65/.581 & 12.67/.489 \\
    ReSplat & 15.69/.347 & 14.47/.440 & 11.66/.580 & 14.00/.455 \\
    MVSplat & 15.07/.359 & 14.38/.441 & 11.53/.565 & 13.76/.455 \\
    DepthSplat & 15.63/.276 & 13.96/.373 & 11.17/.530 & 13.61/.392 \\
    \method, w/o lighting ref. & 15.98/.276 & 15.14/.345 & 12.55/.482 & 14.64/.365 \\
    ReSplat, relit & 16.98/.343 & 15.79/.434 & 13.28/.564 & 15.39/.447 \\
    MVSplat, relit & 16.12/.353 & 15.46/.433 & 13.23/.544 & 15.01/.444 \\
    DepthSplat, relit & 16.96/.271 & 15.12/.367 & 12.64/.512 & 14.90/.383 \\
    \method & \textbf{18.48}/\textbf{.239} & \textbf{17.72}/\textbf{.304} & \textbf{15.05}/\textbf{.439} & \textbf{17.17}/\textbf{.325} \\
    \midrule
    \multicolumn{5}{l}{\emph{Traffic, PSNR}} \\
    Real target frame & 10.73 & 10.06 & 10.81 & 10.44 \\
    Wild3R & 14.45 & 14.33 & 12.58 & 13.88 \\
    ReSplat & 14.52 & 14.73 & 12.57 & 14.08 \\
    MVSplat & 12.99 & 13.94 & 11.97 & 13.15 \\
    DepthSplat & 13.59 & 13.79 & 11.90 & 13.22 \\
    \method, w/o lighting ref. & 16.17 & 15.89 & 14.53 & 15.58 \\
    ReSplat, relit & 16.05 & 16.85 & 14.62 & 16.03 \\
    MVSplat, relit & 14.10 & 15.75 & 14.11 & 14.88 \\
    DepthSplat, relit & 15.01 & 15.61 & 13.71 & 14.93 \\
    \method & \textbf{19.69} & \textbf{19.60} & \textbf{17.54} & \textbf{19.05} \\
    \bottomrule
  \end{tabular}}
\end{table}

\begin{table}[t]
  \centering
  \caption{\textbf{Static PSNR/LPIPS on \bench}, as in Table~\ref{tab:carla_difficulty}, with the sky and the real target frame's transients excluded.}
  \label{tab:carla_static}
  \footnotesize
  \setlength{\tabcolsep}{3.5pt}
  \resizebox{\linewidth}{!}{%
  \begin{tabular}{@{}lcccc@{}}
    \toprule
    & Easy (44) & Medium (81) & Hard (48) & All (173) \\
    \midrule
    Wild3R & 13.99/.388 & 13.14/.412 & 10.60/.524 & 12.65/.437 \\
    ReSplat & 15.85/.309 & 14.53/.387 & 11.67/.533 & 14.07/.407 \\
    MVSplat & 15.28/.313 & 14.45/.384 & 11.54/.522 & 13.86/.404 \\
    DepthSplat & 15.84/.238 & 14.04/.328 & 11.19/.495 & 13.71/.352 \\
    \method, w/o lighting ref. & 16.02/.247 & 15.17/.303 & 12.52/.436 & 14.65/.326 \\
    ReSplat, relit & 17.12/.304 & 15.81/.380 & 13.28/.514 & 15.44/.398 \\
    MVSplat, relit & 16.33/.306 & 15.49/.375 & 13.23/.500 & 15.08/.392 \\
    DepthSplat, relit & 17.17/.232 & 15.17/.321 & 12.65/.475 & 14.98/.341 \\
    \method & \textbf{18.48}/\textbf{.212} & \textbf{17.69}/\textbf{.264} & \textbf{15.00}/\textbf{.393} & \textbf{17.14}/\textbf{.287} \\
    \bottomrule
  \end{tabular}}
\end{table}

\subsection{Re-Rendering the Input Views}
\label{sec:supp_carla_input}
On held-out frames the scored traffic was never in the input of any method.
To test removal where the vehicles are in the input, we re-render the context views themselves and score them against their traffic-free copies.
We use the 50 splits whose context drives have a clean copy: 43 with eight views from one drive, and 7 with eight views from two drives, from the locations with large changes of weather and time of day.
Weights and scoring are the same as for the held-out frames, and no method sees a clean copy.
Because these frames come from the input drives, this test stays separate from the novel-view averages.

\begin{table}[t]
  \centering
  \caption{\textbf{Input views against their traffic-free copies} (PSNR, 200k Gaussians for \method, with masks from its transient head unless noted). Bold marks the best method. Table~\ref{tab:carla} gives the average over all 50 splits, including relit baselines.}
  \label{tab:carla_ctx}
  \footnotesize
  \setlength{\tabcolsep}{3.5pt}
  \resizebox{\linewidth}{!}{%
  \begin{tabular}{@{}lcccc@{}}
    \toprule
    & \multicolumn{2}{c}{One drive (43)} & \multicolumn{2}{c}{Two drives (7)} \\
    \cmidrule(lr){2-3}\cmidrule(lr){4-5}
    & Traffic & Static & Traffic & Static \\
    \midrule
    Real frame & 10.56 & 31.13 & 9.92 & 29.89 \\
    Wild3R & 17.04 & 16.05 & 13.02 & 13.09 \\
    ReSplat & 14.43 & 24.45 & 14.11 & \textbf{20.93} \\
    MVSplat & 13.37 & 19.32 & 12.74 & 15.97 \\
    DepthSplat & 13.00 & \textbf{25.71} & 12.18 & 19.07 \\
    \midrule
    \method, dense, w/o refiner & 18.77 & 22.31 & 14.94 & 17.77 \\
    \method, w/o lighting ref., w/o masks & 19.26 & 22.62 & 15.03 & 18.23 \\
    \method, w/o lighting ref. & \textbf{21.12} & 22.72 & 15.51 & 18.26 \\
    \method, w/o masks & 19.37 & 22.52 & 15.46 & 17.82 \\
    \method & 21.09 & 22.61 & \textbf{15.99} & 17.81 \\
    \bottomrule
  \end{tabular}}
\end{table}

Table~\ref{tab:carla_ctx} separates the two groups.
The margin is smallest with two drives, where ReSplat is the best baseline (16.0 vs.\ 14.1\,dB) and Wild3R drops to 13.0\,dB.
With masks from the transient head, \method has a lower Traffic PSNR than the best baseline on two of the 50 splits.
One has almost no traffic, so the real frame already scores 27.6\,dB, and the other is a two-drive split, by 0.07\,dB; without a reference one more two-drive split is lost, by 1.6\,dB.
\method has the highest static PSNR on only 2 of the 50 splits.
Ground-truth masks give 20.39\,dB on Traffic and masks from the head 20.38\,dB, so the result does not depend on the masks marking exactly the scored pixels.
Removing the masks costs 1.7\,dB with one drive and 0.5\,dB with two.
The dense 1.6M model of our backbone, with neither masks nor refinement, beats the best baseline on 42 of the 50 splits, and the full \method is above this dense model on all 50 splits (48 without a reference).
On static pixels the dense model keeps the lower LPIPS (0.117 vs.\ 0.142).
The even context frames are lighting references for \method; scoring only the odd frames, which are never references, gives the same picture (20.42 vs.\ 16.48\,dB for Wild3R).
Relighting the baselines changes their Traffic PSNR on the input views by at most 0.04\,dB (Table~\ref{tab:carla}).

\begin{figure*}[t]
  \centering
  \includegraphics[width=\textwidth]{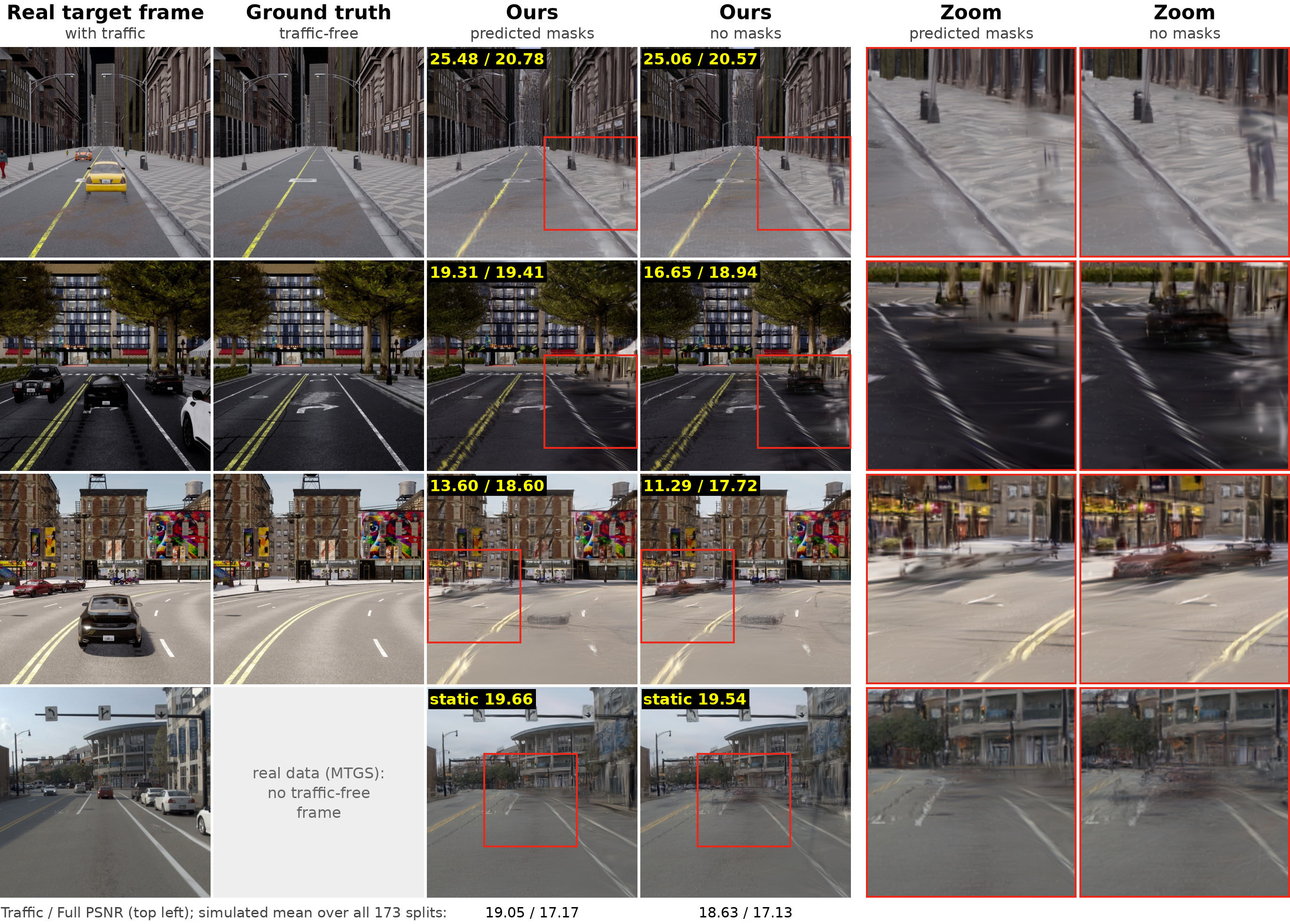}
  \caption{\textbf{Effect of the transient masks} (200k Gaussians, lighting references from the target drive, same weights in both columns). Rows 1--3: \bench, Traffic PSNR\,/\,Full PSNR per frame; the footer gives the means over all 173 splits. Row 4: a held-out MTGS drive, which has no traffic-free frame, so only static PSNR is given. Without masks, vehicles and pedestrians can leave ghosts and smears while Full PSNR barely changes. Masks do not remove every artifact (rows 1, 3), and a ghost outside the target's own traffic lowers Traffic PSNR little (row 1). Frames selected for a visible effect.}
  \label{fig:nomask}
\end{figure*}
\subsection{Predicted, Ground-Truth, and No Masks}
\label{sec:supp_transient_data}
\begin{figure}[t]
  \centering
  \includegraphics[width=\linewidth]{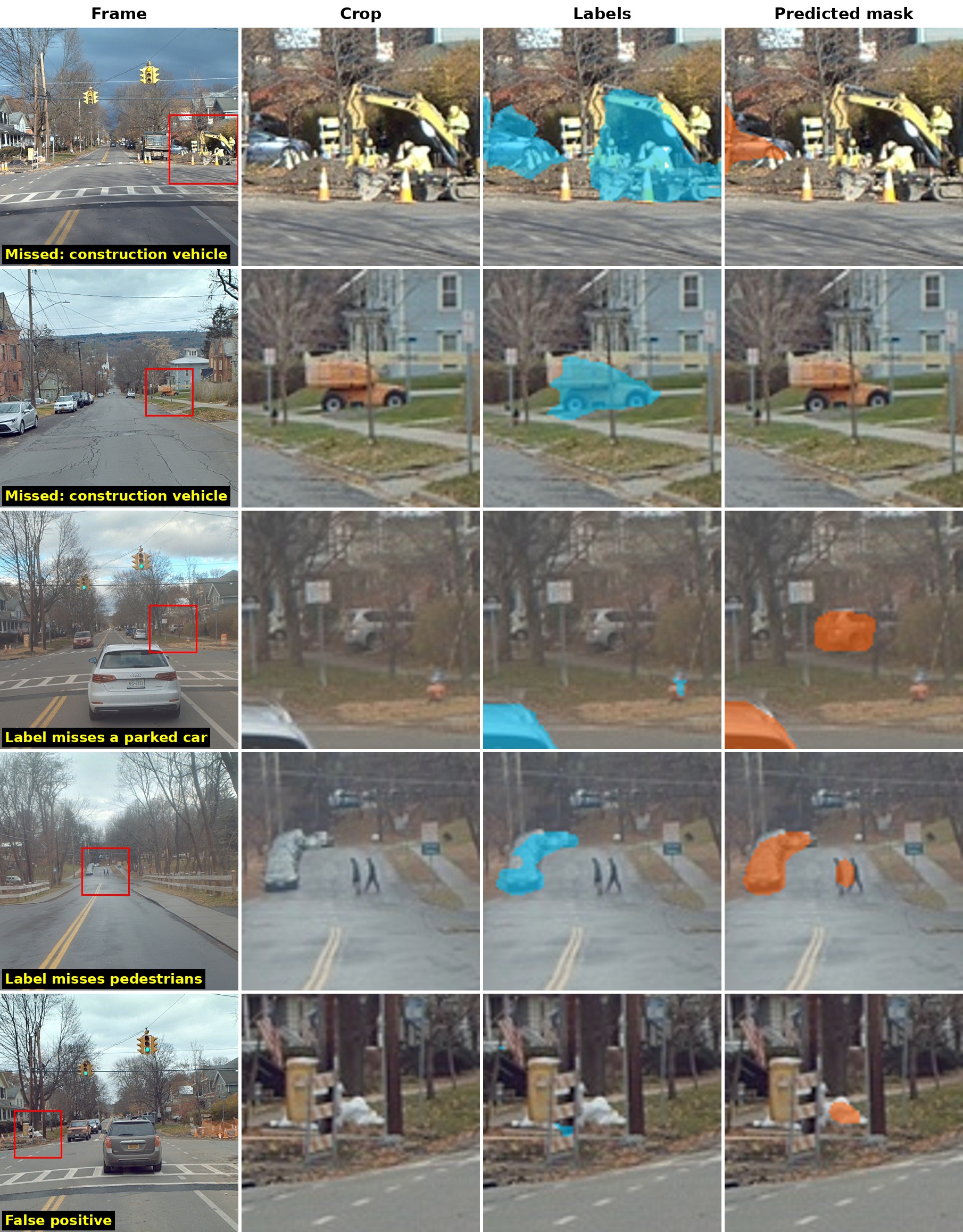}
  \caption{\textbf{Disagreements between predicted masks (orange) and labels (blue)} on Ithaca365 test blocks. The labels are the SegFormer-B2 pseudo-labels used for scoring; human annotations exist only for the 54 frames of the AP/IoU evaluation. Examples picked by hand from the largest. Rows 1--2: construction vehicles the head misses. Rows 3--4: road users the labels miss. Row 5: a false positive on a white bag.}
  \label{fig:mask_failures}
\end{figure}

\paragraph{Mask accuracy.}
The transient head reaches 0.97 AP / 0.84 IoU on six held-out MTGS blocks, against the labels shipped with MTGS, and 0.96 AP / 0.83 IoU on 54 Ithaca365 frames with human annotations.
These annotations are the 2D instance masks that Ithaca365 distributes for about 6.3k of its images (car, truck, pedestrian, bicyclist, bus, motorcyclist).
The 54 frames are all frames of our Ithaca365 data that carry such an annotation, so none was chosen by us.
26 of them lie in the test blocks IT03 and IT05 and 28 in IT00, IT01, IT02, and IT04, whose images the backbone saw during fine-tuning; the head itself never saw Ithaca365.
The labels of our Ithaca365 test splits, used for scoring and in Fig.~\ref{fig:mask_failures}, are SegFormer-B2 pseudo-labels.

\paragraph{Predicted vs.\ label masks on real data.}
To check that the head is accurate enough, we run the full pipeline twice with the same checkpoints: once with the transient masks of the dataset labels (on Ithaca365 the SegFormer-B2 pseudo-labels) and once with the masks predicted by our head, which replace the labels everywhere the pipeline uses transient masks: the sampler, the refiner, and the lighting-reference frames.
Scoring always uses the same labels.
Table~\ref{tab:predmask} shows that the two runs are indistinguishable: at every budget and on both datasets, PSNR differs by at most 0.02\,dB and LPIPS by at most 0.001.
\method therefore needs no external source of transient masks.
It still reads the sky from semantic labels: the sampler down-weights sky pixels (Eq.~\ref{eq:sampling}), and the transient head does not predict sky.
We did not run the pipeline without sky labels; a deployed system would need a sky segmenter or a uniform sky weight.

\begin{table}[t]
  \centering
  \caption{\textbf{Transient masks on \bench} (200k, 173 splits, scored as in Table~\ref{tab:carla}). The masks are replaced in the sampler, the refiner, and the lighting references, with identical weights; \emph{None} uses no transient information. Sky comes from the simulator's labels in every row.}
  \label{tab:carla_masks}
  \footnotesize
  \setlength{\tabcolsep}{3pt}
  \begin{tabular}{lcccccc}
    \toprule
    & \multicolumn{2}{c}{Full} & \multicolumn{4}{c}{Traffic PSNR} \\
    \cmidrule(lr){2-3}\cmidrule(lr){4-7}
    Transient masks & PSNR & LPIPS & Easy & Medium & Hard & All \\
    \midrule
    \multicolumn{7}{l}{\emph{With lighting references}} \\
    Ground truth & 17.19 & 0.324 & 19.71 & 19.56 & 17.53 & 19.03 \\
    Predicted & 17.17 & 0.325 & 19.69 & 19.60 & 17.54 & 19.05 \\
    None & 17.13 & 0.330 & 19.22 & 19.14 & 17.23 & 18.63 \\
    \midrule
    \multicolumn{7}{l}{\emph{No reference}} \\
    Ground truth & 14.65 & 0.365 & 16.19 & 15.85 & 14.51 & 15.56 \\
    Predicted & 14.64 & 0.365 & 16.17 & 15.89 & 14.53 & 15.58 \\
    None & 14.58 & 0.370 & 15.80 & 15.52 & 14.28 & 15.25 \\
    \bottomrule
  \end{tabular}
\end{table}

\begin{table}[t]
  \centering
  \caption{\textbf{Predicted transient masks match the labels} (PSNR/LPIPS, scored as in Table~\ref{tab:main}). Same checkpoints; only the masks given to the sampler, the refiner, and the lighting references differ, and the scored pixels are the same. \emph{Labels}: shipped with MTGS, SegFormer-B2 pseudo-labels on Ithaca365.}
  \label{tab:predmask}
  \footnotesize
  \setlength{\tabcolsep}{4pt}
  \begin{tabular}{lcccc}
    \toprule
    & \multicolumn{2}{c}{MTGS} & \multicolumn{2}{c}{Ithaca365} \\
    \cmidrule(lr){2-3}\cmidrule(lr){4-5}
    Budget & Labels & Predicted & Labels & Predicted \\
    \midrule
    100k & 21.12/.232 & 21.13/.231 & 19.99/.225 & 19.98/.225 \\
    200k & 21.26/.220 & 21.24/.220 & 19.90/.218 & 19.90/.218 \\
    300k & 21.22/.216 & 21.22/.216 & 19.82/.218 & 19.81/.218 \\
    400k & 21.17/.214 & 21.15/.215 & 19.75/.219 & 19.75/.219 \\
    \midrule
    200k, no reference & 19.88/.235 & 19.88/.235 & 18.65/.249 & 18.65/.249 \\
    \bottomrule
  \end{tabular}
\end{table}

\paragraph{No masks on \bench.}
The real-data metrics exclude traffic pixels, so we repeat the comparison on \bench (Sec.~\ref{sec:carlabench}), where the traffic-free frame is the ground truth, and add a run without any transient information.
In that run transient pixels are treated as static in the sampler, and the residuals of the refiner and the lighting references are not masked; the weights are the same in every row and are not retrained.
Table~\ref{tab:carla_masks} shows that predicted masks match ground-truth masks here as well: Full PSNR and LPIPS are unchanged and Traffic PSNR differs by 0.02\,dB, although the transient head was never trained on this data.
Removing the masks shows up where it should, on the pixels of the traffic: Traffic PSNR drops by 0.43\,dB with lighting references and by 0.34\,dB without them, on 138 and 132 of the 173 splits, and by up to 7.9\,dB on single frames.
Full PSNR drops by only 0.04--0.06\,dB, because traffic covers about 5\,\% of the pixels of a frame (8\,\% at the 90th percentile).
Figure~\ref{fig:nomask} shows the ghosts and smears that vehicles and pedestrians can leave without masks, also on a real MTGS drive; over the 18 MTGS test splits, static PSNR changes by only 0.03\,dB without masks, so such ghosts are visible but barely scored on real data.
Without masks and references, \method stays 1.2\,dB above the best un-relit baseline in Traffic PSNR, against 1.5\,dB with masks.
On held-out frames the scored traffic was never in the input; the effect is larger on the input views, where it was (Sec.~\ref{sec:supp_carla_input}).
Table~\ref{tab:abl_masks} separates the sampler from the refiner.

\begin{table*}[t]
  \centering
  \caption{\textbf{\bench per location} (Full PSNR/Traffic PSNR, held-out frames; 200k Gaussians and lighting references for \method). \emph{E/M/H}: splits per difficulty level (9 per location in the first block, 5 in the others). Bold marks the best of all seven columns.}
  \label{tab:carla_locations}
  \footnotesize
  \setlength{\tabcolsep}{3pt}
  \resizebox{\textwidth}{!}{%
  \begin{tabular}{lcccccccccc}
    \toprule
    & & & \multicolumn{3}{c}{No image of the target drive} & \multicolumn{3}{c}{Relit with target-drive references} & \\
    \cmidrule(lr){4-6}\cmidrule(lr){7-9}
    Location & Town & E/M/H & ReSplat & MVSplat & DepthSplat & ReSplat & MVSplat & DepthSplat & \method \\
    \midrule
    Blend & 03 & 1/4/4 & 9.77/10.40 & 11.61/11.60 & 8.84/9.18 & 11.00/12.88 & 12.62/15.07 & 9.68/11.16 & \textbf{14.82}/\textbf{16.69} \\
    Bonsai & 05 & 2/3/4 & 12.72/12.92 & 12.72/13.11 & 11.81/11.73 & 13.50/13.63 & 13.63/14.00 & 12.40/12.27 & \textbf{17.06}/\textbf{17.72} \\
    Bus Station & 10 & 1/4/4 & 10.18/10.43 & 10.26/10.34 & 9.56/9.97 & 10.76/11.54 & 10.74/11.43 & 9.94/10.78 & \textbf{13.33}/\textbf{15.31} \\
    Getty & 10 & 1/4/4 & 10.31/10.34 & 10.83/10.89 & 10.00/9.94 & 11.86/11.85 & 12.57/12.43 & 11.49/11.18 & \textbf{14.22}/\textbf{15.14} \\
    Houses & 02 & 2/3/4 & 14.14/11.88 & 13.71/12.26 & 13.51/11.87 & 15.55/13.67 & 14.93/14.33 & 14.98/13.73 & \textbf{16.75}/\textbf{16.73} \\
    Mega Wide & 03 & 1/4/4 & 11.16/11.06 & 12.00/11.41 & 10.34/10.17 & 15.42/16.02 & 16.11/15.37 & 14.06/14.36 & \textbf{18.20}/\textbf{19.11} \\
    Tree Shade & 10 & 2/3/4 & \textbf{17.01}/13.42 & 15.88/13.99 & 16.59/12.78 & 16.84/13.20 & 15.80/13.82 & 16.45/12.56 & 16.89/\textbf{16.12} \\
    \midrule
    Scene 1 & 12 & 2/3/0 & 15.86/14.08 & 16.20/14.65 & 15.56/13.26 & 17.47/15.72 & 17.49/15.96 & 17.12/14.66 & \textbf{19.50}/\textbf{19.25} \\
    Scene 2 & 12 & 1/4/0 & 11.11/9.04 & 12.25/11.13 & 10.85/8.63 & 14.46/13.38 & 14.75/\textbf{16.43} & 13.70/12.75 & \textbf{16.41}/14.98 \\
    Scene 3 & 12 & 3/2/0 & 18.88/20.73 & 16.78/18.12 & 19.09/18.39 & 19.66/22.22 & 17.41/18.48 & 20.01/20.15 & \textbf{21.28}/\textbf{25.17} \\
    Scene 4 & 12 & 0/5/0 & 17.39/18.78 & 15.23/13.34 & 17.36/17.63 & 18.56/20.72 & 16.68/15.32 & 18.70/19.69 & \textbf{20.25}/\textbf{24.04} \\
    Scene 5 & 12 & 1/4/0 & 13.31/15.75 & 14.14/14.40 & 13.01/14.08 & 14.50/18.06 & 14.72/16.07 & 14.06/15.81 & \textbf{15.83}/\textbf{20.53} \\
    Scene 6 & 12 & 1/4/0 & 12.05/9.17 & 12.32/10.83 & 11.16/8.54 & 14.46/12.20 & 13.97/\textbf{13.95} & 13.09/11.26 & \textbf{16.03}/12.17 \\
    Scene 7 & 12 & 1/4/0 & 16.81/17.22 & 15.72/14.17 & 16.45/16.30 & 17.97/19.07 & 17.17/15.71 & 17.62/18.02 & \textbf{19.81}/\textbf{23.60} \\
    Scene 8 & 12 & 1/4/0 & 17.74/18.50 & 18.08/17.25 & 17.34/17.14 & 19.00/20.77 & 18.36/16.62 & 18.48/18.74 & \textbf{20.98}/\textbf{23.42} \\
    Scene 9 & 12 & 3/2/0 & 14.13/14.95 & 14.01/14.02 & 14.17/14.38 & 14.98/16.65 & 14.91/15.58 & 14.98/15.51 & \textbf{16.31}/\textbf{18.61} \\
    Scene 10 & 12 & 1/4/0 & 15.97/17.18 & 15.70/16.04 & 15.54/15.76 & 16.34/17.75 & 15.45/15.09 & 15.81/16.05 & \textbf{17.85}/\textbf{20.40} \\
    \midrule
    E01 & 10 & 5/0/0 & 16.65/16.11 & 14.49/15.18 & 16.61/15.53 & 17.15/16.56 & 15.23/15.61 & 17.10/15.86 & \textbf{17.19}/\textbf{20.22} \\
    E02 & 12 & 5/0/0 & 12.46/9.63 & 12.80/11.49 & 12.62/9.58 & 15.84/13.33 & 15.49/13.73 & 15.96/12.54 & \textbf{17.23}/\textbf{15.36} \\
    E03 & 13 & 5/0/0 & 15.82/16.86 & 15.32/12.69 & 15.90/14.92 & 16.39/17.49 & 15.95/12.76 & 16.65/15.51 & \textbf{18.12}/\textbf{21.18} \\
    E04 & 13 & 5/0/0 & 17.06/17.21 & 15.15/11.54 & 16.88/15.51 & 18.62/19.42 & 17.10/14.03 & 18.91/18.18 & \textbf{19.52}/\textbf{24.29} \\
    M01 & 10 & 0/5/0 & 14.55/17.49 & 13.52/14.17 & 14.09/17.16 & 14.57/18.20 & 13.69/14.68 & 14.09/17.96 & \textbf{15.50}/\textbf{20.71} \\
    M02 & 12 & 0/5/0 & 12.17/10.63 & 13.38/10.02 & 11.49/9.98 & 15.24/15.20 & 16.06/14.24 & 14.02/14.00 & \textbf{17.67}/\textbf{16.77} \\
    M03 & 13 & 0/5/0 & 17.21/18.82 & 16.16/15.74 & 17.13/17.72 & 17.50/19.30 & 16.39/15.77 & 17.52/18.14 & \textbf{18.38}/\textbf{22.61} \\
    M04 & 13 & 0/5/0 & 16.10/18.79 & 14.74/15.54 & 16.01/18.03 & 17.18/20.75 & 16.11/17.72 & 17.25/20.16 & \textbf{18.10}/\textbf{24.06} \\
    H01 & 10 & 0/0/5 & 14.69/17.91 & 13.95/15.12 & 14.41/16.33 & 14.82/18.03 & 14.02/15.29 & 14.41/16.18 & \textbf{15.98}/\textbf{22.09} \\
    H02 & 12 & 0/0/5 & 10.63/8.40 & 11.11/9.24 & 10.36/8.08 & 14.34/13.08 & 14.47/15.04 & 13.64/12.43 & \textbf{16.76}/\textbf{15.09} \\
    H03 & 13 & 0/0/5 & 15.27/18.50 & 14.95/15.47 & 14.90/17.43 & 16.27/19.79 & 15.45/16.29 & 15.97/18.53 & \textbf{17.22}/\textbf{21.09} \\
    H04 & 13 & 0/0/5 & 15.05/16.55 & 13.60/14.40 & 14.91/16.72 & 16.53/19.85 & 14.99/16.72 & 16.29/19.61 & \textbf{17.98}/\textbf{23.38} \\
    \bottomrule
  \end{tabular}}
\end{table*}

\paragraph{Remaining disagreements.}
Figure~\ref{fig:mask_failures} shows what remains.
We inspected the connected regions where one mask has almost no overlap with the other, over the context and target frames of the 48 test splits.
The predicted-only regions of at least 40 pixels fall on 23 distinct locations, and almost all are real vehicles or people that the labels miss; at most two are false positives, and both are small.
Among the label-only regions, the head misses some construction vehicles, while others are label errors on guardrails and hedges.

\subsection{Results per Location}
Table~\ref{tab:carla_locations} breaks the held-out results down by location, with masks from the transient head, and lists the splits per difficulty level.
Against the un-relit baselines, \method has the highest Traffic PSNR at every location and the highest Full PSNR at 28 of 29; at Tree Shade, ReSplat leads slightly in Full PSNR (17.01 vs.\ 16.89\,dB).
Against the relit baselines, \method has the highest Full PSNR at all 29 locations and the highest Traffic PSNR at 27; relit MVSplat leads in Traffic PSNR at Scene 2 (16.43 vs.\ 14.98\,dB) and Scene 6 (13.95 vs.\ 12.17\,dB).

\section{Geometry and Refinement Ablations}
\label{sec:supp_ablations}
Calibrated unprojection matters more than a more accurate depth map, and accurate cameras matter throughout: registering the Ithaca365 drives to each other gains 0.54\,dB without retraining, and predicted cameras cost 3.8--4.6\,dB.
A head trained with the sampler in the loop beats merging in LPIPS, one head and refiner serve every budget, the fine-tuned refiner beats a frozen ReSplat refiner, two steps give the best LPIPS, and doubling the context views helps without retraining.

\subsection{Calibration and Backbone}
\label{sec:supp_backbone}
\paragraph{Camera conditioning.}
The backbone receives images only.
Known intrinsics and poses enter solely through the unprojection (Sec.~\ref{sec:initialization}) and the renderer; the transformer is not told the camera geometry.
We also tested feeding the intrinsics to the transformer as additional input tokens: the unit camera-frame ray of every pixel center (computed from the cropped $K$, principal point included) is patchified by a zero-initialized $14{\times}14$ convolution and added to the patch tokens before the attention blocks.
This slightly improves the predicted depth but gives worse renders, so we do not use it (Table~\ref{tab:backbone}).

\paragraph{Inverse-depth supervision.}
We also trained the backbone with the inverse-depth loss of DeSiRe-GS~\cite{peng2025desire}, $|1/Z_i-1/(s\,D_i)|$ with weight $4\,e^{-t/8000}$ at step $t$ and LiDAR returns on transient pixels excluded, which weights near geometry far more than distant geometry.
It lowered the depth error close to the camera, though not overall, but made the final renders slightly worse, so we keep the log loss (Table~\ref{tab:backbone}).

\paragraph{Results.}
Injecting ground-truth camera parameters into the unprojection of the pretrained Wild3R improves it by 1.49\,dB PSNR, 0.092 SSIM, and 0.042 LPIPS on average over 34 cross-traversal splits of the six MTGS blocks (better PSNR in 33 of 34), compared with placing each predicted view at its ground-truth pose after the fact.
The gain comes from the intrinsics, whose predicted focal length is off by 6\,\% horizontally and 4\,\% vertically on average (up to 14\,\% in a scene); injecting the poses alone adds 0.02\,dB.
This motivates using calibrated unprojection throughout, and treating camera consistency as a prerequisite for interpreting rendered feedback.

\begin{table}[t]
  \centering
  \caption{\textbf{Backbone ablation} (PSNR/LPIPS, scored as in Table~\ref{tab:main}). \emph{Dense}: the backbone's 1.6M Gaussians; \emph{Full pipeline}: compact head, refiner, and appearance branch trained on it, with lighting references, in the stock-pass adapter version (Sec.~\ref{sec:supp_variants}). Depth: median relative error against LiDAR on 168 Ithaca365 context views, overall and within 10\,m.}
  \label{tab:backbone}
  \footnotesize
  \setlength{\tabcolsep}{2pt}
  \resizebox{\linewidth}{!}{%
  \begin{tabular}{lcccccc}
    \toprule
    & \multicolumn{2}{c}{Dense 1.6M} & \multicolumn{2}{c}{Full pipeline, 200k} & \multicolumn{2}{c}{Depth abs-rel} \\
    \cmidrule(lr){2-3}\cmidrule(lr){4-5}\cmidrule(lr){6-7}
    Backbone & MTGS & Ithaca365 & MTGS & Ithaca365 & all & ${<}10$\,m \\
    \midrule
    \textbf{Ours (log depth)} & 19.18/.227 & 18.35/.255 & \textbf{21.18}/.219 & \textbf{19.93}/\textbf{.215} & .032 & .028 \\
    Inverse depth & \textbf{19.39}/\textbf{.215} & \textbf{18.43}/\textbf{.240} & 20.96/\textbf{.218} & 19.85/.221 & .032 & .026 \\
    \;+ intrinsics tokens & 19.20/.227 & 18.40/.246 & -- & -- & \textbf{.029} & \textbf{.020} \\
    \bottomrule
  \end{tabular}}
\end{table}

Table~\ref{tab:backbone} compares our backbone with one trained with the inverse-depth LiDAR loss of DeSiRe-GS~\cite{peng2025desire}, and with a variant that additionally receives the intrinsics as input tokens.
Inverse depth puts most weight on near geometry and lowers the depth error there (0.026 vs.\ 0.028 within 10\,m), but not overall (0.032 for both), and it is less accurate beyond 40\,m (0.085 vs.\ 0.064).
Its dense renders are better (MTGS 19.39 vs.\ 19.18\,dB, LPIPS 0.215 vs.\ 0.227; Ithaca365 18.43 vs.\ 18.35\,dB), yet the full pipeline built on it is 0.22\,dB worse on MTGS and 0.08\,dB worse on Ithaca365, with equal LPIPS on MTGS (0.218 vs.\ 0.219) and higher LPIPS on Ithaca365 (0.221 vs.\ 0.215).
The advantage of the dense model does not survive compaction: with the compact head trained on identical data, both backbones give the same PSNR before refinement (MTGS 19.21 vs.\ 19.22\,dB at 200k, within the 0.14\,dB spread between two training seeds of the same head), and the gap opens during refinement.
A second seed of our compact head, refined with exactly the same recipe as the inverse-depth pipeline, reproduces our result to within 0.04\,dB, so the gap is not explained by seed or training length.
Only opposing-direction synthesis gains from the inverse-depth backbone (MTGS 20.58 vs.\ 20.55\,dB, LPIPS 0.240 vs.\ 0.250), and we keep the log loss.
Intrinsics tokens reduce the depth error further (0.029 overall, 0.020 within 10\,m) but give worse dense renders (19.20 vs.\ 19.39\,dB on MTGS), so a more accurate depth map does not translate into better novel views either.
Feeding the ground-truth intrinsics only to the unprojection is sufficient.

\subsection{Cross-Drive Registration of Ithaca365}
\label{sec:supp_ithaca_reg}
\begin{figure*}[t]
  \centering
  \includegraphics[width=\textwidth]{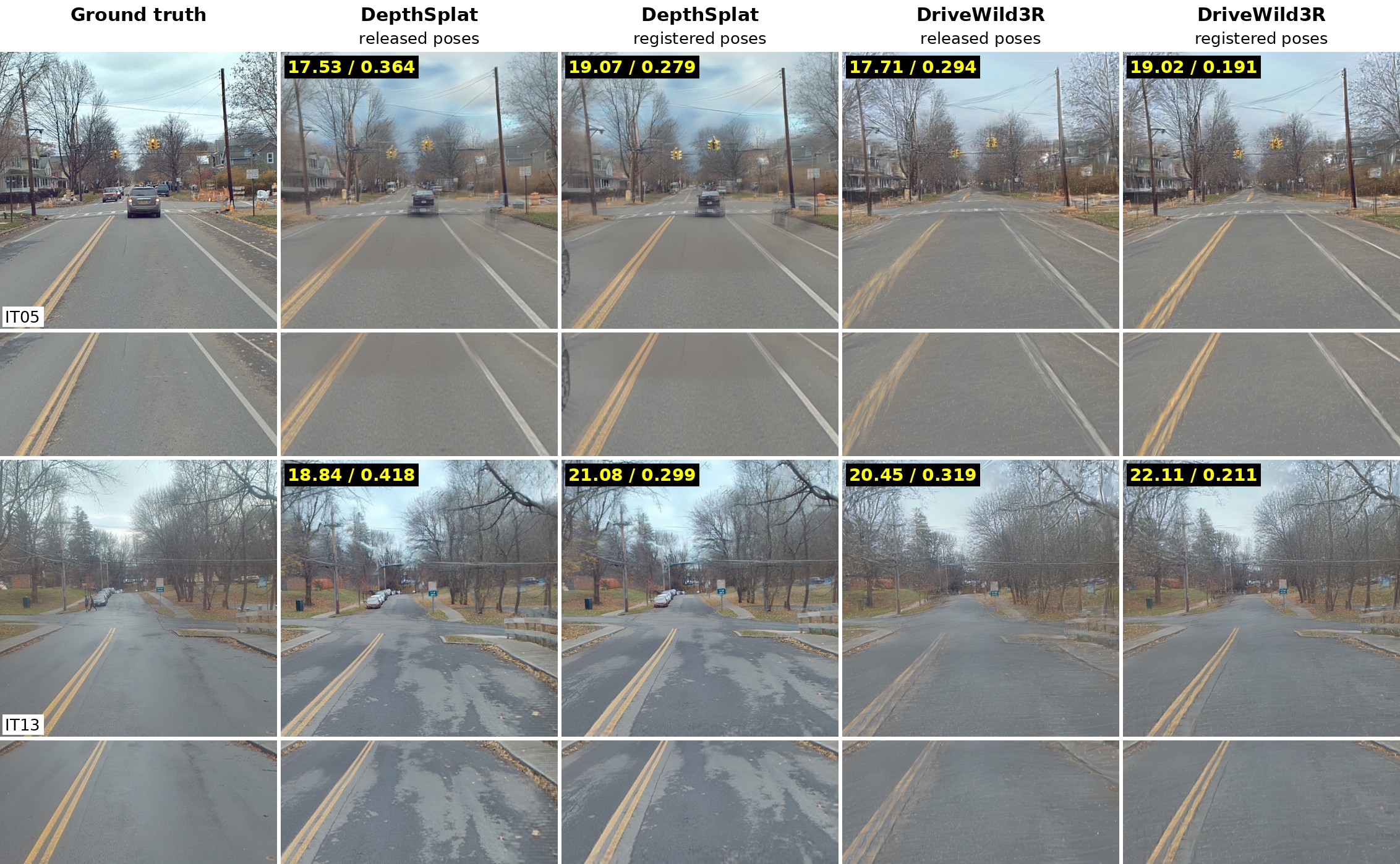}
  \caption{\textbf{Released vs.\ registered Ithaca365 poses.} Two frames of the blocks with the largest cross-drive offsets (IT05, IT13), rendered by DepthSplat and \method (200k, lighting references) with the released and with the LiDAR-registered poses; below each frame a zoom on the road. With the released poses, \method doubles the lane markings and DepthSplat misplaces them. Static PSNR/LPIPS per frame.}
  \label{fig:ithaca_reg}
\end{figure*}

The drives of Ithaca365 are not registered to each other.
When the LiDAR sweeps of two drives are placed with the released poses and aligned by point-to-plane ICP, sweeps of the same drive agree to about 3\,cm, but sweeps of different drives are offset by 7--29\,cm (median per test block; 90th percentile up to 0.9\,m).
MTGS ships traversals that its authors registered to each other, so MTGS is not affected.
We register every drive to the drive with the most sweeps by one rigid ICP of their merged sweeps, which reduces the offsets to 4--7\,cm, and apply the correction to all cameras of that drive; no model is retrained.
Table~\ref{tab:ithaca_reg} compares all methods with the released and the registered poses.
Every method that places its context views with the released poses gains 0.2--0.6\,dB, while Wild3R, which predicts its own cameras, is unchanged, and the ranking is the same in both settings.
The offsets affect the methods differently (Fig.~\ref{fig:ithaca_reg}): \method fuses the drives into one set of Gaussians, so a surface seen by two drives is placed twice and thin markings double, while DepthSplat and ReSplat render sharp but misplaced geometry.
The tables of the main paper use the released poses; the registered poses are used here, in Figs.~\ref{fig:supp_qual_ithaca} and~\ref{fig:shadows}, and in the predicted-camera experiment (Sec.~\ref{sec:supp_posefree}).

\begin{table}[t]
  \centering
  \caption{\textbf{Ithaca365 with released and LiDAR-registered poses} (30 splits, PSNR/LPIPS, scored as in Table~\ref{tab:main}; \method at 200k). Weights are unchanged.}
  \label{tab:ithaca_reg}
  \footnotesize
  \setlength{\tabcolsep}{3pt}
  \begin{tabular}{lccc}
    \toprule
    Method & Released & Registered & $\Delta$ \\
    \midrule
    Wild3R & 16.00/.413 & 16.01/.411 & +0.02/$-$.002 \\
    MVSplat & 18.12/.334 & 18.34/.325 & +0.22/$-$.009 \\
    DepthSplat & 18.53/.268 & 19.13/.241 & +0.60/$-$.027 \\
    ReSplat & 19.04/.333 & 19.51/.310 & +0.47/$-$.022 \\
    \method, no reference & 18.65/.249 & 19.21/.220 & +0.56/$-$.029 \\
    \method, lighting ref. & \textbf{19.90}/\textbf{.218} & \textbf{20.44}/\textbf{.189} & +0.54/$-$.029 \\
    \bottomrule
  \end{tabular}
\end{table}

\subsection{Predicted Cameras}
\label{sec:supp_posefree}
\begin{figure}[t]
  \centering
  \includegraphics[width=\linewidth]{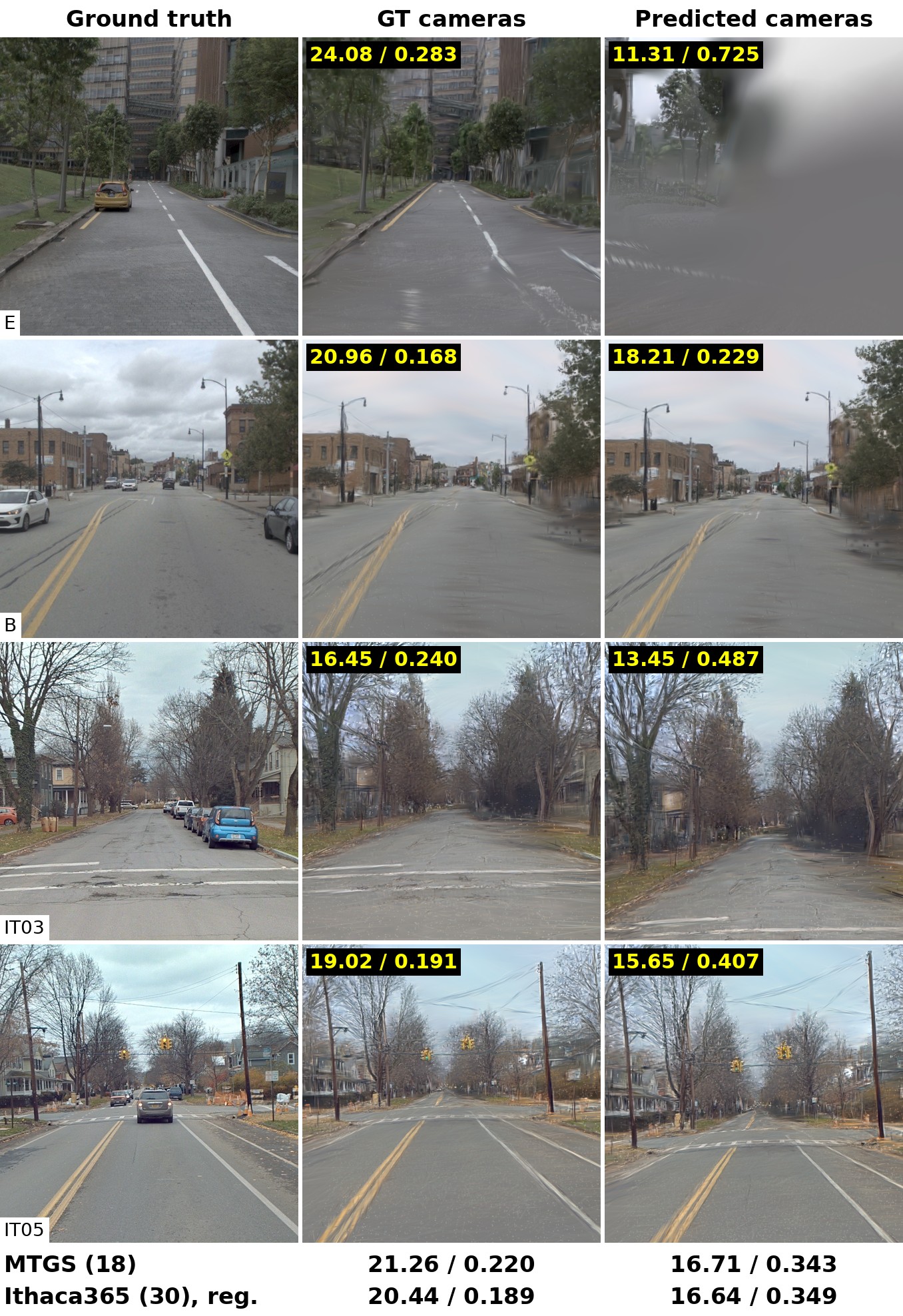}
  \caption{\textbf{Predicted cameras.} \method at 200k with lighting references, with ground-truth input cameras (as in Table~\ref{tab:main}; on Ithaca365 the LiDAR-registered poses of Sec.~\ref{sec:supp_ithaca_reg}) and with the cameras predicted by its backbone, aligned by one similarity transform on the context views; target frames are rendered from their ground-truth (Ithaca365: registered) poses. Rows drawn at random with a fixed seed (row 1 has opposing-direction context). Static PSNR/LPIPS per frame; the footer gives the means over all MTGS and Ithaca365 splits.}
  \label{fig:posefree}
\end{figure}
We also run \method without any camera calibration (Fig.~\ref{fig:posefree}).
The camera head of our backbone predicts poses and intrinsics for all inputs, the context views and the lighting references, in one pass, and one similarity transform fitted on the context views places them in the metric frame, as for Wild3R in Table~\ref{tab:main}; the scored frames are rendered from their ground-truth poses.
On Ithaca365, the ground truth is the LiDAR-registered poses (Sec.~\ref{sec:supp_ithaca_reg}).
Nothing is retrained.
PSNR drops from 21.26 to 16.71\,dB on MTGS and from 20.44 (19.90 with the released poses, Table~\ref{tab:main}) to 16.64\,dB on Ithaca365, and LPIPS rises from 0.220 to 0.343 and from 0.189 to 0.349.
This remains above Wild3R (13.56 and 16.01\,dB; LPIPS 0.423 and 0.411), which uses the same camera protocol, but below ReSplat with ground-truth cameras.
With context from one other drive or from the same direction, the intrinsics cost more than the poses: predicted intrinsics with ground-truth poses lose 2.6\,dB on MTGS and 3.4\,dB on Ithaca365, while predicted poses with ground-truth intrinsics lose 0.8 and 1.6\,dB.
The camera head places the principal point at the image center, 8\,px from the true one on MTGS and 28\,px on Ithaca365, and misestimates the focal length by 2\,\% on MTGS and 11\,\% on Ithaca365 (median).
With opposing-direction context, the camera head does not register the two directions to each other (median rotation error $64^\circ$ after alignment, against $0.13$--$0.17^\circ$ otherwise), and PSNR falls from 20.64 to 12.63\,dB.

\subsection{Sampling and the Gaussian Head}
\label{sec:supp_sampling}

\begin{table}[t]
  \centering
  \caption{\textbf{How to fill the holes of sampling} (100k Gaussians, 18 MTGS splits, odd frames, no refiner or relighting). All rows keep the same entropy-sampled pixels; discard and merge use the dense Gaussian head, ours is trained with the sampler in the loop.}
  \label{tab:sampling}
  \footnotesize
  \setlength{\tabcolsep}{3pt}
  \begin{tabular}{lcccc}
    \toprule
    100k Gaussians, no refiner & PSNR & SSIM & LPIPS \\
    \midrule
    Sample, discard the rest (dense head) & 10.51 & 0.304 & 0.530 \\
    Sample, merge the rest into neighbors & 18.75 & \textbf{0.636} & 0.359 \\
    \textbf{Sample + fine-tuned Gaussian head} & \textbf{18.88} & 0.616 & \textbf{0.274} \\
    \textcolor{gray}{All pixels (dense, 1.6M)} & \textcolor{gray}{19.18} & \textcolor{gray}{0.623} & \textcolor{gray}{0.227} \\
    \bottomrule
  \end{tabular}
\end{table}

\begin{figure*}[t]
  \centering
  \includegraphics[width=\textwidth]{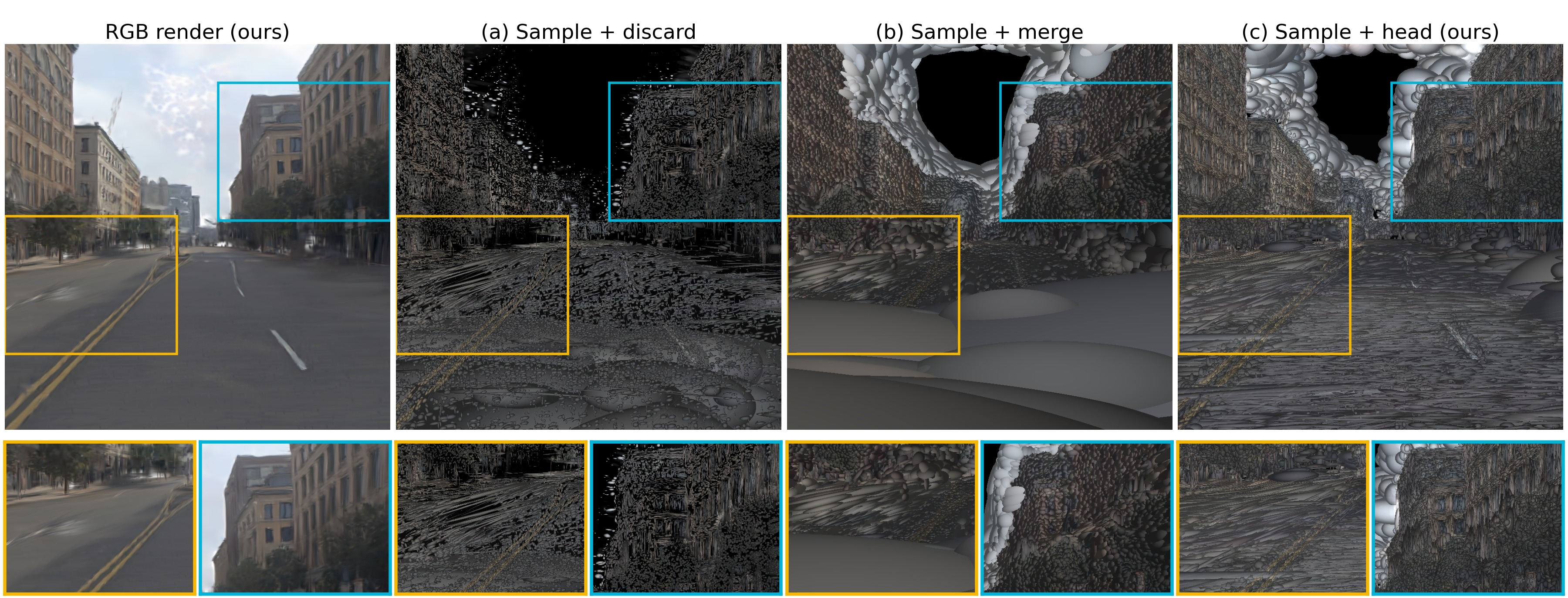}
  \caption{\textbf{Gaussians from the three ways of reducing a cloud}, shown as opaque ellipsoids (100k Gaussians from the same sampled pixels, single-traversal MTGS scene, no refinement; camera on the context drive; left: RGB render of our head; bottom: road-marking and facade crops). (a) Discarding leaves sub-pixel Gaussians with holes. (b) Merging makes the survivors large and round on screen and creates meters-long Gaussians below the road. (c) Our head stretches them along image structure: flat on the road, thin along lane markings and facade edges, very large in the sky.}
  \label{fig:ellipsoids}
\end{figure*}

Our first attempt at reducing the cloud merged instead of predicting: the Gaussians of the kept pixels stayed, and every discarded Gaussian was folded into the nearest kept Gaussian of its own view (moment matching~\cite{kerbl2024hierarchical}, union opacity, averaged color), with no learned head.
We compare this to the head of Sec.~\ref{sec:compact} in Table~\ref{tab:sampling}.
Discarding without a head collapses (10.51\,dB): the dense Gaussians of the kept pixels cover a median of only 0.32 pixels of screen, and a cloud of sub-pixel Gaussians cannot lose 94\,\% of its members.
Merging fills the holes by enlarging the survivors: in the final pipeline at 100k, their median screen size is 1.34\,px against 0.32\,px for the same Gaussians before merging, and 72\,\% of them are nearly opaque (opacity above 0.98).
It comes within 0.43\,dB of the dense cloud and has the best SSIM, but a much worse LPIPS (0.359 vs.\ 0.227).
The fine-tuned head ties merging in PSNR (18.88 vs.\ 18.75\,dB, better on 11 of 18 splits) and is 0.085 better in LPIPS (on 17 of 18 splits).
Without refinement, neither compact cloud reaches the dense one: the head is 0.30\,dB and 0.047 LPIPS behind it with a sixteenth of the Gaussians. Refinement recovers the PSNR (19.74\,dB at 100k without relighting, against 19.18 for dense), and with lighting references the full pipeline has a lower LPIPS than the dense cloud from 200k Gaussians on (Table~\ref{tab:budgets}); without them its LPIPS stays higher at every budget on MTGS.
The two differ in what they do to the Gaussians (Fig.~\ref{fig:ellipsoids}; final pipeline, 100k, 18 MTGS splits, measured in each Gaussian's own context view). Merging makes them large, opaque, and round on screen (median anisotropy of the projected footprint 1.6); their elongation in 3D (3.6:1) points along the viewing ray for 93\,\% of the elongated ones, the trace of moment matching over neighbors at different depths. The head grows them less (median 0.86\,px) but stretches them across the image (median anisotropy 4.8, 62\,\% at least 3:1), along road markings and edges.
Unlike merging, the head also leaves every Gaussian with its own source pixel and view, which the lighting branch relies on.

\paragraph{Why compact Gaussians.}
\label{sec:supp_why_compact}
Refining all 1.6M Gaussians is expensive.
A one-step refiner trained at 1.6M peaks at 62.5\,GiB and takes 6\,h, and two unrolled steps run out of memory.
A two-step refiner trained at 400k takes 20\,GiB and 2\,h.
At inference, four refinement steps peak at 40.3\,GiB at 1.6M and at 11.5\,GiB at 400k; our pipeline with all models resident peaks at 12.4\,GiB at 400k (Table~\ref{tab:runtime}).

\paragraph{Why entropy sampling.}
In the ablation of SparseSplat~\cite{zhang2026sparsesplat}, entropy-based sampling outperforms both uniform random and image-gradient (Laplacian) sampling at 40k Gaussians, and we follow it.
It also suits our head: with entropy sampling, the head's ellipsoids align better with the street and become larger on low-texture road and denser on detailed regions (Fig.~\ref{fig:ellipsoids}).
Down-weighting sky and transient pixels by a factor of ten adds a further 0.27--0.30\,dB at 100k and 200k Gaussians before refinement and removes Gaussians from vehicles in favor of static content.
We did not separate the contributions of the sky and the transient weight.

\subsection{Gaussian Budget}

\begin{table}[t]
  \centering
  \caption{\textbf{Every budget, with and without lighting references} (PSNR/SSIM/LPIPS, one shared head and refiner, scored as in Table~\ref{tab:main}). The 300k budget was never seen in training. Bold marks the best budget per column; the dense row has no reference and is bold where it beats every no-reference budget.}
  \label{tab:budgets}
  \footnotesize
  \setlength{\tabcolsep}{2.5pt}
  \resizebox{\linewidth}{!}{%
  \begin{tabular}{lcccc}
    \toprule
    & \multicolumn{2}{c}{MTGS (18)} & \multicolumn{2}{c}{Ithaca365 (30)} \\
    \cmidrule(lr){2-3}\cmidrule(lr){4-5}
    Budget & No reference & References & No reference & References \\
    \midrule
    100k & 19.74/.624/.247 & 21.12/.646/.232 & \textbf{18.73}/\textbf{.424}/.255 & \textbf{19.99}/\textbf{.461}/.225 \\
    200k & 19.88/.626/.235 & \textbf{21.26}/\textbf{.649}/.220 & 18.65/.419/.249 & 19.90/.456/\textbf{.218} \\
    300k & 19.94/.626/.231 & 21.22/.647/.216 & 18.59/.417/\textbf{.248} & 19.82/.453/\textbf{.218} \\
    400k & \textbf{19.97}/\textbf{.627}/.228 & 21.17/.646/\textbf{.214} & 18.53/.415/\textbf{.248} & 19.75/.450/.219 \\
    \midrule
    Dense, 1.6M & \multicolumn{2}{c}{19.18/.623/\textbf{.227} (no refiner)} & \multicolumn{2}{c}{18.35/.427/.255 (no refiner)} \\
    \bottomrule
  \end{tabular}}
\end{table}

\begin{figure}[t]
  \centering
  \includegraphics[width=\linewidth]{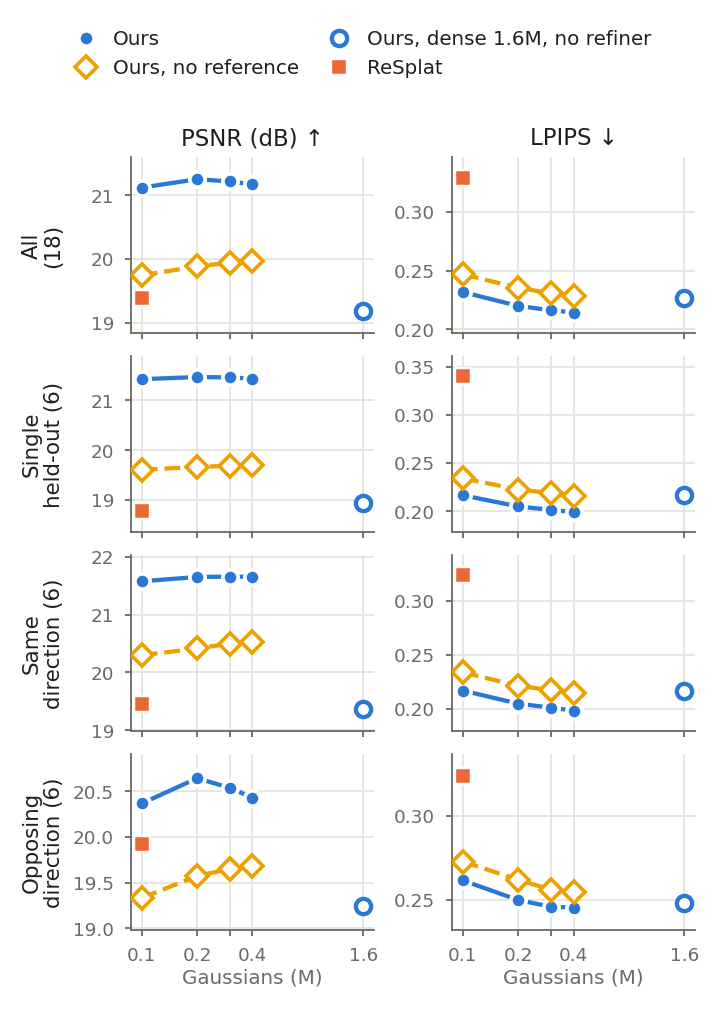}
  \caption{\textbf{Gaussian budget on MTGS}, overall and per setting (rows; splits in parentheses), with one shared head and refiner. \emph{No reference}: no image of the target drive. \emph{Dense}: the adapted backbone's 1.6M Gaussians; ReSplat uses 100k. Scored as in Table~\ref{tab:main}; Wild3R lies below every axis.}
  \label{fig:budget}
\end{figure}

Table~\ref{tab:budgets} and Fig.~\ref{fig:budget} sweep the budget.
With references, PSNR is nearly flat on MTGS, moving between 21.12 and 21.26\,dB, while LPIPS improves steadily with the budget, from 0.232 to 0.214, in every setting.
Ithaca365 follows the same pattern: PSNR falls slightly from 19.99 to 19.75\,dB, and LPIPS drops from 0.225 to 0.218 at 200k and then stays flat.
Without references, PSNR rises slightly with the budget on MTGS (19.74 to 19.97\,dB) and falls slightly on Ithaca365 (18.73 to 18.53\,dB), and LPIPS improves on both (0.247 to 0.228 and 0.255 to 0.248).
Additional Gaussians thus add fine detail rather than correcting color or placement, which a per-pixel error barely registers.
Compared with the dense 1.6M initialization of the same backbone, compaction, refinement, and relighting gain 1.9--2.1\,dB on MTGS and 1.4--1.6\,dB on Ithaca365, and with references every budget from 200k, an eighth of the dense count, also has the lower LPIPS (MTGS 0.220 vs.\ 0.227, Ithaca365 0.218 vs.\ 0.255); at 100k its MTGS LPIPS is still slightly higher (0.232).
Without a reference, PSNR stays above the dense cloud at every budget (by 0.56--0.79\,dB on MTGS and 0.18--0.39\,dB on Ithaca365), while LPIPS is lower only on Ithaca365 from 200k on (0.248--0.249 vs.\ 0.255) and stays higher on MTGS at every budget (0.228--0.247 vs.\ 0.227).
A compact budget therefore sacrifices little: 100k is within 0.14\,dB of the best PSNR on MTGS and the best on Ithaca365.

\subsection{One Model for Every Budget}
\label{sec:supp_shared}

The head and the refiner are shared across budgets (Sec.~\ref{sec:compact}).
We compared them with models trained for a single budget on two earlier pipeline versions (Table~\ref{tab:supp_variants}), over the same 18 MTGS and 30 Ithaca365 splits.
Heads trained for a single budget, in the pre-lighting pipeline, gain at most 0.006 LPIPS before refinement (0.002--0.003 on MTGS), and after refinement they tie with the shared head or fall below it (MTGS at 200k: 20.01 vs.\ 20.11\,dB; at 400k: 19.99 vs.\ 20.19\,dB).
Refiners trained for a single budget, in the reference-slot pipeline at 100k and 400k, are 0.07--0.27\,dB better in PSNR on MTGS; on Ithaca365 they are 0.05--0.06\,dB worse at 100k and 0.13--0.17\,dB better at 400k; LPIPS differs by at most 0.006, in favor of the shared refiner at 100k and of the single-budget ones by at most 0.003 at 400k.
In exchange, the shared models accept any number of Gaussians without retraining.
The 300k budget in Table~\ref{tab:budgets} was never seen in training, yet it is as good as the trained budgets (21.22\,dB and SSIM 0.647 on MTGS, vs.\ 21.26\,dB and 0.649 at 200k; SSIM 0.453 on Ithaca365).

\subsection{Refiner Transfer}
\label{sec:supp_transfer}

\begin{table}[t]
  \centering
  \caption{\textbf{Refiner transfer}. Every row refines the same compact cloud; only the refiner weights differ. No row uses target-drive information, so the last row equals \emph{no reference} in Table~\ref{tab:main}.}
  \label{tab:transfer}
  \footnotesize
  \setlength{\tabcolsep}{3pt}
  \resizebox{\linewidth}{!}{%
  \begin{tabular}{lcccccc}
    \toprule
    & \multicolumn{3}{c}{MTGS (18)} & \multicolumn{3}{c}{Ithaca365 (30)} \\
    \cmidrule(lr){2-4}\cmidrule(lr){5-7}
    Refiner (200k) & PSNR & SSIM & LPIPS & PSNR & SSIM & LPIPS \\
    \midrule
    None (compact head) & 19.13 & 0.618 & 0.262 & 18.32 & 0.426 & 0.277 \\
    Frozen ReSplat, 2 steps & 19.40 & 0.600 & 0.294 & 18.48 & \textbf{0.429} & 0.320 \\
    Frozen ReSplat, 4 steps & 19.17 & 0.594 & 0.313 & 18.54 & 0.428 & 0.337 \\
    \textbf{Fine-tuned (ours), 2 steps} & \textbf{19.88} & \textbf{0.626} & \textbf{0.235} & \textbf{18.65} & 0.419 & \textbf{0.249} \\
    \bottomrule
  \end{tabular}}
\end{table}

Table~\ref{tab:transfer} applies the pretrained ReSplat refiner, frozen, to the Gaussians of our compact head.
It raises PSNR a little (MTGS $+0.27$\,dB at two steps, $+0.03$\,dB at its native four; Ithaca365 $+0.16$ and $+0.23$\,dB) but makes LPIPS worse on all but one of the 48 splits ($+0.03$ to $+0.06$), and SSIM drops on MTGS.
The fine-tuned refiner improves all three on MTGS ($+0.75$\,dB, $-0.027$ LPIPS, lower LPIPS on all 18 splits) and PSNR and LPIPS on Ithaca365 ($+0.34$\,dB on 28 of 30 splits, $-0.028$ LPIPS).
The same holds at 100k (frozen: $+0.28$ to $+0.34$\,dB, LPIPS $+0.028$ to $+0.048$; fine-tuned: $+0.93$\,dB / $-0.028$ on MTGS, $+0.37$\,dB / $-0.033$ on Ithaca365, better on every Ithaca365 split).
Pretrained updates thus do not carry over to our setting: the frozen refiner trades perceptual quality for a small fidelity gain, while the fine-tuned one improves both.
We do not separate the causes; the compact Gaussians differ from ReSplat's in density, size and placement, driving scenes differ from its DL3DV training scenes, and the condition features come from our adapter rather than from ReSplat's own encoder.

\subsection{Refinement Steps}
\label{sec:supp_steps}

\begin{table}[t]
  \centering
  \caption{\textbf{Number of refinement steps} (PSNR/LPIPS, final pipeline, 200k, far reference, all target frames scored; Table~\ref{tab:supp_variants}). \emph{Trained}: steps $N$ in training; \emph{Run}: steps at test time. $N$ geometry steps are followed by $N$ color steps.}
  \label{tab:steps}
  \footnotesize
  \setlength{\tabcolsep}{2.5pt}
  \begin{tabular}{llccc}
    \toprule
    & & \multicolumn{2}{c}{MTGS} & Ithaca365 \\
    \cmidrule(lr){3-4}\cmidrule(lr){5-5}
    Trained & Run & all & Same direction & all \\
    \midrule
    2 steps & 1 & 20.23/.247 & 20.56/.235 & 19.13/.250 \\
    2 steps & \textbf{2} & 20.54/\textbf{.233} & 20.93/\textbf{.219} & \textbf{19.32}/\textbf{.239} \\
    2 steps & 3 & 20.63/.237 & 20.97/.222 & 19.25/.247 \\
    2 steps & 4 & \textbf{20.69}/.247 & \textbf{21.05}/.230 & 19.28/.259 \\
    \midrule
    4 steps & 2 & 19.83/.258 & 20.16/.242 & 19.05/.258 \\
    4 steps & 3 & 20.11/.252 & 20.49/.233 & 19.11/.254 \\
    4 steps & 4 & 20.29/.251 & 20.73/.229 & 19.22/.253 \\
    \bottomrule
  \end{tabular}
\end{table}

Table~\ref{tab:steps} varies the number of recurrent steps $N$, with the full lighting-aware refiner (far reference).
\emph{Running more steps than trained.}
A refiner trained for two steps has its best LPIPS at two steps on both datasets, and its best PSNR there on Ithaca365.
One step leaves error uncorrected (20.23\,dB / 0.247 on MTGS); three and four steps raise LPIPS (0.237, 0.247 on MTGS; 0.247, 0.259 on Ithaca365).
On MTGS, PSNR rises slightly with more steps (20.54 $\to$ 20.69\,dB at four steps, in every setting), so the extra steps trade perceptual quality for a small fidelity gain there; on Ithaca365 they lose both.
We did not test the cause; our guess is that the refiner learns a correction sized for two moves, so further moves overshoot.
The same trend appears for the refiner without color steps (0.241 $\to$ 0.245 $\to$ 0.255 LPIPS at two, three, four steps) and for the adjacent-reference refiner (0.223 $\to$ 0.227 $\to$ 0.244).
\emph{Training for more steps.}
Four steps are what ReSplat uses, so we also trained a four-step refiner (Sec.~\ref{sec:supp_losses}), backpropagating through all eight passes at a peak of about 85\,GB.
It is worse than the two-step refiner at every setting: at four steps it reaches 20.29\,dB / 0.251 on MTGS and 19.22\,dB / 0.253 on Ithaca365, against 20.54\,dB / 0.233 and 19.32\,dB / 0.239 for two steps, and the gap persists at 400k (20.26\,dB / 0.249 vs.\ 20.60\,dB / 0.227 on MTGS).
Its validation LPIPS was still falling when training ended (0.349 after 4k steps, against 0.326 for the two-step run).
We therefore use $N{=}2$ throughout.
Longer training or stopping gradients between steps may change this and is left for future work.

\subsection{Geometry of the Refined Gaussians}
\label{sec:supp_refined_depth}
\begin{table}[t]
  \centering
  \caption{\textbf{Depth of the rendered Gaussians on held-out traversals} (median relative error against LiDAR, before $\rightarrow$ after refinement; protocol in Sec.~\ref{sec:supp_metrics}). 18 MTGS and 28 Ithaca365 splits; two IT08 splits have no LiDAR on the scored frames.}
  \label{tab:refined_depth}
  \footnotesize
  \begin{tabular}{lcc}
    \toprule
    & MTGS & Ithaca365 \\
    \midrule
    ReSplat & .185 $\rightarrow$ .170 & .102 $\rightarrow$ .107 \\
    \method 100k & .092 $\rightarrow$ .102 & .092 $\rightarrow$ .139 \\
    \method 200k & .100 $\rightarrow$ .092 & .082 $\rightarrow$ .095 \\
    \method 400k & .118 $\rightarrow$ .113 & .087 $\rightarrow$ .083 \\
    \bottomrule
  \end{tabular}
\end{table}
Table~\ref{tab:refined_depth} scores the depth rendered from the Gaussians at the scored frames of the held-out traversal.
On MTGS, ReSplat's mean is dominated by block F (Table~\ref{tab:supp_manifest}), where its depth collapses to 0.3--0.6 times the LiDAR depth; without block F, both methods reach about 0.09 at 200k.
On Ithaca365, ReSplat has the lower error at 100k, its own budget (0.107 vs.\ 0.139), and ours is lower at 200k and 400k (0.095 and 0.083).
The refiner is trained with image losses only.
It increases the depth error on Ithaca365 at 100k (on 26 of 28 splits) and slightly at 200k (+0.013; 95\,\% bootstrap interval over splits +0.002 to +0.025); every other change in the table lies within its interval, so refinement produces no measurable change in depth there.

\subsection{Number of Context Views}
\label{sec:supp_views}

\begin{table*}[t]
  \centering
  \caption{\textbf{Number of context views} (PSNR/LPIPS, odd frames scored as in Table~\ref{tab:main}). 50\,m windows with the same query frames at both view counts; \method at a fixed total budget of 200k Gaussians, trained with 8 views only and not retrained. ReSplat receives no target-drive image and is not relit, so it compares with \emph{\method, no reference}. $^\dagger$Relative to the 8-view checkpoint at 8 views.}
  \label{tab:views}
  \footnotesize
  \setlength{\tabcolsep}{2.5pt}
  \begin{tabular}{lcccccc}
    \toprule
    & \multicolumn{3}{c}{MTGS (18 splits)} & \multicolumn{3}{c}{Ithaca365 (10 splits)} \\
    \cmidrule(lr){2-4}\cmidrule(lr){5-7}
    Method & 8 views & 16 views & $\Delta$PSNR & 8 views & 16 views & $\Delta$PSNR \\
    \midrule
    ReSplat, 8-view ckpt & 19.13/.333 & 19.33/.327 & $+0.20$ & 17.90/.364 & 17.90/.373 & $-0.00$ \\
    ReSplat, 16-view ckpt & -- & 19.62/.325 & $+0.49^\dagger$ & -- & 18.17/.373 & $+0.27^\dagger$ \\
    \method, no reference & 19.43/.249 & 19.69/.243 & $+0.27$ & 17.49/.264 & 17.61/.257 & $+0.12$ \\
    \method & \textbf{20.92}/\textbf{.234} & \textbf{21.23}/\textbf{.229} & $+0.31$ & \textbf{18.16}/\textbf{.242} & \textbf{18.31}/\textbf{.236} & $+0.14$ \\
    \bottomrule
  \end{tabular}
\end{table*}

Table~\ref{tab:views} doubles the context from 8 to 16 views at a fixed total of 200k Gaussians.
We use 50\,m windows (the six MTGS test blocks re-cut to 50\,m, and three windows of Ithaca365 IT03 and two of IT05, each with the settings of Sec.~\ref{sec:protocol}), because 8 views already cover the 20\,m MTGS segments.
Every stage of \method was trained with 8 views and none is retrained, so this evaluates \method zero-shot.
The extra views still add 0.31\,dB on MTGS and 0.14\,dB on Ithaca365, with a small LPIPS gain (0.005--0.006).
The gain needs budget: at 100k it is 0.20 and 0.07\,dB, and at 400k it is 0.37 and 0.13\,dB.
On MTGS most of it comes from opposing-direction context (${+}0.57$\,dB), where 8 views leave the most surfaces uncovered; the other two settings gain 0.17--0.19\,dB.
Halving the context to 4 views costs 0.86 and 0.33\,dB.
ReSplat behaves differently: with the 8-view checkpoint it gains 0.20\,dB on MTGS from 16 views, mostly from opposing-direction context (${+}0.41$\,dB), and nothing on Ithaca365, where its LPIPS gets worse; its checkpoint trained for 16 views gains 0.49 and 0.27\,dB.
At 16 views and with the same information, \method without a reference matches ReSplat's 16-view checkpoint in PSNR on MTGS (19.69 vs.\ 19.62\,dB) and trails it on Ithaca365 (17.61 vs.\ 18.17\,dB), while its LPIPS is lower on both (0.243 vs.\ 0.325 and 0.257 vs.\ 0.373).
With lighting references, \method leads on both datasets (MTGS 21.23\,dB, Ithaca365 18.31\,dB), but ReSplat is not relit in this table, so that comparison mixes conditions.

\begin{figure*}[t]
  \centering
  \includegraphics[width=\textwidth]{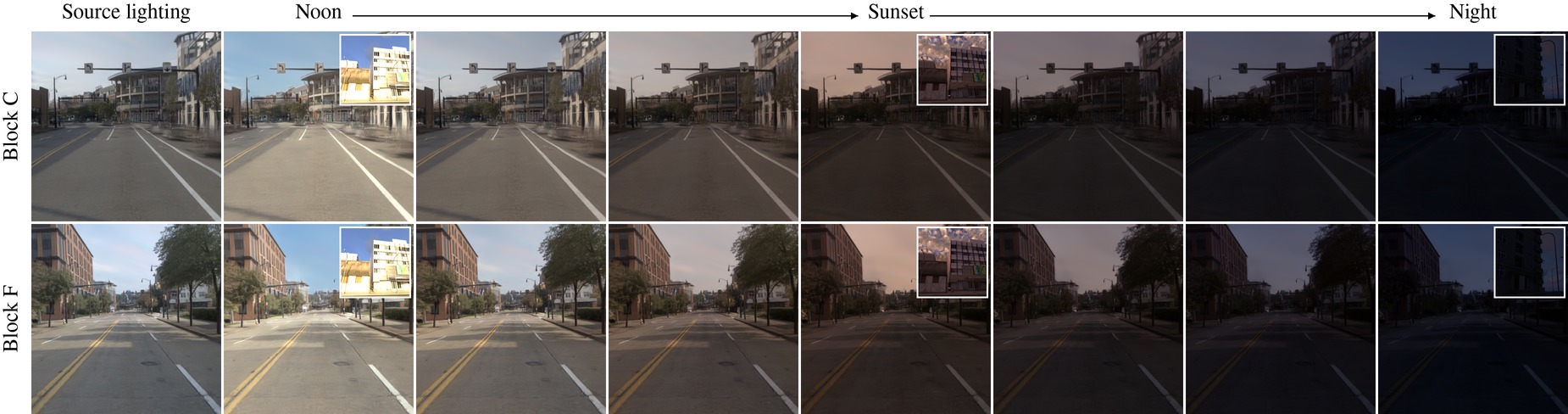}
  \caption{\textbf{Appearance transfer from arbitrary images.} Two reconstructed MTGS streets (200k Gaussians, refined) rendered with their source lighting (left), then relit with lighting codes interpolated between three synthetic WildCity images (insets): noon $\rightarrow$ sunset $\rightarrow$ night. The cameras lie on the context drive. The geometry model never sees the lighting images. The transfer changes color only, so lighting effects that require new light sources or shadows (street lights at night, cast shadows) are not produced. More lightings in Fig.~\ref{fig:relight_wildcity}.}
  \label{fig:relight}
\end{figure*}
\begin{figure*}[t]
  \centering
  \includegraphics[width=\textwidth]{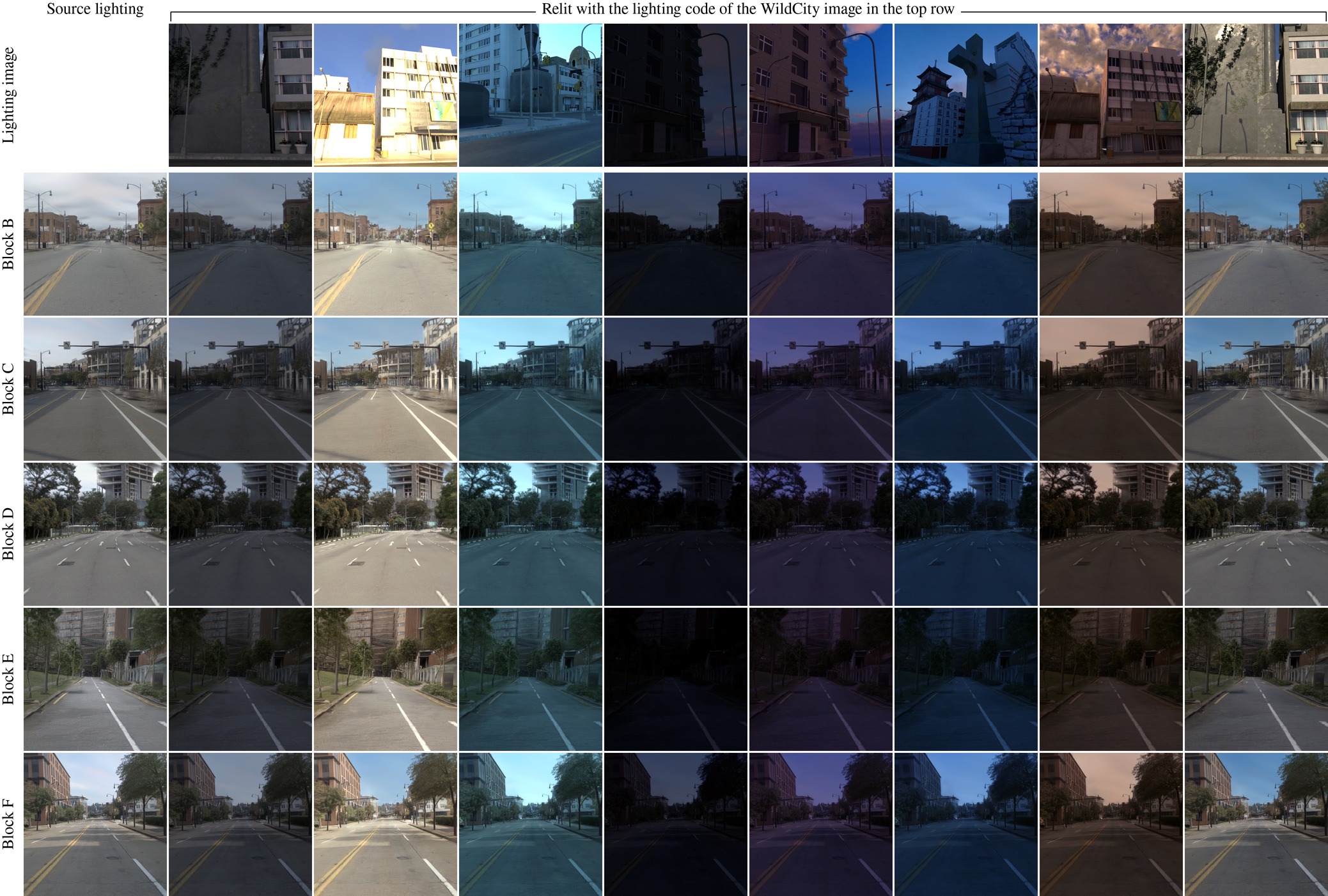}
  \caption{\textbf{Relighting to arbitrary images.} Five MTGS blocks (200k Gaussians, refined), each rendered with its source lighting (left column) and relit with the lighting code of each synthetic WildCity image in the top row. The cameras lie on the context drive. The codes are computed from the lighting images alone; geometry is unchanged across a row.}
  \label{fig:relight_wildcity}
\end{figure*}
\section{Appearance Transfer}
\label{sec:supp_appearance_eval}
A lighting code from any frame of the target drive transfers its lighting, and the color steps of the refiner add about 0.5\,dB more when the references observe the scored surfaces.
The geometry of every refined cloud is bit-identical with and without the reference (Sec.~\ref{sec:appearance}); the reference only changes the colors of the same Gaussians.
In every setting it raises PSNR by 0.8--1.8\,dB but improves LPIPS only slightly (by 0.010--0.018 on MTGS and 0.024--0.031 on Ithaca365), as expected for a color change (Sec.~\ref{sec:supp_metrics}).
The LPIPS gap to ReSplat therefore measures reconstruction quality, not lighting.

\subsection{Target Lighting}
\label{sec:supp_target_code}

\begin{table}[t]
  \centering
  \caption{\textbf{Choice of target lighting} (PSNR/LPIPS, 200k, color transferred after refinement, no color steps, all target frames scored). $^\dagger$Reference-slot pipeline (Sec.~\ref{sec:supp_variants}): backbone and refiner retrained with a target-drive frame in Wild3R's reference slot.}
  \label{tab:appearance_code}
  \footnotesize
  \setlength{\tabcolsep}{4pt}
  \begin{tabular}{lcc}
    \toprule
    Target lighting & MTGS & Ithaca365 \\
    \midrule
    None (source lighting) & 19.92/.242 & 18.48/.252 \\
    Far target-drive frame & 20.39/.241 & 19.17/.244 \\
    Target drive, leave-one-out & \textbf{20.58}/\textbf{.239} & \textbf{19.18}/\textbf{.243} \\
    \textcolor{gray}{Own frame (leaky bound)} & \textcolor{gray}{20.74/.239} & \textcolor{gray}{19.20/.243} \\
    \midrule
    Reference as view 0$^\dagger$ & 20.70/.231 & 18.91/.240 \\
    \bottomrule
  \end{tabular}
\end{table}

Table~\ref{tab:appearance_code} isolates the choice of target code, without the refiner's color steps.
Relighting to the target drive's own lighting adds 0.66\,dB on MTGS and 0.70\,dB on Ithaca365 over source lighting, at nearly equal LPIPS ($-$0.003 and $-$0.009).
The gain is largest when the context comes from a single other drive: on MTGS it rises from 19.64 to 20.57\,dB.
A single frame of the target drive outside the reconstructed window is nearly as good as the leave-one-out code (${-}0.19$/${-}0.01$\,dB), so the branch does not need images of the scored location.
Using the scored frame's own code adds only 0.16\,dB on MTGS and 0.02\,dB on Ithaca365, which bounds how much scene content leaks into the 32-d code.

\subsection{Where the Lighting Reference Comes From}
\label{sec:supp_refdist}

\begin{table}[t]
  \centering
  \caption{\textbf{Where the lighting reference comes from} (200k, no retraining). PSNR/LPIPS gain over no reference (absolute values in the first row; \emph{first half}: first half of the window); the scored frames per row are given in the text. \emph{code}: the reference sets $\mathbf{e}_{\mathrm{tgt}}$ only; \emph{+ refiner}: it also drives the refiner's color steps.}
  \label{tab:refdist}
  \footnotesize
  \setlength{\tabcolsep}{1.5pt}
  \begin{tabular}{@{}lcccc@{}}
    \toprule
    & \multicolumn{2}{c}{MTGS} & \multicolumn{2}{c}{Ithaca365} \\
    \cmidrule(lr){2-3}\cmidrule(lr){4-5}
    Reference & code & + refiner & code & + refiner \\
    \midrule
    None (abs.) & \multicolumn{2}{c}{19.88/.235} & \multicolumn{2}{c}{18.65/.249} \\
    \midrule
    Outside window & $+$.64/$-$.003 & $+$.67/$-$.003 & $+$.71/$-$.010 & $+$.67/$-$.010 \\
    First half & $+$.85/$-$.004 & $+$1.13/$-$.009 & $+$.70/$-$.010 & $+$.82/$-$.016 \\
    Adjacent (ours) & $+$.83/$-$.004 & $+$1.37/$-$.015 & $+$.74/$-$.011 & $+$1.25/$-$.030 \\
    \textcolor{gray}{Scored (leaky)} & \textcolor{gray}{$+$.98/$-$.005} & \textcolor{gray}{$+$1.36/$-$.015} & \textcolor{gray}{$+$.75/$-$.011} & \textcolor{gray}{$+$1.30/$-$.033} \\
    \bottomrule
  \end{tabular}
\end{table}

Table~\ref{tab:refdist} moves the reference away from the scored views without retraining.
Each row is scored on frames it does not use as references: the odd frames for a single frame outside the window (at least 10.9\,m from the nearest scored view) and for the adjacent even frames, and the second half of the window when its first half is the reference; the leaky row scores the even frames themselves, and its code-only entry uses each frame's own code.
The lighting code saturates early: a single frame outside the reconstructed window already gives $+$0.64\,dB on MTGS and $+$0.71\,dB on Ithaca365, and even the scored frame's own code adds only 0.34 and 0.04\,dB more.
The color steps behave differently.
With a reference outside the window they add nothing ($+$0.03 and $-$0.04\,dB), with the first half of the window $+$0.28 and $+$0.12\,dB, and with adjacent frames $+$0.54 and $+$0.51\,dB together with the largest LPIPS gain ($-$0.011 and $-$0.019).
Giving the color steps the scored frame itself is no better than its neighbors on MTGS (21.21 vs.\ 21.26\,dB) and only 0.05\,dB better on Ithaca365 (19.95 vs.\ 19.90\,dB), so two compact refinement steps do not overfit the reference pixels.
The color steps thus correct the appearance of surfaces that the reference also observes, which is why we place the references next to the rendered views; the lighting itself is transferred by the code from any frame of the target drive.

\begin{figure}[t]
  \centering
  \includegraphics[width=\linewidth]{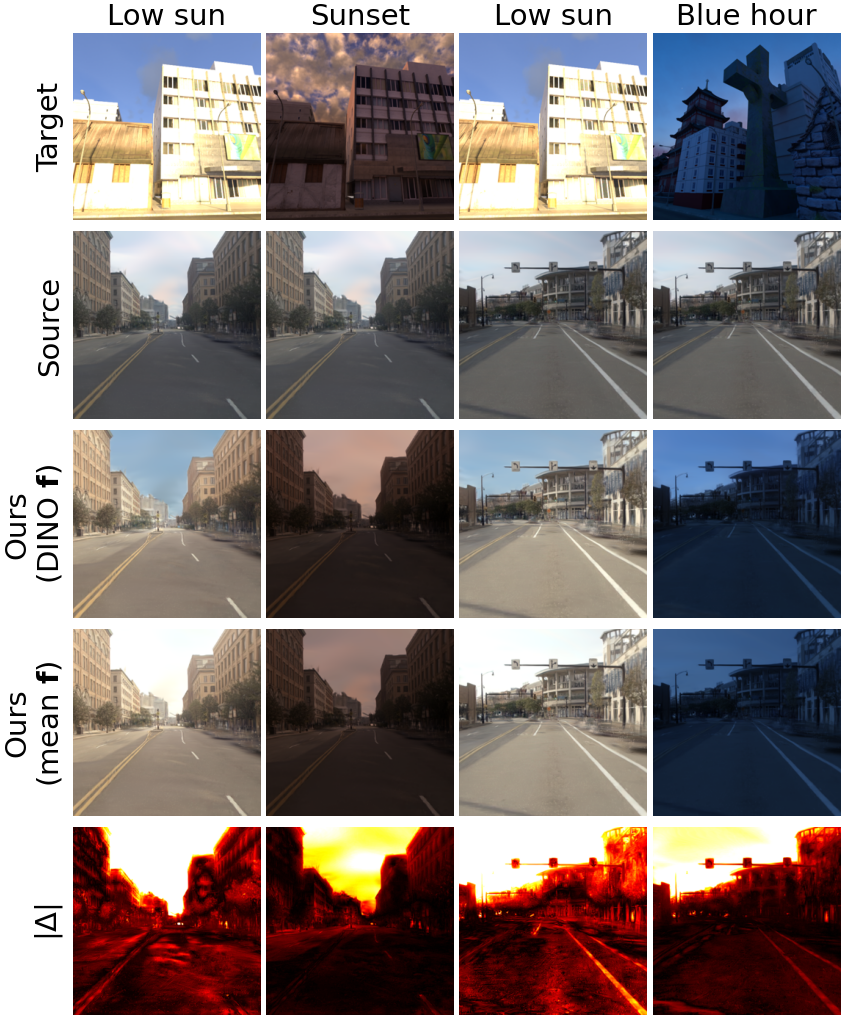}
  \caption{\textbf{What the local feature does.} Two MTGS scenes (200k, refined) relit to WildCity lightings. Rows: target image, source lighting, relit with $\mathbf{f}_j$, relit with its scene mean, and their absolute difference (white ${\ge}40$ levels). With $\mathbf{f}_j$ the sky keeps its color; the mean treats it like the ground.}
  \label{fig:localfeat}
\end{figure}
\subsection{Lighting Inside the Refiner}
\label{sec:supp_light_refiner}

\begin{table}[t]
  \centering
  \caption{\textbf{Lighting inside the refiner} (PSNR/LPIPS, 200k, $N{=}2$). \emph{Far}: one target-drive frame outside the window gives code and reference; all frames scored. \emph{Adjacent}: the even frames give both; odd frames scored. \emph{Relight} and \emph{+ relit residuals} build on the row above; each \emph{+ refiner} row adds color steps to the last row without them (\emph{Relight} in the far block, \emph{+ relit residuals} in the adjacent block), so the three \emph{+ refiner} rows of the adjacent block are alternatives, and \emph{Ref.\ as view 0} is a separate pipeline. \emph{Relit residuals}: geometry steps compare with the cloud relit to each context view's code. \emph{+ refiner}: the refiner's color steps; \emph{far-/adj.-trained}: refiner trained with far or adjacent references, color steps after the geometry steps (\emph{geo.\ first}) or interleaved (\emph{interl.}); the gray row interleaves them at test time and leaks. $^\dagger$Reference-slot pipeline (Sec.~\ref{sec:supp_variants}); all other rows use the final pipeline.}
  \label{tab:appearance_refiner}
  \footnotesize
  \setlength{\tabcolsep}{2.5pt}
  \begin{tabular}{lccc}
    \toprule
    & \multicolumn{2}{c}{MTGS} & Ithaca365 \\
    \cmidrule(lr){2-3}\cmidrule(lr){4-4}
    Refiner variant & all & Same dir. & all \\
    \midrule
    \multicolumn{4}{l}{\emph{Far reference; all target frames scored}} \\
    No transfer & 19.92/.242 & 20.39/.227 & 18.48/.252 \\
    Relight (far code) & 20.39/.241 & 20.73/.227 & 19.17/.244 \\
    Ref.\ as view 0$^\dagger$ & 20.26/.239 & 20.84/.221 & 18.78/.244 \\
    + refiner & \textbf{20.54}/\textbf{.233} & \textbf{20.93}/\textbf{.219} & \textbf{19.32}/\textbf{.239} \\
    \midrule
    \multicolumn{4}{l}{\emph{Adjacent references (even frames); odd frames scored}} \\
    No transfer & 19.94/.241 & 20.40/.226 & 18.49/.251 \\
    Relight & 20.62/.238 & 20.94/.224 & 19.18/.243 \\
    + relit residuals & 20.62/.235 & 20.96/.222 & 19.26/.246 \\
    + refiner, far-trained (ours) & 21.26/\textbf{.220} & 21.65/\textbf{.205} & 19.90/\textbf{.218} \\
    + refiner, adj.-trained, interl. & 21.60/.223 & 22.09/.208 & \textbf{20.28}/\textbf{.218} \\
    + refiner, adj.-trained, geo.\ first & \textbf{21.74}/.228 & \textbf{22.31}/.208 & 20.07/.226 \\
    \textcolor{gray}{\hspace{0.6em}interleaved at test (leaks)} & \textcolor{gray}{22.31/.206} & \textcolor{gray}{22.90/.188} & \textcolor{gray}{20.47/.205} \\
    Ref.\ as view 0$^\dagger$ & 20.71/.230 & 21.07/.214 & 18.93/.239 \\
    \bottomrule
  \end{tabular}
\end{table}

Table~\ref{tab:appearance_refiner} adds the refiner changes of Sec.~\ref{sec:appearance} one at a time.
Rendering each context view in its own lighting has a small effect (${-}0.003$ LPIPS on MTGS at equal PSNR; ${+}0.08$\,dB and ${+}0.003$ LPIPS on Ithaca365).
\emph{Far reference.}
With a single target-drive frame outside the window, the color steps add 0.15\,dB and ${-}0.008$ LPIPS on MTGS and 0.15\,dB and ${-}0.005$ LPIPS on Ithaca365 over relighting with the same code.
The PSNR gain comes from single held-out ($+0.27$\,dB on MTGS) and same-direction synthesis ($+0.20$\,dB); on opposing-direction synthesis only LPIPS improves (0.259 vs.\ 0.268). The gain holds at 100k, 300k, and 400k Gaussians (LPIPS $0.244$, $0.229$, $0.227$ vs.\ $0.249$, $0.237$, $0.235$ without the color steps; the same refiner serves all budgets).
\emph{Adjacent references.}
Adjacent target-drive frames give a much larger gain: the color steps of our refiner add 0.64\,dB on both datasets and lower LPIPS by 0.015 (MTGS) and 0.028 (Ithaca365) over the lighting-consistent refiner, and a refiner trained on the adjacent references adds 0.98 and 1.02\,dB (interleaved-trained row).
Because those references are next to the scored views, this gain comes from the references observing the scored surfaces rather than from lighting transfer alone: swapping only the code (Table~\ref{tab:appearance_code}) is worth ${+}0.19$\,dB over the far code (see also Table~\ref{tab:refdist}).
\emph{Geometry first.}
Interleaving geometry and color steps at inference raises PSNR further (MTGS 22.31 vs.\ 21.60\,dB), but the refined means differ with and without references: target-drive information reaches the geometry through the recurrent state.
A refiner trained with geometry first roughly matches the interleaved-trained refiner run geometry first (21.74 vs.\ 21.60\,dB on MTGS at 200k; 20.07 vs.\ 20.28\,dB on Ithaca365), so the color steps do not need to be trained with the schedule used at inference; on MTGS, training geometry first helps at 100k (21.84 vs.\ 21.37\,dB) and not at 400k (21.44\,dB for both, with higher LPIPS).

\subsection{Separate Branch vs.\ Reference View}
\label{sec:supp_refslot}
In the reference-slot pipeline (Sec.~\ref{sec:supp_variants}), a target-drive frame takes Wild3R's reference slot, and the backbone and refiner are adapted to it.
Compared with our appearance branch on the pipeline version that this model was trained from, it is 0.17\,dB better on MTGS with lower LPIPS (20.70 vs.\ 20.53\,dB, 0.231 vs.\ 0.239), but 0.32\,dB worse on Ithaca365 (18.91 vs.\ 19.23\,dB).
Because the reference enters the backbone, it also changes the geometry: the same backbone evaluated without a reference falls 0.3--0.5\,dB below a backbone never trained with references.
Our branch leaves the geometry model untouched, so the no-reference case is exactly the geometry model.

\begin{figure*}[t]
  \centering
  \includegraphics[width=\textwidth]{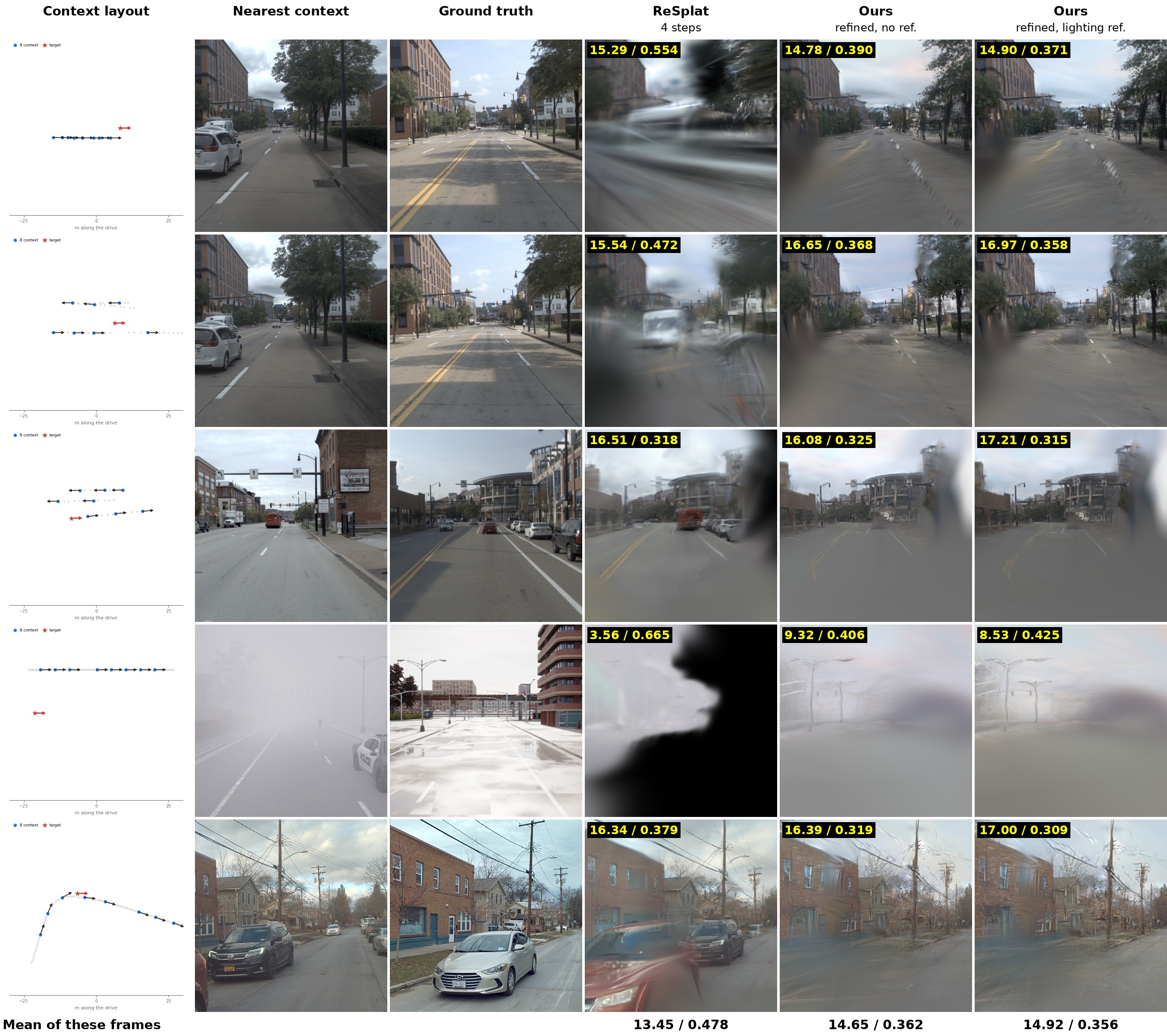}
  \caption{\textbf{Failure cases.} Five frames chosen by hand from the lowest-PSNR frames of the 48 test splits (rows 1--3 and 5) and of \bench (row 4, Blend, Hard), so this is not a fair sample. The left column shows the eight context cameras (blue) and the target camera (red) from above. In row 4 the ground truth is the traffic-free twin of the target frame. Numbers are static PSNR/LPIPS at 200k Gaussians, with the means over these frames in the footer.}
  \label{fig:failures}
\end{figure*}
\subsection{Per-Gaussian Local Feature}
\label{sec:supp_localfeat}

\begin{table}[t]
  \centering
  \caption{\textbf{Per-Gaussian local feature} (PSNR/LPIPS, no retraining). Real data: full pipeline at 200k with $\mathbf{f}_j$ replaced in every call of $\Gamma_\psi$; \emph{static} as in Table~\ref{tab:main}, \emph{sky} PSNR on sky pixels only. \emph{WildCity}: image-space transfer on 1{,}440 held-out relit pairs. Oracle: a $3{\times}4$ color map fitted on the target.}
  \label{tab:localfeat}
  \footnotesize
  \setlength{\tabcolsep}{2.5pt}
  \begin{tabular}{lccccc}
    \toprule
    & \multicolumn{2}{c}{MTGS} & \multicolumn{2}{c}{Ithaca365} & WildCity \\
    \cmidrule(lr){2-3}\cmidrule(lr){4-5}\cmidrule(lr){6-6}
    Local feature & static & sky & static & sky & pairs \\
    \midrule
    No transfer & 19.88/.235 & 19.92 & 18.65/.249 & 19.90 & 17.78/.205 \\
    $\mathbf{f}_j$ (ours) & \textbf{21.26}/.220 & \textbf{21.37} & \textbf{19.90}/\textbf{.218} & \textbf{22.38} & \textbf{25.95}/\textbf{.088} \\
    $\mathbf{f}_j :=$ scene mean & 21.22/.220 & 20.27 & 19.76/.219 & 21.87 & 25.00/.097 \\
    $\mathbf{f}_j := \mathbf{0}$ & 21.24/\textbf{.219} & 20.39 & 19.72/.219 & 21.54 & 24.35/.107 \\
    \textcolor{gray}{Oracle global affine} & -- & -- & -- & -- & \textcolor{gray}{27.75/.111} \\
    \bottomrule
  \end{tabular}
\end{table}

Table~\ref{tab:localfeat} replaces $\mathbf{f}_j$ by its scene mean or by zero, so that all Gaussians of a context view share one color map per target code.
On the static metric of Table~\ref{tab:main} this costs only 0.04\,dB on MTGS and 0.14\,dB on Ithaca365 at nearly equal LPIPS.
That metric excludes the sky, where most of the effect lies: the rendered colors with and without $\mathbf{f}_j$ differ three times as much in the sky as on static surfaces (7.7 vs.\ 2.4 of 255 levels on MTGS).
On sky pixels $\mathbf{f}_j$ is worth 1.1\,dB on MTGS and 0.5\,dB on Ithaca365, and on MTGS it carries most of the sky gain of the transfer (19.92\,dB without transfer, 20.27 with the mean feature, 21.37 with $\mathbf{f}_j$).
On WildCity relit pairs, which include the sky, the local feature is worth 0.95\,dB and 0.009 LPIPS, and 1.7\,dB on bright targets (mean luminance ${\ge}0.35$).
On static surfaces the lighting changes of real drives are small: relighting a real scene to its target drive changes the mean color by about 4 of 255 levels, and the codes of different real drives move it by about 2 levels against each other, compared with 34 levels between WildCity lightings.
What $\mathbf{f}_j$ adds is semantic (Fig.~\ref{fig:localfeat}).
Relighting the six MTGS traversal scenes to a low-sun WildCity code darkens the rendered sky by 38 levels while the road brightens by 44; with $\mathbf{f}_j$ replaced by its mean the sky brightens by 31 levels along with the ground and saturates, and the spread of the per-class mean color change shrinks from 29 to 6 levels.
Under sunset and blue-hour codes $\mathbf{f}_j$ keeps a glow in the sky that the mean feature turns dark (sky $-$61 vs.\ $-$99 levels at sunset), while the road changes by less than 4 levels.
It does not recover local lighting: on WildCity the transfer stays below the global affine oracle in PSNR (25.95 vs.\ 27.75\,dB), and dark targets, where street lights and headlights would matter, gain only 0.5\,dB from $\mathbf{f}_j$.

\subsection{Relighting to Arbitrary Images}
\label{sec:supp_relight}

Figure~\ref{fig:relight} interpolates lighting codes between three WildCity images, and Fig.~\ref{fig:relight_wildcity} extends it to five MTGS blocks and more lightings: each code is computed from a single WildCity image, and the geometry is identical across a row, so only colors change.

\section{Runtime and Failure Cases}
\label{sec:supp_runtime}
A reconstruction takes 0.9--1.6\,s at 8.8--12.4\,GiB, and most failures are regions that the context barely observes.

\subsection{Measured Reconstruction and Rendering Cost}
Table~\ref{tab:runtime} splits the runtime of \method by stage.
The backbone, the masks, and the compact head do not depend on the budget, so reconstruction time grows only with refinement.
Refinement finds each Gaussian's nearest neighbors with the steps of Morton-order $k$-NN graph construction~\cite{connor2010knng}, applied per block of points on the GPU rather than per point: points are sorted in Morton order, a search among adjacent blocks gives each block an upper bound on its neighbor distances, and every block whose bounding box lies beyond that bound is skipped, which leaves the result exact except for the choice among equidistant points.
Every budget renders with gsplat~\cite{ye2025gsplat} at more than 800 views per second.
The compact head sets the memory peak up to 200k, and refinement sets it above that.
At ReSplat's budget of 100k, \method takes 0.93\,s against 0.38\,s: more than half of the difference is transient masking, which ReSplat does not perform, and the rest is the larger backbone, lighting, and the per-view relighting and reference frames during refinement.
Training time per stage is given in Table~\ref{tab:supp_recipe}.

\begin{table}[t]
  \centering
  \caption{\textbf{Runtime of \method by stage} on MTGS (18 splits, eight $448{\times}448$ context views, 9--12 lighting references, one RTX PRO 6000, without model loading and disk access). Peak memory with all models resident. Transient masks: 0.03\,s for the context views, 0.28\,s for the references.}
  \label{tab:runtime}
  \footnotesize
  \setlength{\tabcolsep}{5pt}
  \begin{tabular}{lcccc}
    \toprule
    Stage (s) & 100k & 200k & 300k & 400k \\
    \midrule
    Backbone & 0.21 & 0.21 & 0.21 & 0.21 \\
    Transient masks & 0.31 & 0.31 & 0.31 & 0.31 \\
    Entropy map & 0.02 & 0.02 & 0.02 & 0.02 \\
    Compact head & 0.07 & 0.07 & 0.07 & 0.07 \\
    Lighting codes & 0.03 & 0.03 & 0.03 & 0.03 \\
    Adapter & 0.03 & 0.03 & 0.03 & 0.03 \\
    Refinement (2+2 steps) & 0.26 & 0.48 & 0.70 & 0.90 \\
    \midrule
    Total & 0.93 & 1.15 & 1.37 & 1.57 \\
    Peak GPU memory (GiB) & 8.8 & 8.8 & 10.5 & 12.4 \\
    Rendering (views/s) & 1{,}579 & 1{,}199 & 975 & 801 \\
    \bottomrule
  \end{tabular}
\end{table}

\begin{table}[t]
  \centering\small
  \caption{\textbf{Lighting references against runtime} (200k, MTGS, same odd frames scored, timed as in Table~\ref{tab:runtime}). \emph{Up to 12}: all even frames (9--12); \emph{6 frames}: six evenly spaced even frames for the color steps and the code; \emph{None}: no image of the target drive.}
  \label{tab:supp_refcount}
  \setlength{\tabcolsep}{5pt}
  \begin{tabular}{lcccc}
    \toprule
    Lighting references & Runtime & PSNR$\uparrow$ & SSIM$\uparrow$ & LPIPS$\downarrow$ \\
    \midrule
    Up to 12 frames & 1.15\,s & \textbf{21.26} & \textbf{0.649} & \textbf{0.220} \\
    6 frames & 1.00\,s & 21.21 & 0.647 & 0.221 \\
    None & \textbf{0.80\,s} & 19.88 & 0.626 & 0.235 \\
    \bottomrule
  \end{tabular}
\end{table}

Table~\ref{tab:supp_refcount} shows the trade-off between the number of lighting references and runtime.
Six references save about 0.15\,s at every budget and cost 0.03--0.05\,dB PSNR and at most 0.001 LPIPS.
Without any reference, \method is 0.33--0.36\,s faster and loses 1.2--1.4\,dB.

\subsection{Coverage and Failure Cases}
Figure~\ref{fig:failures} shows frames from the low end of the 48 test splits and of \bench.
Most failures are places that the context barely observes.
When the target drives in another lane or the context bends away from it, the road and the near facades smear (rows 1, 5).
Opposing-direction context leaves surfaces that face the target unobserved, which appear as blurred patches or missing regions (rows 2, 3).
When the context drive is recorded in dense fog, the fog is part of the reconstructed scene: \method renders the haze into a clear target frame, and lighting references, which change only colors, cannot remove it (row 4).
ReSplat fails on the same frames: its PSNR is slightly higher only in row 1, it collapses in the fog of row 4, and its LPIPS is worse on all of them.

\end{document}